\documentclass[pdflatex,sn-mathphys-num]{sn-jnl}% Math and Physical Sciences Numbered Reference Style 
\usepackage{graphicx}%
\usepackage{multirow}%
\usepackage{amsmath,amssymb,amsfonts}%
\usepackage{amsthm}%
\usepackage{mathrsfs}%
\usepackage[title]{appendix}%
\usepackage{xcolor}%
\usepackage{textcomp}%
\usepackage{manyfoot}%
\usepackage{booktabs}%
\usepackage{algorithm}%
\usepackage{algorithmicx}%
\usepackage{algpseudocode}%
\usepackage{listings}%
\usepackage{times}%
\usepackage{latexsym}%
\usepackage{booktabs}%
\usepackage{multicol}%
\usepackage{tabularx}%
\usepackage{subfig}%
\usepackage{subcaption}%
\usepackage{caption}%
\usepackage{tcolorbox}%
\usepackage{lmodern}%
\usepackage[svgnames]{xcolor}
\newtcolorbox{infobox}[1][]{
colback=SeaGreen!12, % soft green background
colframe=SeaGreen!55!black, % slightly darker green frame
boxrule=0.6pt,
arc=1mm,
left=6pt,right=6pt,top=6pt,bottom=6pt,
before skip=8pt, after skip=8pt,
title=#1
}

\begin{document}

\title[Article Title]{Introducing TrGLUE and SentiTurca: A Comprehensive Benchmark for Turkish General Language Understanding and Sentiment Analysis}

%%=============================================================%%
%% GivenName	-> \fnm{Joergen W.}
%% Particle	-> \spfx{van der} -> surname prefix
%% FamilyName	-> \sur{Ploeg}
%% Suffix	-> \sfx{IV}
%% \author*[1,2]{\fnm{Joergen W.} \spfx{van der} \sur{Ploeg} 
%%  \sfx{IV}}\email{iauthor@gmail.com}
%%=============================================================%%

\author{\fnm{Duygu} \sur{Altinok}}\email{duygu@turkish-nlp-suite.com}

\affil{\orgname{Independent Researcher}, \orgaddress{\state{Berlin}, \country{Germany}}}

%%==================================%%
%% Sample for unstructured abstract %%
%%==================================%%

\abstract{Evaluating the performance of various model architectures, such as transformers, large language models (LLMs), and other NLP systems, requires comprehensive benchmarks that measure performance across multiple dimensions. Among these, the evaluation of natural language understanding (NLU) is particularly critical as it serves as a fundamental criterion for assessing model capabilities. Thus, it is essential to establish benchmarks that enable thorough evaluation and analysis of NLU abilities from diverse perspectives. While the GLUE benchmark has set a standard for evaluating English NLU, similar benchmarks have been developed for other languages, such as CLUE for Chinese, FLUE for French, and JGLUE for Japanese. However, no comparable benchmark currently exists for the Turkish language. To address this gap, we introduce TrGLUE, a comprehensive benchmark encompassing a variety of NLU tasks for Turkish. In addition, we present SentiTurca, a specialized benchmark for sentiment analysis. To support researchers, we also provide fine-tuning and evaluation code for transformer-based models, facilitating the effective use of these benchmarks. TrGLUE comprises Turkish-native corpora curated to mirror the domains and task formulations of GLUE-style evaluations, with labels obtained through a semi-automated pipeline that combines strong LLM-based annotation, cross-model agreement checks, and subsequent human validation. This design prioritizes linguistic naturalness, minimizes direct translation artifacts, and yields a scalable, reproducible workflow. With TrGLUE, our goal is to establish a robust evaluation framework for Turkish NLU, empower researchers with valuable resources, and provide insights into generating high-quality semi-automated datasets.}

%%\pacs[JEL Classification]{D8, H51}

%%\pacs[MSC Classification]{35A01, 65L10, 65L12, 65L20, 65L70}

\maketitle

\section{Introduction}\label{sec1}

Given the advancements in transformer models, such as BERT \cite{devlin-etal-2019-bert} and large language models (LLMs) \cite{minaee2024largelanguagemodelssurvey}, it has become essential to benchmark these models from various perspectives. Evaluating natural language understanding (NLU) is particularly important, and having a benchmark for evaluating and analyzing NLU abilities is crucial. The General Language Understanding Evaluation (GLUE) benchmark \cite{wang-etal-2018-glue} has been widely used for English. However, benchmarks for languages other than English have also been developed, including CLUE for Chinese \cite{xu-etal-2020-clue}, FLUE for French \cite{le2019flaubert}, JGLUE for Japanese \cite{kurihara-etal-2022-jglue}, and KLUE for Korean \cite{park2021kluekoreanlanguageunderstanding}.

In the case of Turkish, several datasets have been created for tasks like text classification, sentiment analysis \cite{demirtas}, and hate speech \cite{toraman2022large}. However, there is a lack of datasets for paraphrase, similarity, and inference tasks, with the exception of a Semantic Textual Similarity (STS) dataset translated by Beken et al. \cite{beken-fikri-etal-2021-semantic}. As a result, there is currently no standard NLU benchmark for Turkish, and the existing datasets are scattered and not comprehensive. To address this gap, we have developed a centralized and easily accessible benchmark called TrGLUE, which is hosted on Hugging Face.

Despite the availability of transformer models like BERTurk \cite{stefan_schweter_2020_3770924} and even some LLMs, there is a scarcity of training and evaluation datasets in Turkish. Most models are trained on OSCAR \cite{oscar1, oscar2} and/or mC4 corpora \cite{mc4}, which lack variability, and the situation is even worse for evaluation datasets. BERTurk has been evaluated in limited areas such as sequence classification, named entity recognition (NER), part-of-speech (POS) tagging, and question answering. However, there is little information available about the construction process or corpus statistics of the question answering dataset used by BERTurk.

Existing evaluations in the field lack comprehensive coverage and reproducibility, either focusing superficially on a few dimensions or relying on local datasets that are not shared. Moreover, the absence of a standardized benchmarking dataset for Turkish creates challenges in comparing results across different models. Another issue is using translations instead of building from natural text data, which introduces low-quality translations, cultural biases as well as not considering agglutinative nature of Turkish, also introducing evaluation metric defects. 

To address these issues, we propose the creation of the Turkish General Language Understanding Evaluation (TrGLUE) benchmark. Similar to the widely used GLUE benchmark for English, TrGLUE aims to establish a standardized evaluation framework for Turkish NLP models. It comprises diverse tasks of text classification and sentence pair classification, carefully designed to assess various aspects of natural language understanding in Turkish. 

We also introduce SentiTurca, a sentiment analysis focused benchmark. SentiTurca encompasses datasets from different domains and covers a wide range of sentiments, including the Turkish Hate Map (TuHaMa), the most extensive Turkish hate speech dataset available. These resources enable researchers to explore sentiment analysis in diverse contexts and address the challenges of hate speech in the Turkish language.

During the construction of TrGLUE, we mirrored the task design and domain coverage of the original GLUE-style datasets. Source texts were collected from native Turkish resources, including Wikipedia (tr), news outlets, social platforms, and public forums, to ensure linguistic naturalness and domain comparability.

For STS-B, whose task formulation is language-agnostic, we opted for translation followed by multiple human review passes to adapt cultural references and idiomatic usage for Turkish. Aside from STS-B, we avoided direct translations and relied on Turkish-native corpora.

To scale annotation while maintaining quality, we adopted a semi-automated, disagreement-driven pipeline. For each task, we first assembled large unlabeled Turkish instance pools and translated the corresponding English task definitions into Turkish guidelines. We trained a lightweight sentence-transformer\cite{reimers-2020} classifier to generate preliminary labels and, in parallel, prompted a permissively licensed, state-of-the-art LLM, Snowflake Arctic \cite{arctic2024} to produce label hypotheses for the same instances. We then reconciled the two signals: examples with classifier-LLM disagreement were prioritized for human annotation, while agreement cases underwent stratified spot checks. This triage concentrates expert effort on ambiguous or difficult items, preserves scalability for clear-cut cases, and yields measurable quality control. We document all prompts, model versions, and decision thresholds to ensure reproducibility and to facilitate re-annotation or extension to new domains.

Additionally, we provide detailed documentation and guidelines for dataset creation in non-English languages, facilitating the development of benchmarks for other languages. By sharing the dataset creation process and relevant prompts, we encourage researchers to contribute to the expansion of NLU benchmarks for a wide range of languages.

Our contributions can be summarized as follows:

\begin{itemize}
    \item We introduce TrGLUE, an open-source NLU benchmarking dataset specifically designed for evaluating Turkish language models. It serves as a comprehensive and standardized benchmark, addressing the lack of resources for evaluating Turkish NLP across various tasks.
    \item Each dataset included in TrGLUE is unique and carefully crafted to ensure quality and diversity, enabling robust evaluations and comparisons of Turkish language models.
    \item SentiTurca provides a large-scale and diverse sentiment analysis dataset for the Turkish language, including the extensive Turkish Hate Map (TuHaMa), facilitating research on hate speech challenges in Turkish.
    \item To facilitate the use of TrGLUE and SentiTurca, we provide fine-tuning and evaluation scripts, simplifying the adaptation and evaluation of models on these benchmarks.
    \item By utilizing our tools, we conducted assessments on various Turkish transformer models as well as popular open-source and proprietary LLMs. As a result, our research also delves into exploring the Turkish NLU proficiencies of certain well-established LLMs. 
    \item Synthetic data quality is verified through a semi-automated pipeline combining cross-model agreement checks and targeted human validation, keeping the process cost-efficient and reproducible.
\end{itemize}

Both TrGLUE and SentiTurca are freely available through their respective repositories on Hugging Face\footnote{\parbox{\linewidth}{TrGLUE: \url{https://huggingface.co/datasets/turkish-nlp-suite/TrGLUE}\\ SentiTurca: \url{https://huggingface.co/datasets/turkish-nlp-suite/SentiTurca}}} and GitHub. \footnote{\parbox{\linewidth}{TrGLUE: \url{https://github.com/turkish-nlp-suite/TrGLUE}\\ SentiTurca: \url{https://github.com/turkish-nlp-suite/SentiTurca}}}

\begin{table}[h]
\caption{All tasks involve classifying single sentences or sentence pairs, except for STS-B, which is a task involving regression. MNLI consists of three classes, while the remaining classification tasks have two classes. For TrSTS-B, no test set is released due to the limited dataset size; all evaluations are conducted on the development split. QA stands for question answering.}\label{tab:tasks}%
\begin{tabular}{lrrrll}
 \toprule
\textbf{Corpus} & \textbf{Train} & \textbf{Dev} & \textbf{Test} & \textbf{Task} & \textbf{Metrics}  \\
\midrule
\multicolumn{6}{c}{Single-Sentence Tasks}\\
\midrule
TrCoLA & \textbf{7.9K} & \textbf{1K} & \textbf{1K}& acceptability & Matthews' corr. \\
TrSST-2 & \textbf{60K} & \textbf{8.9K} & \textbf{8.9K} & sentiment & acc./F1  \\
\midrule
\multicolumn{6}{c}{Similarity and Paraphrase Tasks}\\
\midrule
TrMRPC & \textbf{3.18K} & \textbf{1.0K} & \textbf{1.0K} & paraphrase & acc./F1  \\
TrSTS-B & \textbf{2.46K} & \textbf{0.6K} & \textbf{-} & sentence similarity & Pearson/Spearman corr.  \\
TrQQP & \textbf{309K} & \textbf{30K} & \textbf{30K} & paraphrase & acc./F1  \\
\midrule
\multicolumn{6}{c}{Inference Tasks} \\
\midrule
TrMNLI & \textbf{165K} & \textbf{18.4K} & \textbf{18.4K}  & NLI & matched/mismatched acc. \\
TrQNLI & \textbf{120K} & \textbf{10K} & \textbf{10K} & QA/NLI & acc.  \\
TrRTE & \textbf{3.78K} & \textbf{1.0K} & \textbf{1.0K} & NLI & acc.  \\
\bottomrule
\end{tabular}
\end{table}

\section{Related Work}
The GLUE benchmark, consisting of nine datasets, initially paved the way for evaluating NLU models. It includes tasks such as sentence classification and sentence pair classification, with a focus on natural language inference (NLI). GLUE's success has inspired the creation of similar benchmarks in various languages, such as CLUE \cite{xu-etal-2020-clue}, FLUE \cite{le2019flaubert}, KLUE \cite{park2021kluekoreanlanguageunderstanding}, IndicGLUE \cite{indicnlpsuite}, ALUE \cite{seelawi-etal-2021-alue}, JGLUE \cite{kurihara-etal-2022-jglue}, Napolab \cite{Chaves_Rodrigues_napolab_2023}, UINAUIL \cite{basile-etal-2023-uinauil}, SuperGLEBer \cite{pfister-hotho-2024-supergleber}  and CLUB \cite{club}, targeting Chinese, French, Korean, Indian languages, Arabic, Japanese, Portuguese, Italian, German and Catalan, respectively.

The development of multilingual benchmarks has seen significant progress with the emergence of resources such as XGLUE \cite{liang-etal-2020-xglue}, XTREME \cite{xtreme}, and XTREME-R \cite{xtremer}, which incorporate datasets spanning multiple languages. However, Turkish remains underrepresented in these benchmarks, offering only limited and relatively small datasets. This underrepresentation has highlighted the growing need for comprehensive and large-scale natural language understanding (NLU) benchmarks tailored specifically to the Turkish language.

\subsection{Constructing NLI datasets for Turkish}
\label{sec:nli-cons}
One of the key efforts in creating cross-lingual natural language inference (NLI) datasets is the XNLI benchmark \cite{conneau-etal-2018-xnli}, which includes data for 15 languages, including Turkish. This benchmark was constructed by translating 7,500 instances from the MultiNLI dataset \cite{williams-etal-2018-broad} into each target language using crowd-sourced human translators. For evaluation, the authors employed back-translation approaches and multilingual sentence encoders. Specifically, the multilingual sentence encoders were trained to align sentence embeddings and word embeddings across languages. For Turkish, the back-translation approach performed the best. While the XNLI benchmark provides accurate translations due to human involvement, the costs associated with human translation were a significant consideration.

A notable effort focused specifically on Turkish NLI datasets is the work by \cite{budur-etal-2020-data}. In this study, the authors translated both the SNLI and MultiNLI datasets into Turkish using the Amazon Translate engine. To ensure dataset quality, the authors employed human annotators to evaluate the translations and verify that semantic relations were preserved. A total of 500 example sets were distributed to four annotators, who assessed both the translation quality and label correctness. Their findings indicated that the datasets were of high quality, demonstrating that machine translation is a feasible and efficient way to construct Turkish NLI datasets. These datasets also offer a commercial license, making them accessible for various applications.

Our work builds on these efforts with three key advancements. First, we minimize translation: with the exception of TrSTS-B, all datasets are sourced from native Turkish text. Second, we leverage recent advances in LLMs to scale annotation while maintaining quality. Third, we retain full provenance and licensing information for all sources to ensure transparency and reuse.

Concretely, we sample inputs (premises, sentences, questions, reviews) from licensed Turkish sources across multiple registers and genres. We adapt the English task definitions into Turkish annotation guidelines and train a lightweight sentence transformer to assign coarse labels. In parallel, our instruction-tuned, permissively licensed LLM of choice produces label hypotheses via prompting. Instances with classifier-LLM disagreement are routed to human annotators; agreement cases undergo stratified spot checks. This disagreement-driven triage concentrates human effort on ambiguous items and keeps the process cost-effective. Human annotators act as gatekeepers: they reject non-native or awkward phrasing, correct labels where necessary, and allow only minor orthographic fixes. To protect evaluation integrity, we balance lexical overlap across labels, cap trivial negation or heuristic patterns, diversify distractors, and rigorously deduplicate to prevent near-duplicates and train-dev-test leakage.

In practice, we pre-clean inputs and, when relevant, stratify by register/genre. Prompting budgets, temperatures, and overlap/length constraints bound variance, and a double-annotated subset with adjudication is used to calibrate quality. We enforce class balance per split and, when relevant, per genre; we also track rejection and relabeling rates. Task-specific exceptions apply: TrCoLA is entirely hand-crafted by linguists without LLM generation; TrRTE hypotheses are written by humans; TrMNLI hypotheses are generated by the LLM conditioned on labels; TrSST-2 labels are derived directly from review ratings with human spot checks; and TrWNLI is omitted (see Section \ref{sec:wnli} for a Turkish-specific rationale).

\begin{table}[h]
\captionsetup{width=0.8\textwidth}
\caption{Comparison of sizes between the original GLUE datasets and our translated datasets. The last column indicates the percentage of data retained after translation, reflecting some data loss during the process.}
\label{tab:size-comp}
\centering
\begin{tabular}{|l|l|l|l|}
\hline
Corpus & Original size & TR size & Size percent. \\ \hline
CoLA & 10.7K & 9.92K & 0.92 \\ 
SST-2 & 70K & 78K & 1.11 \\
MRPC & 5.8K & 5.18K & 0.89 \\
STS-B & 8.63K & 3.06K &  0.35\\
QQP & 795K  & 369K &  0.46 \\
MNLI & 432K & 202K & 0.46 \\
QNLI & 116K & 140K &  1.20 \\
RTE & 5.77K &  5.78K & 1.00 \\
\hline
\end{tabular}
\end{table}

Finally, we provide a size comparison to situate TrGLUE relative to GLUE (Table \ref{tab:size-comp}). Unlike translation-based pipelines that often lose substantial volume (especially for sentence similarity), our native-first, semi-automated approach produces high-quality datasets quickly and at lower cost while preserving linguistic naturalness. Beyond NLI, we cover the full range of GLUE-style tasks—single-sentence and sentence-pair classification, and regression for TrSTS-B—yielding a complete, reproducible benchmark for Turkish NLU.

\section{TrGLUE Benchmark}

TrGLUE covers single-sentence and sentence-pair classification tasks, summarized in Table \ref{tab:tasks}. We provide Turkish counterparts for all GLUE tasks except WNLI (omitted; see Section \ref{sec:wnli}).

Where possible, we mirror the construction philosophy of GLUE while avoiding translation: inputs come from native, human-written Turkish sources; labels are produced via an LLM-assisted, human-verified workflow. Professional verification and spot-checks were conducted by Co-one (Istanbul)\footnote{Co-one, \url{https://co-one.co/}} with a team of ten native Turkish annotators (balanced by gender) holding advanced degrees in language-related fields. Task-specific procedures are provided in the subsequent subsections, with the following task-level exceptions:

\begin{itemize}
\item TrRTE: hypotheses are fully human-written under linguist-designed guidelines, with linguists serving as lead annotators and adjudicators.
\item TrCoLA: hand-crafted by linguists; no LLM generation.
\item TrSST-2: labels derived directly from review ratings with human spot-checks.
\item TrSTS-B: constructed via translate-then-edit—machine translations were corrected by humans to ensure natural Turkish usage and cultural fit.
\end{itemize}

To maximize tool compatibility, our Hugging Face release preserves GLUE-style split names and metadata, enabling drop-in use with the official GLUE metric\footnote{\url{https://huggingface.co/spaces/evaluate-metric/glue}} and common evaluation pipelines.

\subsection{Single-sentence tasks}
\subsubsection{TrCoLA}
The TrCoLA dataset is a Turkish replica of the original CoLA dataset \cite{cola} and was compiled in a similar manner. The sentences in TrCoLA were sourced from Turkish textbooks and linguistic books, just as in the original dataset. Like CoLA, TrCoLA consists of sentences that exhibit morphological, syntactic, and semantic violations. Each instance in the dataset is a pair $(sentence, label)$, where the label indicates whether the sentence is acceptable or unacceptable.

\begin{table}[h]
\captionsetup{width=0.8\textwidth}
\caption{The linguistic textbooks utilized for TrCoLA sentences are sourced from Anadolu Universitesi's Open and Distance Education System publications \url{https://www.anadolu.edu.tr/en}, which are freely accessible on the internet.}\label{tab:cola-books}%
\begin{tabular}{|l|l|l|}
\hline
Source & \% & Topic \\ \hline
Türk Dili-I & 25 & Overview  \\
Türkçe Sözlü Anlatım & 20  & Verbal expression \\
Türkçe Cümle Bilgisi & 30 & Syntax  \\
Türkçe Ses Bilgisi & 15 & Phon(etics/ology) \\
\hline
\end{tabular}
\end{table}

We collected a total of 3,630 sentences from publicly available Turkish textbooks and linguistic books, as shown in Table \ref{tab:cola-books}. Next, we tasked Snowflake Arctic with generating three variations for each sentence, each containing a violation. This process resulted in 10,890 variations. However, not all of these variations were usable due to known issues of hallucinations in generative models \cite{mckenna2023sourceshallucinationlargelanguage}. Therefore, we sought human review for these variations. The review process was conducted by our annotation-data company Co-one. For each variation, we provided the original sentence, the variation itself, the type of violation, and the human annotators decided whether to keep or discard the instances. After the review, the final dataset size was reduced to 6,686 instances, highlighting the importance of human oversight despite the advancements in generative models. The final dataset size is 9,916 instances, which were then divided into train, development, and test splits. 

Each instance in the dataset was annotated by 4 annotators, and only instances with at least 3 agreeing votes were accepted. Krippendorff's Alpha \cite{krippendorff2004} was computed for the dataset, yielding a high agreement score of $0.91$, indicating strong inter-annotator reliability. Despite their expertise, annotators reported that the dataset was challenging, and a significant portion of instances required multiple readings to make accurate judgments. The complete annotation guidelines are made publicly available in the TrGLUE GitHub repository.

To the best of our knowledge, this dataset is the first of its kind for the Turkish language. We also provide a standalone version of this dataset in its Hugging Face repository\footnote{\url{https://huggingface.co/datasets/turkish-nlp-suite/TrCoLA}}, with a slightly different format than the original CoLA dataset. In the standalone version, we provide the original sentence, the variation, the variation type, and the label, allowing for more in-depth research and analysis.

\subsubsection{TrSST-2}
\label{sec:movies}
The TrSST-2 dataset, utilized for sentiment analysis of film critiques, was constructed by aggregating data from online sources, specifically Sinefil.com and Beyazperde.com. Sinefil.com contributed around 40K reviews, while Beyazperde.com provided approximately 38K reviews, resulting in a combined total of about 78K reviews. Initially, the dataset comprised pairs of $(sentence, star)$ entries, where 'star' denoted a rating from 0 to 10 with a precision of 1. To align with the binary classification structure of the original SST-2\cite{socher-etal-2013-recursive}, reviews rated with 6 stars were excluded due to their mixed sentiments, with some leaning negative and others positive; subsequently, reviews with stars below 5 were categorized as negative, and those with stars above 5 were classified as positive. The final dataset comprised 78K entries, with 67K for training and 8.9K each for validation and testing. For those seeking the complete challenge, the dataset is available with all 10 classes in its dedicated Hugging Face repository.\footnote{\url{https://huggingface.co/datasets/turkish-nlp-suite/BuyukSinema}}

\begin{figure}[!ht]
\centering
\begin{tabular}{ccc}
\subfloat[Distribution of star ratings in the dataset.]{\includegraphics[width=0.32\textwidth]{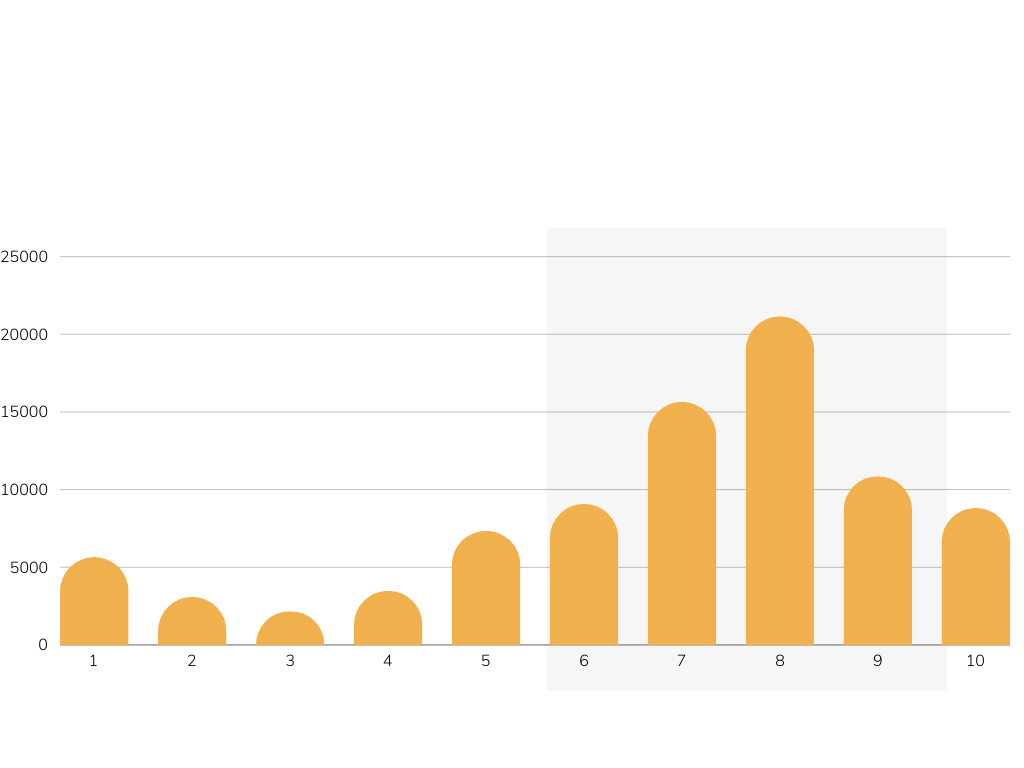}} & \subfloat[Binary classification labels of the dataset.]{\includegraphics[width=0.32\textwidth]{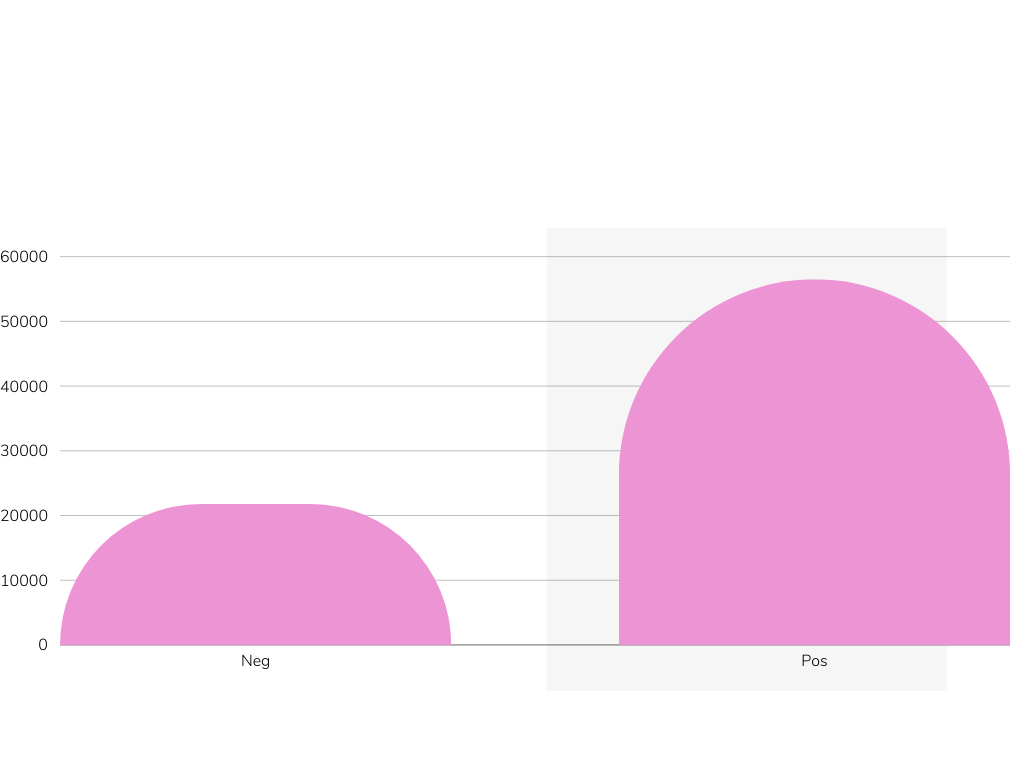}} & \subfloat[Distribution of the number of words in review texts.]{\includegraphics[width=0.32\textwidth]{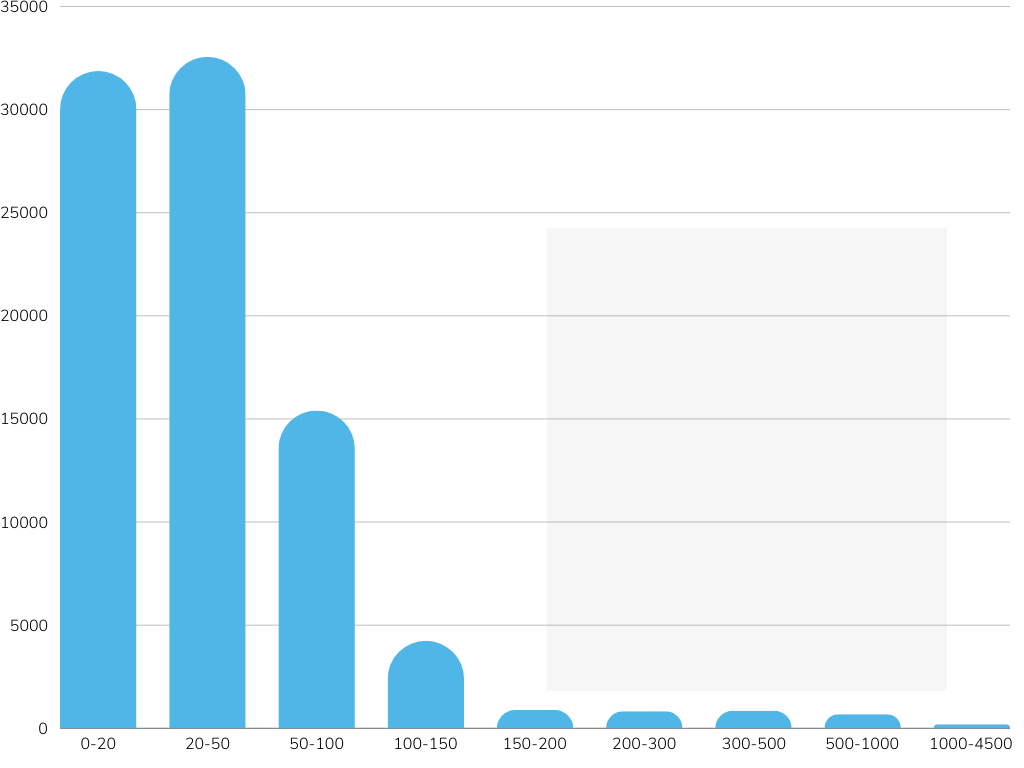}}
\end{tabular}
\caption{Statistical information about the movie reviews dataset.}
\label{fig:movie-stats}
\end{figure}

The distribution of stars within the dataset, illustrated in Figure \ref{fig:movie-stats}, reveals that a significant portion of reviews fall within the \textgreater 7 star range, indicating that the majority of viewers enjoyed the movies they reviewed. However, this also presents a challenge due to the skewed distribution of classes. To address this class imbalance, we utilize the both binary accuracy and F1-score, whereas the original GLUE task employs binary accuracy. Analysis of review lengths, as shown in Figure \ref{fig:movie-stats}, indicates that most reviews contain 0-50 words, with an average length of 48 words. The reviews are generally detailed, offering substantial context, and may include emoticon characters.

It is worth mentioning that there is another Turkish movies dataset by \cite{demirtas}, which contains 10K instances consisting of 5K positive and 5K negative reviews. However, this dataset only includes single-sentence reviews.

\subsection{Sentence-pair tasks}

\subsubsection{TrMRPC} The Microsoft Research Paraphrase Corpus (MRPC) \cite{mrpc} is a dataset consisting of sentence pairs extracted from online news sources. These pairs are annotated by humans to indicate whether the sentences are semantically equivalent. The task associated with this dataset is to classify the sentence pairs as either equivalent or not equivalent. 

We construct a Turkish paraphrase corpus in the style of MRPC using the Havadis news corpus (\(\approx\)745K articles; 2.88\,GB; \(\approx\)315M tokens).\footnote{\url{https://huggingface.co/datasets/turkish-nlp-suite/Havadis}} Articles are stored by newspaper, which facilitates efficient filtering by outlet and author metadata when available \cite{bellaturca}.

We follow a three-phase harvesting pipeline to obtain sentence pairs that include both likely paraphrases and plausible ``near-miss'' non-paraphrases. Our phases mirror the original MRPC methodology, with Turkish-specific adaptations to handle rich morphology. We describe these phases in detail as follows:
 \begin{itemize}
     \item \textbf{Phase 1: Document- and string-based retrieval} Each article is sentence-segmented. We generate initial candidate links using (i) normalized Levenshtein distance over sentence strings; and
     (ii) sentence-position proximity within the same document. To limit topical drift in cross-document candidates, we further restrict pairs to articles from the same outlet and (when available) the same author.
     \item \textbf{Phase 2: Lexical/length filtering with lemmas} We adapt MRPC's word-level heuristics by operating on lemmas rather than surface forms to account for Turkish agglutination. A pair \((s_1, s_2)\) is retained iff all of the following hold:
        \begin{enumerate}
         \item \(5 \leq \text{len}_{\text{words}}(s_1), \text{len}_{\text{words}}(s_2) \leq 40\);
         \item \(\geq 3\) shared lemmas between \(s_1\) and \(s_2\);
          \item \(\min(\text{len}_{\text{words}}(s_1), \text{len}_{\text{words}}(s_2)) \geq 0.666 \texttimes \max(\text{len}_{\text{words}}(s_1), \text{len}_{\text{words}}(s_2))\);
        \item Bag-of-lemmas edit distance \(\geq 8\).
         \end{enumerate}
      For lemmatization we used spaCy Turkish models \cite{altinok-2023-diverse}, which include a statistical morphosyntactic component (the Morphologizer). After Phase 2, we obtain \(\approx 40{,}000\) candidate pairs.
     \item \textbf{Phase 3: Model-assisted triage} We translate the original MRPC to Turkish and fine-tune a sentence transformer\footnote{\url{sentence-transformers/all-MiniLM-L6-v2}} classifier on the translated set (used only as a weak semantic filter; we do not rely on translated labels for evaluation).
     We score all Phase 2 candidates with this classifier and apply temperature scaling for probability calibration.
     Pairs with high confidence (e.g., \(p \geq 0.9\) paraphrase, \(p \leq 0.1\) non-paraphrase) are provisionally auto-labeled subject to random audits; the remainder proceed to manual annotation.
    \item \textbf{LLM disagreement sampling} To better surface borderline cases, we additionally obtain labels from our LLM Snowflake Arctic on the provisionally selected set. We do not expose model or LLM labels to annotators to avoid anchoring. Instead, we route to humans only those instances where the LLM and the classifier disagree or both exhibit low confidence. In our build, \(\approx 1.5\text{k}\) instances met this criterion and were prioritized for human review.

    \item \textbf{Human annotation and adjudication} Annotators receive only the raw sentence pairs and task guidelines; no pre-labels are shown. We use double annotation with adjudication on a substantial subset to monitor inter-annotator agreement (IAA). Final labels are derived from majority vote with expert adjudication on disagreements.

    \item \textbf{Outcome and class balance} The final Tr-MRPC comprises \(\approx 5.18\text{k}\) sentence pairs, with a class distribution similar to MRPC: \(\approx 1.6\text{k}\) paraphrase (entailment) and \(\approx 3.5\text{k}\) non-paraphrase (non-entailment).
    \end{itemize}

\paragraph{Labeling policy difference from English MRPC}
One issue in constructing a Turkish MRPC-style corpus is that our labeling rules necessarily diverge from English MRPC. In Turkish news, exact sentence-level equivalence across outlets is uncommon—headlines are highly compressed, body sentences vary stylistically, and agglutinative morphology amplifies surface variation. To obtain a usable number of positive pairs without rewarding mere topical overlap, we adopt a relaxed but principled criterion: two sentences are paraphrases if they assert the same main claim, even when one adds non-contradictory detail. We keep the core MRPC exclusions: changes in truth conditions, polarity, modality strength, or key factual quantities remain non-paraphrase. This policy is applied consistently throughout TrMRPC.

We treat the following as paraphrase (label 1) when the main claim is shared.
\begin{itemize}
\item Headline $\leftrightarrow$ detailed sentence: A short headline can pair with a longer sentence if both communicate the same event/claim; extra who/where/how detail is acceptable.\
Ex: ``Paris'te ilk: Sinek alarmı harekete geçirdi." | ``Paris'te sağlık yetkilileri, kaplan sivrisinekleri nedeniyle ilk kez bölgeleri dezenfekte etti." $\rightarrow$ 1
\item Attribution frames: Constructions like ``X farkı", ``X etkisi" are acceptable if the underlying result matches.\
Ex: ``Fenerbahçe'de İsmail Kartal farkı! Sağlam savunma…" | ``Fenerbahçe'de artık savunma daha sağlam." $\rightarrow$ 1
\item Wording/structure changes: Order, phrasing, synonyms, active$\leftrightarrow$passive, or minor detail that does not change truth conditions.\
Ex: ``Kaza sabah saatlerinde oldu." | ``Kaza sabah 8'de meydana geldi." $\rightarrow$ 1
\end{itemize}

We treat the following as non-paraphrase (label 0).
\begin{itemize}
\item Topic overlap only: Same subject, different claims.\
Ex: ``İnsanlar kumruları beslemek için simit atıyor olabilir." | ``Kumru güvercine benzer." $\rightarrow$ 0
\item Mismatched certainty: Hedge vs.\ confirmation changes the claim.\
Ex: ``Şirketin iflas ettiği iddia edildi." | ``Şirketin iflas ettiği doğrulandı." $\rightarrow$ 0
\item Contradictory numbers/dates/names/places: Factual quantity changes the event.\
Ex: ``Bakanlık 500 öğretmen atadı." | ``Bakanlık 600 öğretmen atadı." $\rightarrow$ 0
\item Negation or polarity change.\
Ex: ``Sözleşme imzalandı." | ``Sözleşme imzalanmadı." $\rightarrow$ 0
\item Role swap or incompatible coreference.\
Ex: ``Ali, Veli'yi yendi." | ``Veli, Ali'yi yendi." $\rightarrow$ 0
\item Headline mismatch: Same context, different asserted claims.\
Ex: ``Milli Piyango sonuçları kazananları gösterecek." | ``Sıralı tam liste cnnturk.com'da olacak." $\rightarrow$ 0
\end{itemize}

Annotators apply a substitution test: if one sentence can replace the other in a news story without changing the intended main claim, assign label 1; if the substitution changes the claim, assign label 0. When uncertain, assign label 0. We retain Turkish orthography and do not penalize benign morphological or word-order variation provided the main claim is preserved.

\subsubsection{TrSTS-B} 
The Semantic Textual Similarity Benchmark (STS-B) \cite{stsb} contains sentence pairs drawn from news headlines, and from image/video captioning and NLI sources, each labeled by humans with a similarity score from 1 to 5. A Turkish translation exists \cite{beken-fikri-etal-2021-semantic}, but it was created in 2021 via Google Translate without human quality control. In contrast, our TrSTS-B follows a translate-then-edit pipeline with human verification: we translate English pairs and then apply targeted edits to ensure natural, idiomatic Turkish and cultural appropriateness, followed by annotator spot-checks and format validation.

We choose translation for STS-B because a large fraction of the data—especially captions and simple event descriptions—is effectively language-agnostic (e.g., ``A man is riding a horse"), so faithful translation preserves task semantics. Nonetheless, some phrases are culturally specific or uncommon in Turkish, such as ``mechanical bull," for which a literal form (``mekanik boğa") would be opaque or unnatural. In these cases we substitute culturally appropriate counterparts that preserve the underlying scene/event while keeping difficulty comparable; for example, ``mechanical bull" becomes ``çarpışan araba" (bumper cars). Similarly, where flora, fauna, or place names are unfamiliar in Türkiye (e.g., lemur, Delaware), we replace them with natural Turkish equivalents that maintain semantic roles (e.g., ``lemur" → ``sincap," ``Delaware" → ``İzmir"). We documented these interventions and counted approximately 40 such substitutions. Proper-noun conventions are also normalized; for instance, ``Alexander (the Great)" is rendered as the conventional Turkish ``İskender."

Turkish's lack of articles and different casing/inflectional conventions cause some distinct English sentences to collapse into the same Turkish string. For example, ``The woman is playing the flute." and ``A woman is playing a flute." both translate to ``Bir kadın flüt çalıyor." We remove pairs that collapse to identical sentences because they no longer provide a meaningful similarity gradient. Different English pairs can also map to the same Turkish pair, creating duplicates; we detect and remove these via normalized exact matching and MinHash-based near-duplicate filtering.

Translation was produced with our LLM Snowflake Arctic, under a constrained output format. Human annotators then performed targeted edits and spot-checks, correcting literalisms, proper-noun handling, and culturally odd renderings. We filter outputs that violate the expected format or exhibit hallucinations/unfaithful content; the translation prompt and format constraints are provided in Appendix \ref{sec:stsb-prompt}. Finally, we run a consistency pass to ensure that numeric scales, named entities, and temporal expressions remain aligned across pairs.

Original STS-B has 8.63K English pairs. After cultural substitutions, collapsed-pair removal and deduplication (approximately 3.7K instances), and filtering of malformed/hallucinated outputs and low-quality translations, TrSTS-B contains 3.06K pairs. We withhold a test set due to small size; evaluation is reported on the development split

\subsubsection{TrQQP} The Quora Question Pairs2 (QQP) dataset\footnote{\url{https://data.quora.com/First-Quora-Dataset-Release-Question-Pairs}} is a collection of question pairs obtained from the community-driven question-answering platform, Quora. The task is to determine whether a pair of questions are semantically equivalent, and the labels include ``duplicate" and ``not duplicate."

We construct a Turkish QQP-style resource (TrQQP) by harvesting question pairs from multiple Turkish Q\&A and opinion platforms that span both formal and informal registers. Sources and sizes are summarized in Table \ref{tab:qqp-sources}.

\begin{table}[h]
\captionsetup{width=0.8\textwidth}
\caption{Sizes and source of QQP sentence pairs.}
\label{tab:qqp-sources}
\centering
\begin{tabular}{|l|l|l|}
\hline
Website & Topics & Num of pairs \\ \hline
SETA & politics, economy, society & 58K  \\ 
Yetkin Report & politics, economy, society & 15.5K  \\
Türkiye Raporu & politics, economy, society & 58.1K \\
Ekşi Duyuru & life & 73.6K \\
Sorucevap & life  & 73.7K  \\
KızlarSoruyor & life & 89.6K \\
\hline
\end{tabular}
\end{table}

Data are gathered from several websites with heterogeneous metadata. Some platforms expose explicit ``similar question'' links/tags, enabling guided harvesting of likely duplicates; others lack such signals and require unsupervised retrieval. We unify these sources in a multi-stage pipeline that mixes metadata-driven pairing, text similarity, and model-assisted triage. As with TrMRPC, automatic labels are not shown to annotators. We consulted our LLM, Snowflake Arctic for disagreement sampling and prioritization, not as ground truth. Our phases of compiling QQP are as follows:

\begin{itemize}
    \item \textbf{Phase 1: Crawling and normalization} We crawl publicly accessible question pages and threads across the listed websites, preserving (i) question text, (ii) timestamps, (iii) topic/tags where available, (iv) intra-site similar question edges when present. We apply light normalization only (whitespace/punctuation normalization and Unicode NFC); we do not correct casing, slang, typos, or emojis.

    \item \textbf{Phase 2: Guided pairing via site metadata} For sites that expose similar/related question links, we form positive candidate pairs by materializing each such edge as a pair and adding 1-hop transitive neighbors within thread/topic scopes (while avoiding trivial duplicates by ID). We deduplicate within-site pairs by hashing normalized text and keep provenance for audit.

    \item \textbf{Phase 3: Unguided retrieval for sites without links} For platforms without similar question' tags, we retrieve candidate pairs using a two-pass approach:
     (i) lexical BM25 over a question index to obtain top-$k$ neighbors per query ($k\in[10,50]$ depending on site size), then
    (ii) semantic reranking with a sentence embedding model to select the top-$m$ nearest neighbors.
    We additionally enforce simple length and overlap constraints to avoid trivial matches: 90\% of the tokens per question, at least two shared content lemmas after lemmatization, and a minimum character-level similarity, namely Jaro-Winkler similarity, threshold tuned per site.

    \item \textbf{Phase 4: Negative sampling} To construct non-duplicate pairs with realistic topical overlap, we sample hard negatives by:
   (i) picking high lexical/semantic similarity neighbors below a duplicate threshold,
   (ii) mixing cross-topic pairs within the same site section to prevent overly easy negatives, and (iii) ensuring that negatives do not coincide with any known metadata-linked duplicate.

   \item \textbf{Phase 5: Model-assisted triage (no annotator anchoring)} We train a weak duplicate detector by translating English QQP to Turkish and fine-tuning a Turkish sentence-pair classifier on the translated set. This model is used solely for triage. We calibrate its probabilities with temperature scaling and score all candidate pairs. High-confidence ends ($p \ge 0.9$ duplicate; $p \le 0.1$ non-duplicate) are provisionally auto-labeled and sampled for random audit; the remaining uncertain cases are routed to human annotation.

   \item \textbf{LLM-based disagreement sampling} We obtain independent judgments from Snowflake Arctic on provisionally labeled or uncertain items. Pairs where Arctic and the classifier disagree, or where both are low-confidence, are prioritized for human review. Annotators never see model or LLM labels or rationales.

   \item \textbf{Human annotation} Annotators receive only raw question pairs and task guidelines (duplicate vs.\ not duplicate). We use double annotation with expert adjudication on disagreements and track inter-annotator agreement. We sent \(\approx 35\text{K}\) uncertain/borderline pairs to annotators and audited an additional \(\approx 5\text{K}\) high-confidence ends, totaling \(\approx 40\text{K}\) human-judged items.
   \end{itemize}

Descriptive statistics and construction notes are as follows:
\begin{itemize}
   \item \textbf{Label distribution} The resulting label distribution is skewed toward non-duplicates, similar to English QQP: 219{,}329 not-duplicate (0) and 149{,}465 duplicate (1), i.e., $\approx 63\%$ vs.\ $37\%$.

    \item \textbf{Emoji prevalence and orthography} Emojis are frequent in the informal platforms; we count 15.9K emoji occurrences with 1{,}998 distinct symbols. We retain emojis, slang, abbreviations, and capitalization noise to preserve real-world difficulty (contrast: Quora prohibits emojis).

    \item \textbf{Redundancy and uniqueness} Across all English QQP pairs, there are 980{,}004 unique questions; 240{,}623 (24.55\%) repeat at least once, with a maximum frequency of 391. Counts are heavy-tailed: 739{,}381 appear once, 135{,}202 twice, 47{,}086 thrice, and only a handful exceed 100 repetitions. Across all Turkish TrQQP pairs, there are 360{,}965 unique questions; 49{,}831 (13.80\%) repeat at least once, with a maximum frequency of 470. Compared to English, Turkish is smaller in scale ($\approx$361k vs.\ $\approx$980k unique) and less repetitive overall (13.80\% vs.\ 24.55\%), but exhibits a higher single-question maximum (470 vs.\ 391), indicating a few highly popular questions.

    \item \textbf{De-duplication and split hygiene} We canonicalize questions by Unicode normalization and lightweight token rules to compute equivalence classes of near-duplicates for split hygiene. Train/dev/test splits are created by clustering questions and ensuring clusters do not straddle splits, preventing leakage via repeated questions.
\end{itemize}

\subsubsection{TrMNLI} The Multi-Genre Natural Language Inference Corpus (MNLI) \cite{mnli} is a crowd-sourced dataset consisting of sentence pairs with annotations for textual entailment. The task involves predicting whether a given premise sentence entails, contradicts, or is neutral towards a given hypothesis sentence. The premise sentences are sourced from ten different genres, including transcribed speech, fiction, and government reports. 

MNLI was designed as an ambitious benchmark to capture the full complexity of modern English, extending beyond single-domain or short-caption datasets. It draws from ten distinct genres of written and spoken American English, selected to approximate the linguistic and stylistic diversity of contemporary usage. Five of these genres—fiction, government, slate, telephone, and travel—are included in the training set, while the remaining five—9/11 reports, letters, Oxford University Press non-fiction, verbatim speech, and face-to-face conversation—appear only in evaluation. This split enables systematic assessment of both in-domain (matched) and cross-domain (mismatched) generalization in natural language inference models.

\paragraph{Dataset compilation}
TrMNLI provides a large-scale, genre-balanced Turkish NLI benchmark. Premises are sampled from the BellaTurca collection \cite{bellaturca}, drawing on subcorpora that span distinct communicative styles (e.g., \textit{AkademikDerlem/AcademicCrawl}\footnote{\url{https://huggingface.co/datasets/turkish-nlp-suite/AkademikDerlem}}, \textit{ÖzenliDerlem/CraftedCrwal}\footnote{\url{https://huggingface.co/datasets/turkish-nlp-suite/OzenliDerlem}}, \textit{ForumSohbetleri}\footnote{\url{https://huggingface.co/datasets/turkish-nlp-suite/ForumSohbetleri/ForumChats}}). To mirror MNLI, we curate eight genres and organize them into matched and mismatched splits.

The matched portion (used for training and in-domain evaluation) includes:
\begin{itemize}
  \item \textbf{Folk tales} (MasalMasal / ÖzenliDerlem): narrative and descriptive storytelling;
  \item \textbf{Product complaints} (Serzenişler / ÖzenliDerlem): semi-formal, problem-oriented discourse;
  \item \textbf{Travel blogs} (GeziNotları / ÖzenliDerlem): informal, experiential writing;
  \item \textbf{Forum content} (\textit{ForumSohbetleri}): conversational online communication.
\end{itemize}

The mismatched portion (held out for out-of-domain evaluation) contains four genres with distinct tone and structure (all from \textit{Özenli Derlem}):
\begin{itemize}
  \item \textbf{Movie reviews} (PerdeArkasıYorumlar): subjective and evaluative prose;
  \item \textbf{Fashion magazines} (SüslüTrendler): semi-formal, trend-focused articles;
  \item \textbf{Cultural magazines} (KültürHaritası): formal expository pieces on heritage and events;
  \item \textbf{Viral media content} (ViralMedya): casual, humorous, and stylistically varied online text.
\end{itemize}

While MNLI uses ten genres (five matched, five mismatched), suitable Turkish sources with clear licensing and genre delineation were more limited. Accordingly, TrMNLI comprises eight genres: four matched (used for training and in-domain evaluation) and four mismatched (evaluation only). This design preserves the core matched/mismatched generalization test while maintaining genre diversity across narrative, evaluative, expository, and conversational registers.

\paragraph{Construction methodology}
We adopt an LLM-assisted, human-verified procedure. For each sampled premise sentence, our LLM Snowflake Arctic was prompted to generate three hypotheses—one \emph{entailment}, one \emph{neutral}, and one \emph{contradiction}—under three prompting styles (factuality-focused, linguistic-variation, and free style; prompts in Appendix \ref{sec:mnli-prompt}). Human annotators then conducted a manual validation pass with the following rules: (i) do not edit model text for content; (ii) discard the pair if the hypothesis is ungrammatical or non-native Turkish; (iii) correct the label if it is incorrect; and (iv) optionally fix minor typos/orthography (no semantic edits). Annotators did not see model scores or rationales. We double-annotated a subset with expert adjudication to calibrate quality. To limit artifacts, we balanced lexical overlap across labels, capped overt-negation contradictions, and deduplicated near-duplicates via MinHash and embedding screening. We retain genre tags and provenance (corpus/subcorpus IDs) for audit, and release only premise-hypothesis-label triples with split and genre metadata.

\begin{table}[h]
\captionsetup{width=0.8\textwidth}
\caption{Sizes and sources of TrMNLI sentence pairs across genres.}
\label{tab:trmnli-sources}
\centering
\begin{tabular}{|l|r|r|r|}
\hline
\textbf{Genre} & \textbf{Train} & \textbf{Dev} & \textbf{Test} \\ 
\hline
\multicolumn{4}{|c|}{\textbf{Matched Genres}} \\ \hline
Folk tales & 23.5K & 1.3K & 1.3K \\
Product complaints  & 47.1K  & 2.62K & 2.62K \\
Travel blogs  & 47.1K & 2.62K & 2.62K \\
Forum content  & 47.1K & 2.62K & 2.62K \\ 
\hline
\multicolumn{4}{|c|}{\textbf{Mismatched Genres}} \\ \hline
Movie reviews  & -- & 2.3K & 2.3K \\
Fashion magazines  & -- & 2.3K & 2.3K \\
Cultural magazines  & -- & 2.3K & 2.3K \\
Viral media content  & -- & 2.3K & 2.3K \\ 
\hline
\textbf{TrMNLI overall} & \textbf{165K} & \textbf{18.4K} & \textbf{18.4K} \\
\hline
\end{tabular}
\end{table}

\paragraph{Diversity of linguistic phenomenon}
Although the original MNLI study reports quantitative counts of predefined syntactic or semantic phenomena, we provided diversity in such phenomena during corpus creation. Namely, we prompted Arctic with specific instructions focusing in three fields, factuality, linguistic variations and free style. Appendix \ref{sec:mnli-prompt} presents the prompts used for this purpose. We also provide a few illustrative examples below. The examples below illustrate typical sentence-pair relations in Turkish along two primary dimensions: (i) factuality and (ii) linguistic variation.

Many instances require evaluating whether the hypothesis is grounded in objective information expressed in the premise.  
\begin{itemize}
   \item \textbf{Entailment:}  
   \textit{Premise:} Türkiye'nin başkenti Ankara'dır.  
   \textit{Hypothesis:} Ankara Türkiye'nin başkentidir.  
   (\emph{The hypothesis repeats an objectively true fact stated in the premise.})
   \item \textbf{Contradiction:}  
   \textit{Premise:} Türkiye'nin başkenti Ankara'dır.  
   \textit{Hypothesis:} Türkiye'nin başkenti İstanbul'dur.  
   (\emph{The hypothesis contradicts an objective fact from the premise.})
   \item \textbf{Neutral:}  
   \textit{Premise:} Türkiye'nin başkenti Ankara'dır.  
   \textit{Hypothesis:} Ankara İç Anadolu Bölgesi'nde yer alır.  
   (\emph{The hypothesis introduces new factual information that is not entailed by the premise.})
\end{itemize}

A variety of transformations highlighting sentence-level lexical and structural diversity in Turkish included in the dataset.
\begin{itemize}
   \item \textbf{Paraphrasing:}  
   \textit{Premise:} Toplantı iptal edildi.  
   \textit{Hypothesis:} Toplantıyı yapmamaya karar verdiler.  
   (\emph{Same meaning expressed with different lexical choices.})
   \item \textbf{Summarization:}  
   \textit{Premise:} Konferans, birçok uzmanın katıldığı uzun bir tartışmayla sonuçlandı.  
   \textit{Hypothesis:} Konferans tartışmalarla bitti.  
   (\emph{Condensed version preserving the main meaning.})
   \item \textbf{Lexical Transformation:}  
   \textit{Premise:} Çocuk hızlı koştu.  
   \textit{Hypothesis:} Çocuk seri adımlarla ilerledi.  
   (\emph{Synonym and lexical substitution.})
   \item \textbf{Syntactic Restructuring:}  
   \textit{Premise:} Doktor hastayı muayene etti.  
   \textit{Hypothesis:} Hasta doktor tarafından muayene edildi.  
   (\emph{Active-passive alternation preserving entailment.})
   \item \textbf{Focus Shifting:}  
   \textit{Premise:} Öğrenci öğretmene bir mektup verdi.  
   \textit{Hypothesis:} Öğretmen öğrenciden bir mektup aldı.  
   (\emph{Same event with shifted focus.})
   \item \textbf{Negation:}  
   \textit{Premise:} Kadın tiyatroya gitti.  
   \textit{Hypothesis:} Kadın tiyatroya gitmedi.  
   (\emph{Contradiction created by negation.})
   \item \textbf{Addition of Unrelated Details:}  
   \textit{Premise:} Adam parka yürüdü.  
   \textit{Hypothesis:} Adam parka yürüyüp arkadaşlarıyla kahve içti.  
   (\emph{Neutral example by adding unrelated new information.})
\end{itemize}

Overall, these patterns demonstrate that TrMNLI contains a wide range of natural linguistic phenomena beyond simple lexical overlap, including factual reasoning, rephrasing, and structural variation, reflecting the expressive morphology and flexible word order characteristic of Turkish.

\subsubsection{TrQNLI} The Stanford Question Answering Dataset (QNLI) \cite{squad} is a question-answering dataset comprising question-paragraph pairs. Each paragraph is extracted from Wikipedia and contains the answer to the corresponding question, which is generated by an annotator. The benchmark authors convert the task into sentence pair classification by creating pairs between each question and each sentence in the corresponding context, filtering out pairs with low lexical overlap. The objective is to determine whether the context sentence contains the answer to the question. 

For TrQNLI, we construct a native Turkish counterpart using Turkish Wikipedia. We sample approximately 24K paragraphs and prompt our LLM Snowflake Arctic to generate question-answer pairs, requesting exactly 10 question-answer-answer-index triplets per paragraph. While we do not require span extraction at evaluation time, we still guide the LLM to respect Turkish morphology when proposing answers. In particular, the prompt requests both lemma and surface forms where relevant and enforces case morphology when it is semantically required by the question (e.g., a genitive question such as ``Kimin …?" expects a genitive-marked answer). The full prompt appears in Appendix \ref{sec:qnli-prompt}.

We obtain about 240K raw triplets and apply automatic filtering to remove malformed outputs, format violations, and likely hallucinations, resulting in roughly 150K remaining instances. Human annotators then review a stratified subset and discard low-quality items (approximately 10K), yielding around 140K validated question-answer-context triplets.

To convert to the QNLI format, we split each context into sentences, identify the answer-bearing sentence for each triplet, and form question-sentence pairs labeled as entails (the sentence contains the answer) or not. For each positive pair we sample a matched negative from the same paragraph to control topical overlap, producing a 1:1 class balance. Train/dev/test splits are stratified to maintain label balance and to avoid paragraph leakage across splits.

\subsubsection{TrRTE}
The Recognizing Textual Entailment (RTE) datasets are derived from a series of annual entailment challenges. The benchmark combines RTE1 \cite{rte1}, RTE2 \cite{rte2}, RTE3 \cite{rte3}, and RTE5 \cite{rte5}, with examples built from news and Wikipedia. For consistency, all datasets are converted to a two-class setting by collapsing neutral and contradiction into a single ``not entailment'' class.

RTE was compiled within the PASCAL challenges to provide an application-driven benchmark for inference. Curators sourced short texts (premises, $P$) from realistic pipelines—IR, IE, QA, MT—primarily from newswire and web documents. For each text, trained annotators authored or edited a single-sentence hypothesis ($H$) that was either licensed by the text (entailed) or deliberately plausible yet unsupported/contradicted (not entailment), under guidelines targeting diverse phenomena (negation, quantification, temporal relations, coreference, lexical inference). Candidate pairs were double-annotated and adjudicated to produce gold binary labels. The datasets are small but carefully curated (e.g., RTE-1: 567 dev, 800 test), with short, self-contained texts and hypotheses, balanced across application categories and labels, and standardized as $(P,H,\text{label})$ pairs.

We construct a Turkish RTE corpus in a similar spirit by sampling short, self-contained premises from licensed newspapers and academic texts, and instructing native-speaker annotators to generate single-sentence hypotheses under three complementary styles. 
\begin{itemize}
\item (i) \emph{Linguistic}: annotators target a specific phenomenon (e.g., negation, quantifiers, comparatives, modality, coreference, word order/topicalization, agreement/case, derivational/inflectional morphology) and author minimal, fluent hypotheses that isolate that phenomenon while preserving truth conditions. 
\item (ii) \emph{Free}: open-ended, natural paraphrasing that stays on topic while varying lexical choice and syntax to reduce superficial overlap artifacts. 
\item (iii) \emph{Factuality}: minimal edits to factual attributes (entities, dates, locations, quantities, roles) to yield clear entailment/not-entailment cases. 
For each premise, annotators draft candidates for both labels (entailed, not entailment) using any styles, with the constraint that at least one pair per premise uses the linguistic style to ensure systematic coverage. A separate validation pass with different annotators labels the pairs; disagreements are adjudicated. We maintain a 50/50 label balance within each split (train/dev/test).
\end{itemize}

To promote consistency, we provide concise labeling guidelines: an entailment hypothesis must follow from the premise under common background knowledge without adding new specific details; a not-entailment hypothesis is either plausible but unwarranted (neutral) or incompatible (contradiction). Turkish-specific cautions include using both clausal (değil, -ma/-me) and morphological (-maz, yok) negation without over-reliance on any single cue; careful treatment of quantifiers and determiners (tüm/bazı/hiçbir) and their scopes; correct agreement and case marking; attention to evidentiality and modality (-miş vs.\ -di; olabilir, muhtemelen), especially for neutrality; and leveraging SOV flexibility and focus particles (bile, sadece) without altering truth conditions.

We control artifacts by (i) balancing lexical overlap across labels, (ii) limiting the fraction of pairs where overt negation alone flips the label, (iii) diversifying contradiction mechanisms (number, date, role, entity swaps; world knowledge), (iv) prohibiting verbatim copying of premises, and (v) deduplicating near-duplicates via MinHash and embedding screening. Quality assurance includes two independent labels per pair, majority-vote gold with a ``no-consensus'' bucket excluded from evaluation for auditability.

\subsubsection{Absence of TrWNLI: Why WNLI Does Not Transfer to Turkish}
\label{sec:wnli}

We argue that a Turkish adaptation of WNLI (Winograd-style NLI) is linguistically ill-posed and largely redundant. The core ambiguity exploited by WNLI in English—reference resolution for third-person pronouns under minimal lexical cues—does not naturally arise in Turkish due to (i) pervasive subject and object omission (pro-drop/zero anaphora), (ii) rich case and agreement morphology, (iii) productive voice and reflexive morphology, and (iv) flexible word order that encodes information structure. Literal translations of WNLI items either (a) suppress the relevant ambiguity by distributing information across morphology and structure, or (b) yield stilted, non-native uses of overt pronouns, thereby testing translation artifacts rather than reasoning. In other words, the construct measured by WNLI in English is anchored in properties of the English pronominal system (limited morphology, obligatory overt pronouns, and fixed SVO word order). Turkish departs from these properties along several dimensions: pronouns are frequently omitted when recoverable, argument roles are morphologically marked, and discourse prominence is signaled by word order and particles. As a result, when the same scenarios are expressed idiomatically in Turkish, the intended ambiguity is either eliminated by overt morpho-syntactic cues or becomes unnatural to maintain. This misalignment undermines both the face validity (native acceptability) and the construct validity (is the benchmark still testing pronoun resolution?) of a direct TrWNLI port.

WNLI descends from Winograd Schemas: minimally contrasted sentence pairs in which a pronoun (\emph{he/she/it/they}) admits two plausible antecedents, and where a subtle lexical or world-knowledge cue determines the correct resolution. In English, the pronominal system provides limited morpho-syntactic disambiguation: third-person pronouns lack rich case morphology (beyond nominative/accusative), have limited agreement features, and rely heavily on syntactic position and semantics for interpretation. Consequently, the benchmark meaningfully probes pragmatic reasoning and commonsense under tight surface controls. Crucially, the English design assumes that the pronoun must be overt and that replacing it with a full noun phrase would change the distribution of cues the model sees, weakening the diagnostic. The minimality constraints (few function word changes, no explicit re-mention of antecedents) are tailored for English. When transported to languages where such constraints are not natural—because pronouns are optional or because case and agreement overtly encode roles—the same minimal edits no longer isolate pragmatic knowledge; instead, they repackage or erase it via grammar.

In Turkish, the same functional space is occupied and often obviated by morphological and structural mechanisms. First, pro-drop and zero anaphora are pervasive: null subjects and (to a lesser extent) null objects are licensed when recoverable from context. Overt third-person \emph{o} is optional and stylistically marked; native renderings prefer dropping it. This removes the surface pronoun token that WNLI manipulates and shifts interpretation to verbal morphology and discourse coherence. Practically, contexts designed to hinge on ``it" or ``she" often have no overt pronoun at all in Turkish. The reference is resolved through the agreement morphology on the verb and through discourse structure. Forcing an overt \emph{o} to mimic English typically sounds marked or literary and introduces unnaturalness that contaminates the evaluation signal.

Second, case and postpositions play a central role. Productive case marking—accusative (\emph{-\,(y)I}), dative (\emph{-\,(y)A}), locative (\emph{-DA}), ablative (\emph{-DAn})—and postpositions encode grammatical/semantic roles, often forcing a unique reading that English leaves ambiguous. Ambiguity that hinges on bare ``it" in English is resolved by \emph{havuçta} (LOC) vs.\ \emph{havucun} (GEN) vs.\ \emph{havucu} (ACC), etc. Case morphology does not merely add redundancy; in many minimal pairs, it deterministically signals affectedness, definiteness/specificity, and thematic roles. Thus, tiny morphological choices (e.g., locative vs.\ genitive) collapse the ambiguity without appeal to world knowledge. This shifts the task from pragmatic inference to morphological parsing.

Third, agreement, voice, and reflexives further constrain interpretation. Verbal inflection marks person/number and often identifies the elided subject. Voice alternations (passive, causative, reflexive, reciprocal) restructure argument prominence, while reflexive \emph{kendi/kendisi} and possessive suffixes (\emph{-sI}) disambiguate coreference that English encodes with bare pronouns. For example, reflexive morphology encodes self-directed actions that English indicates with ``himself/herself" or even bare ``him/her" under certain constraints. In Turkish, choosing between \emph{kendi(ni)} and \emph{onu} is a categorical signal of binding, not a subtle cue. Similarly, passive voice can suppress the agent, altering salience and reference in ways that make Winograd-style ambiguity poorly formed.

Fourth, flexible word order and information structure allow speakers to signal intended referents via placement and particles (e.g., \emph{sadece}, \emph{bile}) rather than through pronominal choice. Native phrasing tends to encode the disambiguating cue structurally. Information-structural choices are not ornamental: fronting a constituent or marking focus can make one antecedent prominent and the other backgrounded, which in turn guides pronoun resolution. English relies more on prosody and fixed order; Turkish openly uses syntactic flexibility. Consequently, a literal Winograd item translated word-for-word often becomes trivial for native readers, because the ``right" antecedent is foregrounded by structure.

Consider two canonical WNLI-style examples and their Turkish realizations. Carrot/pin (English): \emph{I stuck a pin through a carrot. When I pulled the pin out, it had a hole.} Hypothesis: \emph{The carrot had a hole.} Literal Turkish: \emph{Bir havucun içinden iğne geçirdim. İğneyi çıkardığımda bir delik vardı. / Havucun bir deliği vardı.} Natural Turkish: \emph{Bir havucun içinden iğne geçirdim. İğneyi çıkardığımda \textbf{havuçta} bir delik vardı. / \textbf{Havuçta} bir delik vardı.} In English, the pronoun ``it" is ambiguous between \emph{pin} and \emph{carrot}; world knowledge resolves it in favor of \emph{carrot}. In Turkish, the natural formulation avoids an overt pronoun entirely and uses the locative \emph{havuçta}, which encodes the affected location. The morphological marking makes the intended referent explicit, so the Winograd-style ambiguity evaporates without invoking commonsense.

Tatyana/mother (English): \emph{When Tatyana reached the cabin, her mother was sleeping. She was careful not to disturb her, undressing and climbing back into her berth.} Literal Turkish: \emph{Tatyana kulübeye vardığında annesi uyuyordu. Onu rahatsız etmemeye dikkat ederek soyunup ranzasına geri döndü.} Natural Turkish: \emph{Tatyana kulübeye vardığında annesi uyuyordu. \textbf{Annesi} onu rahatsız etmemeye dikkat ederek soyunup ranzasına geri döndü. / \textbf{Annesi} \textbf{Tatyana'yı} rahatsız etmemeye dikkat ederek soyunup ranzasına geri döndü.} The English item leverages two female referents and ambiguous ``she/her." Turkish resolves this by reintroducing the subject noun phrase (\emph{annesi}) and case-marking the object (\emph{Tatyana'yı}). These choices are natural and preferred; keeping bare pronouns (\emph{o}/\emph{onu}) is either marked or unclear. The resulting sentences no longer rely on delicate pronoun cues; reference is disambiguated morpho-syntactically.

Attempting to port WNLI to TrWNLI raises methodological issues. First, translation artifacts are unavoidable if one tries to preserve English-like ambiguity. Translators must either overuse overt pronouns or suppress normal case marking, both of which produce sentences that native speakers judge unnatural. Systems trained on such data risk learning to prefer translations rather than mastering Turkish discourse. Second, there is label instability under minimal edits. Edits that are pragmatically inert in English—such as introducing a locative or switching bare to accusative objects—can deterministically fix reference in Turkish and flip labels. This violates the minimal-pair design principle and makes evaluation brittle with respect to morpho-syntactic variation. Third, there are construct validity concerns: when disambiguation is carried by obligatory or strongly preferred morphology, the benchmark ceases to isolate commonsense or pragmatic reasoning and instead measures whether models detect overt markers. This conflates the target competence and can inflate scores through shallow cues. Finally, TrWNLI would be redundant with Turkish-native phenomena that we already evaluate elsewhere (e.g., in our linguistic-style tasks): zero anaphora resolution, reflexive vs.\ non-reflexive alternations (\emph{kendi/kendisi} vs.\ \emph{onu}), possessive disambiguation (\emph{annesi} vs.\ \emph{annesini}), case-driven specificity (ACC vs.\ bare), and voice alternations (passive/causative) that reshape salience.

We do not provide a Turkish WNLI (TrWNLI) split in TRGLUE. As argued above, Winograd-style pronoun disambiguation does not transfer to Turkish due to pro-drop, rich case/agreement morphology, and flexible information structure. Direct translations either eliminate the ambiguity or yield translationese, undermining construct validity. Accordingly, TrWNLI is omitted from TRGLUE. For transparency and comparability, we include a short note in the benchmark documentation and tables clarifying that omission is due to linguistic inapplicability rather than resource constraints.

Although we omit TrWNLI, we still assess reference-related reasoning in Turkish via more appropriate tasks (e.g., anaphora and morphology-sensitive inference) described elsewhere in this paper. 

A direct TrWNLI port misconstrues the locus of ambiguity in Turkish and risks measuring conformity to Anglophone pronominal patterns rather than genuine inference. A morphology- and discourse-aware evaluation regime—without a WNLI-style split—offers a valid and linguistically grounded path for Turkish NLI. Accordingly, we omit TrWNLI from TRGLUE and document the linguistic rationale to guide future cross-lingual benchmark design.

\subsection{Comparison of corpus statistics}
English and Turkish differ substantially in linguistic structure. Turkish is agglutinative with rich inflectional morphology and relatively free word order \cite{Oflazer1994}. For instance, a single Turkish token such as \emph{gideceksin} corresponds to multiple English words (``you will go"). These properties affect parsing and tokenization: in ``Kediyi köpek kovaladı," the dependencies \texttt{(Kediyi obj kovaladı, köpek nsubj kovaladı, kovaladı root ROOT)} differ from ``The dog chased the cat" \texttt{(The det dog, dog nsubj chased, chased root ROOT, the det cat, cat obj chased)}. Case marking and flexible constituent order yield different head-dependent configurations, often reducing orthographic token counts in Turkish while increasing the diversity of surface forms per lemma. Consistent with this, Turkish translations may contain fewer words on average, as reflected in the ``Avg. words" column of Table \ref{tab:vocab-comp}.

Table \ref{tab:vocab-comp} presents vocabulary statistics for each dataset in both English and Turkish. For each dataset, we computed the total word count, unique word count (vocabulary), and distinct lemma count. While lemma and suffix counts are less meaningful for English, they are significant for Turkish due to its morphosyntactic complexity. These statistics were computed using spaCy Turkish models, recognized for their industrial-grade accuracy and speed. The computation scripts are available in the TrGLUE GitHub repository.

\begin{table}[ht]
    \centering
    \setlength{\tabcolsep}{3pt}
    \caption{Comparison of vocabulary statistics between the English and Turkish GLUE datasets for each individual dataset. We note that original QQP and MNLI are twice the size of our datasets, STS-B is almost 3 times size.}
    \label{tab:vocab-comp}
    \begin{tabular}{lcccccccc}
        \toprule
        \multirow{2}{*}{\textbf{Dataset}} & \multicolumn{4}{c}{\textbf{English}} & \multicolumn{4}{c}{\textbf{Turkish}} \\
        \cmidrule(lr){2-5} \cmidrule(lr){6-9}
        & \textbf{Tot wrds} & \textbf{Uniq wrds} & \textbf{Lemmas} & \textbf{Avg words} & \textbf{Tot wrds} & \textbf{Uniq wrds} & \textbf{Lemmas} & \textbf{Avg words} \\
        \midrule
        CoLA  & 83.6K  & 6.3K  & 4.3K & 7.81  & 123.9K & 17.6K & 12.1K & 12.48 \\
        SST-2 & 646K & 16.2K & 13.5K & 9.22  & 3.6M & 297K & 233K &  47.3 \\
        MRPC  & 226.6K & 18.5K & 15.2K & 19.5 & 170K & 37.5K & 16.1K  & 16.3 \\
        STS-B & 148K &  6.4K & 5K & 9.26  & 37.8K & 3.8K & 1.9K & 6.17 \\
        QQP  & 17.9M & 175K & 157K & 11.25 & 5.5M & 112K & 42K & 7.52 \\
        MNLI & 12M & 40K & 25K & 13.85 & 4.2M  & 273.8K & 124K &  21.13 \\
        QNLI & 4.3M & 89K & 78K & 18.5 & 4.2M & 211K & 110.4K & 15.9 \\
        RTE   & 300K & 26.5K & 21.8K & 25.95 & 141K & 35.8K  &13.9K  & 12.23  \\
        \bottomrule
    \end{tabular}
\end{table}

The figures in Table \ref{tab:vocab-comp} reflect systematic cross-lingual effects rather than preprocessing artifacts. Turkish splits typically have fewer words per instance but many more unique surface forms relative to lemma counts, consistent with agglutinative morphology and productive suffixation. This yields higher type-token ratios on the Turkish side (e.g., SST-2, MNLI) even when total word counts are comparable or smaller (e.g., MRPC, RTE). Pairwise tasks (QQP, MRPC, STS-B) compress more aggressively in Turkish, reducing average length while preserving core semantics. These properties informed our design choices: lemma-aware preprocessing for retrieval and filtering, relaxed equivalence for TrMRPC to handle headline-body asymmetries, and sentence-level segmentation for TrQNLI with attention to morphology. Practically, subword tokenization will fragment Turkish words more, potentially increasing sequence lengths despite fewer orthographic tokens; morphologically informed tokenizers and normalization choices therefore have outsized impact. We release scripts and configuration details (model versions, seeds, and normalization settings) to facilitate exact replication and ablation.

\subsection{Linguistic diversity in TrGLUE}
Turkish features agglutinative morphology, relatively free word order, and pervasive pro-drop, posing modeling challenges that differ from English GLUE. We quantify TrGLUE's coverage of these core phenomena and report their per-task distributions.

All measurements below are derived with spaCy's Turkish pipeline, which jointly predicts POS, morphological features, NER, and dependency parses \cite{Honnibal_spaCy_Industrial-strength_Natural_2020}. Because spaCy's components are coupled (morphological analyses inform POS and vice versa, and both condition the dependency parse), our counts reflect integrated morpho-syntactic analyses rather than isolated taggers. Where relevant, we exclude punctuation and follow standard UD conventions for token-head relations and feature inventories.

\subsubsection{Morphological richness and coverage}
We measure (i) morphemes-per-token, (ii) distinct inflectional feature bundles, and (iii) derivational suffix productivity using spaCy-based tagging and the Zeyrek finite-state analyzer\footnote{\url{https://github.com/obulat/zeyrek}}. TrGLUE contains on average 2.25 morphemes/token (median: 2; 95th pct: 5; 99th pct: 8), with 1,095 distinct inflectional feature bundles (POS×Case×Number×Person×Tense×Aspect×Mood×Polarity×Evidentiality). Verbal negation via -ma/-me appears in 11.75\% of finite verbs, evidential/mirative marking (-miş) in 9.21\%, and derivational morphology is frequent (e.g., -lI, -sIz, -CI, -lIk).

Across datasets, the morphemes-per-token histograms exhibit a consistent core with a heavy tail characteristic of Turkish agglutination (Figure \ref{fig:histo-all}). The overall median is 2 and the mean is 2.25, with the 95th and 99th percentiles at 5 and 8, respectively. Most tokens fall in the 1-4 range, while a meaningful tail (≥5 morphemes) reflects stacked case, agreement, derivation, and polarity/evidential markers. For readability, the rightmost mass is aggregated into a ``10+'' bin, and we annotate the median and p95 to emphasize the tail. Per-dataset shapes broadly mirror the overall distribution, with NLI/QA skews slightly heavier (more verbs and possessed/case-marked nominals) and sentiment lighter. These histograms contextualize token-level complexity behind length expansion and help explain performance sensitivity in morphologically rich segments.

\subsubsection{Syntax and word order variation}
Across TrGLUE, Turkish displays the expected SOV-dominant profile with limited but measurable flexibility: among clauses with overt subjects and objects, non-canonical permutations (OSV/OVS/VSO/VOS) account for 3.07\%, consistent with topicalization and information-structure-driven reordering under a generally head-final preference. The pooled average dependency distance is 3.26 tokens (non-punctuation arcs), slightly higher than typical English estimates and compatible with a largely right-branching written register in a morphologically rich language with postverbal clausal complements and freer constituent placement. Subjects are omitted in 73.64\% of finite clauses, underscoring robust pro-drop even in formal task data; this rate remains high under conservative coordination-based subject recovery, indicating that null subjects are pervasive rather than a parsing artifact. Overall, TrGLUE preserves core typological properties of Turkish—SOV alignment, flexible alternations, and extensive subject drop—with dependency lengths in a reasonable range for mixed-genre corpora.

For English GLUE, we calculated average dependency distance as 2.5 tokens; non-canonical S/O/V orders occur in 1.1\% of clauses with overt subjects and objects; and subject omission is 0.5\% of finite clauses. Relative to this baseline, Turkish TrGLUE shows longer dependencies (3.26 vs. 2.5), a higher rate of non-canonical orders (3.07\% vs. 1.1\%), and dramatically more subject omission (73.64\% vs. 0.5\%), highlighting English's rigid SVO, non-pro-drop profile versus Turkish's flexible word order and robust pro-drop.

We compute average dependency distance as a pooled mean over non-punctuation dependency arcs: for each token-head pair within a sentence, distance is the absolute difference in token indices, and the corpus average is the sum of distances divided by the number of counted arcs. Clause word order is estimated from dependency parses by locating, for each finite verb (VerbForm=Fin), the nearest overt subject and object dependents within the same sentence. We consider subjects labeled nsubj/csubj (including passive variants) and, for objects, obj/iobj. Orders are assigned by the linear positions of S, O, and V; non-canonical rates are computed among clauses where both S and O are present.

\subsubsection{Named entities}
Using a unified NER pipeline and tokenization regime, TrGLUE exhibits a high overall density of named entities. Aggregating across all subsets, we counted 35,635,200 tokens and 3,278,439 entity tokens, yielding an average of 0.092 entity tokens per token (9.2\%). In broad terms, this pooled rate indicates that entity-bearing content is frequent across TrGLUE—consistent with the prevalence of newsy or informational text in several tasks—so models repeatedly encounter person, organization, location, and miscellaneous entities, making NER coverage and entity-aware tokenization consequential for downstream performance. Because Turkish morphology packs case and agreement onto proper nouns, entity recognition and normalization are sensitive to segmentation choices; nevertheless, the aggregate density suggests ample signal for evaluating how models handle named entities. Totals are computed by summing non-overlapping entity spans and token counts over all instances, using the same lowercasing, no-punctuation counting, and digit-collapsing conventions as elsewhere for consistency

\begin{figure}[h!]
\centering
% Top row
\subfloat[Pooled average dependency distance by task. Bars show the mean token-head distance over all dependency edges within each dataset (excluding punctuation), with the pooled ``Overall'' summary on the right; higher values indicate looser, more right-branching structure.]{
\includegraphics[width=0.47\textwidth]{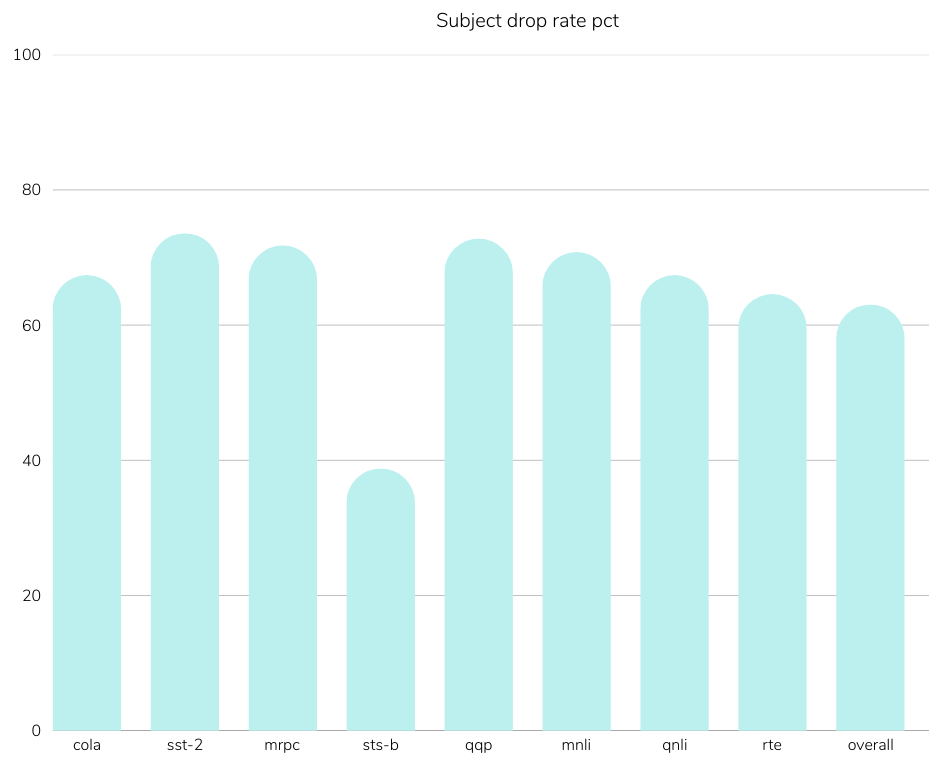}
}
\hfill
\subfloat[Subject-drop rate (\%) in finite clauses. Bars report the proportion of finite clauses lacking an overt subject (null subject), highlighting pervasive pro-drop across tasks; ``Overall'' gives the pooled estimate.]{
\includegraphics[width=0.47\textwidth]{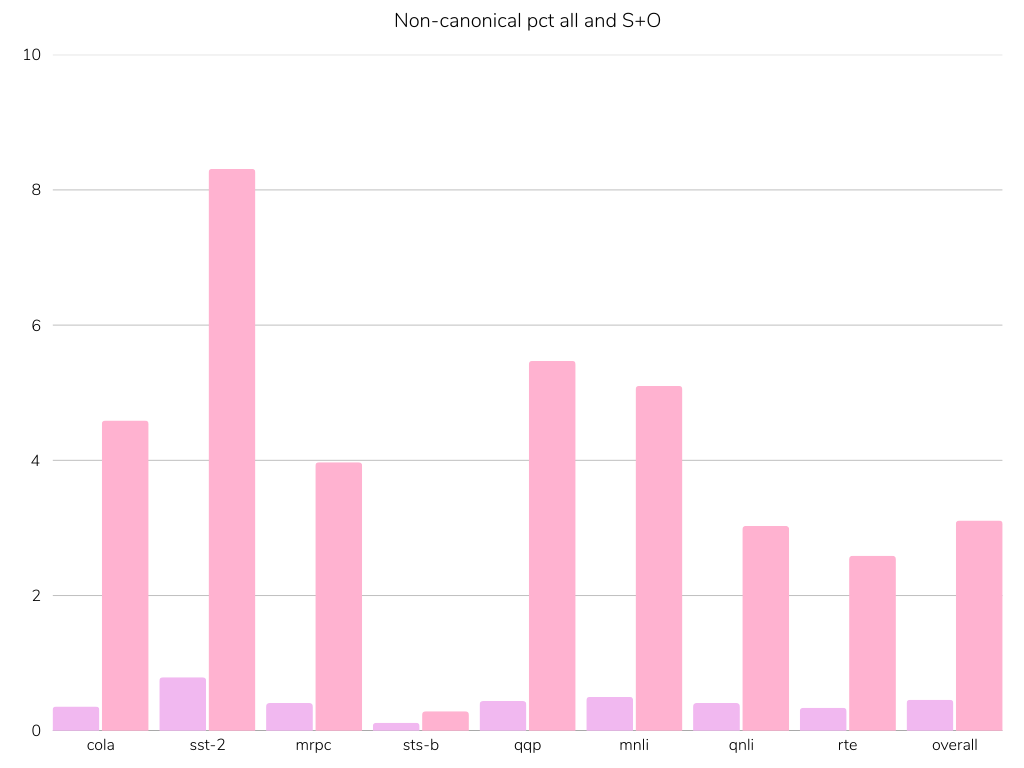}
}

\vspace{0.75em}

% Bottom row
\subfloat[Non-canonical word order rates (\%) by task. ``All'' = share of OSV/OVS/VSO/VOS among all finite clauses; ``S+O'' = restricted to clauses where both subject and object are overt.]{
\includegraphics[width=0.47\textwidth]{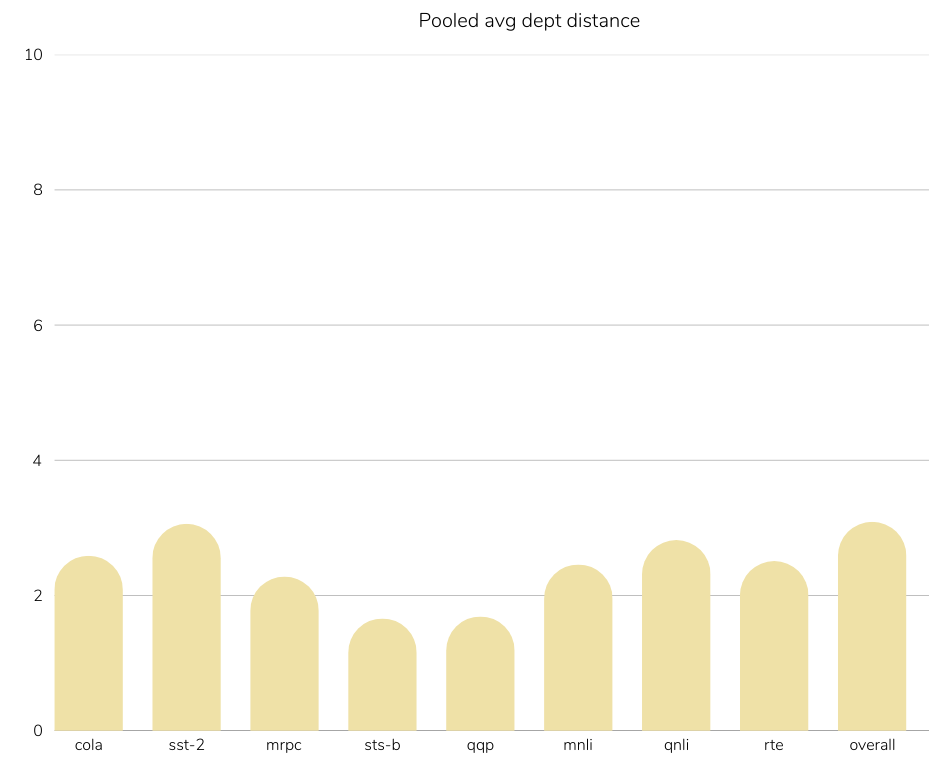}
}
\hfill
\subfloat[Entity density by subset and overall. Each panel reports the proportion of tokens that are part of named entities (entity tokens/tokens, \%)]{\includegraphics[width=0.47\textwidth]{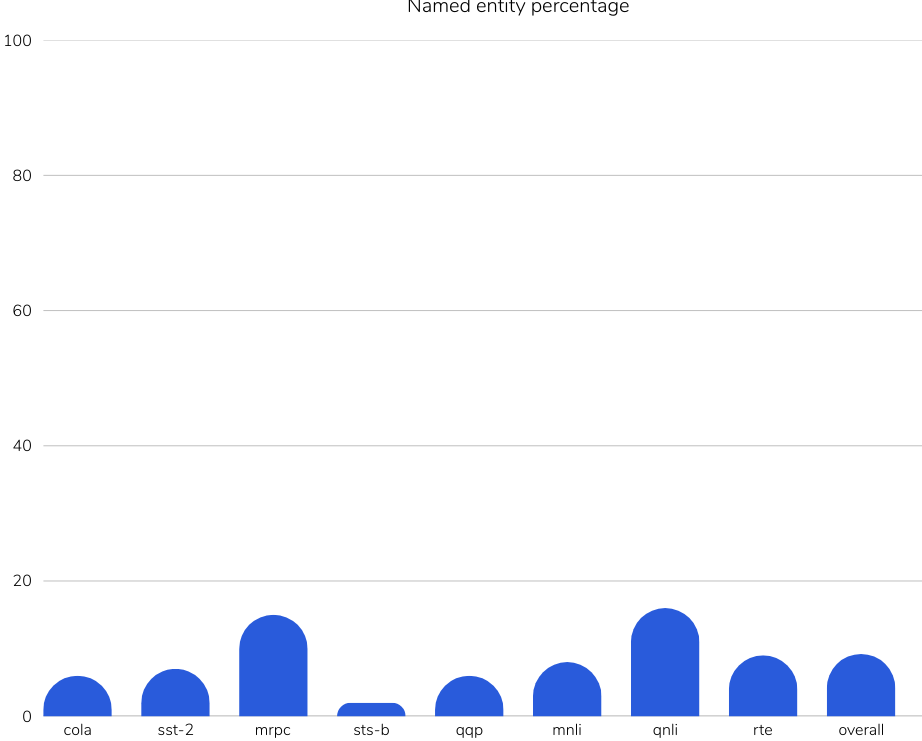}}

\caption{Cross-task syntactic and named entity profiles for TrGLUE. Subfigures (a)-(c) report (a) average dependency distance (tokens), (b) subject-drop rate (\%), and (c) non-canonical word order rates (\%), with ``Overall'' computed over all finite clauses and ``S+O'' restricted to clauses where both subject and object are overt. Together they illustrate SOV-dominant syntax with pervasive pro-drop, modest but consistent word-order flexibility, and task-specific variation.}
\label{fig:dep-stats-per-subset}
\end{figure}

\subsubsection{Task-relevant phenomena}
Across subsets, the morphemes-per-token profiles share a common Turkish core but differ in tail weight and inventory breadth in ways that track genre and task design. The pooled distribution is centered at 2 morphemes with a heavy tail (p95=5, p99=8), yet means span from lighter CoLA (2.14) and QQP/SST-2/MRPC/QNLI (2.37-2.42) to heavier MNLI/STS-B/RTE (2.50-2.58). Heavier subsets exhibit more mass in 3-5 and occasional ≥6 morpheme bins, driven by stacked case/possessive/agreement on nominals and tense-aspect-mood-polarity/evidential layering on verbs; lighter subsets concentrate in 1-3, reflecting shorter, template-like prompts and fewer derivational chains. Inflectional feature diversity follows suit: bundle counts are smallest in CoLA (340) and largest in the pooled ``Overall'' set (1,095), with intermediate diversity for MNLI/RTE (745-812) and broad coverage for QQP/QNLI/MRPC/SST-2 (902-1,072), indicating that genre mixture and corpus breadth—not just size—expand the space of inflectional combinations.

\begin{figure}[t]
\centering

% Row 1
\subfloat[CoLA]{\includegraphics[width=0.3\textwidth]{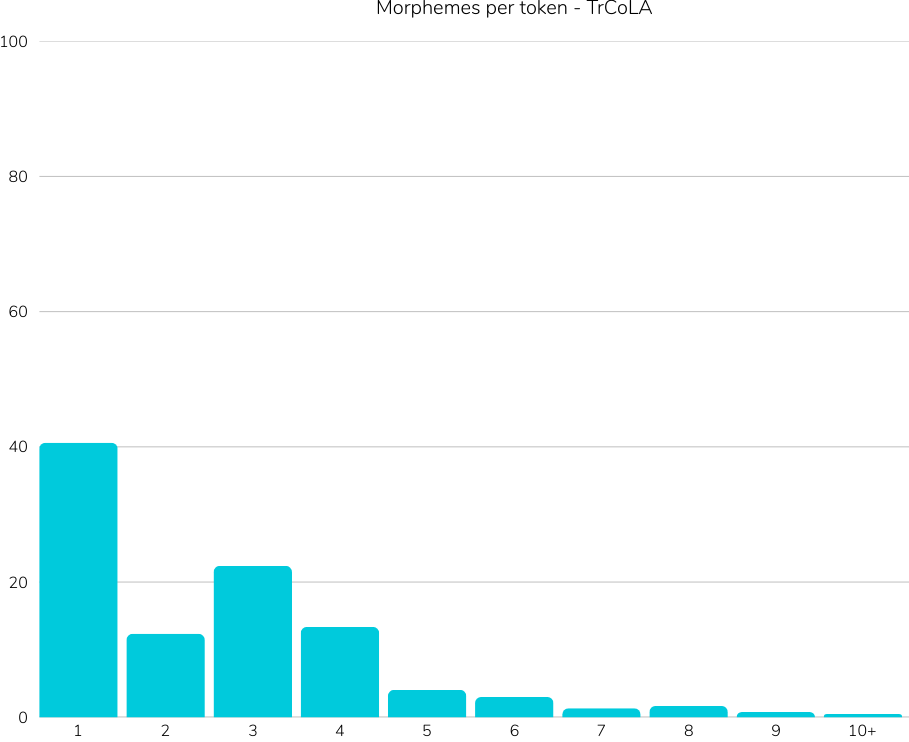}}\hfill
\subfloat[MNLI]{\includegraphics[width=0.3\textwidth]{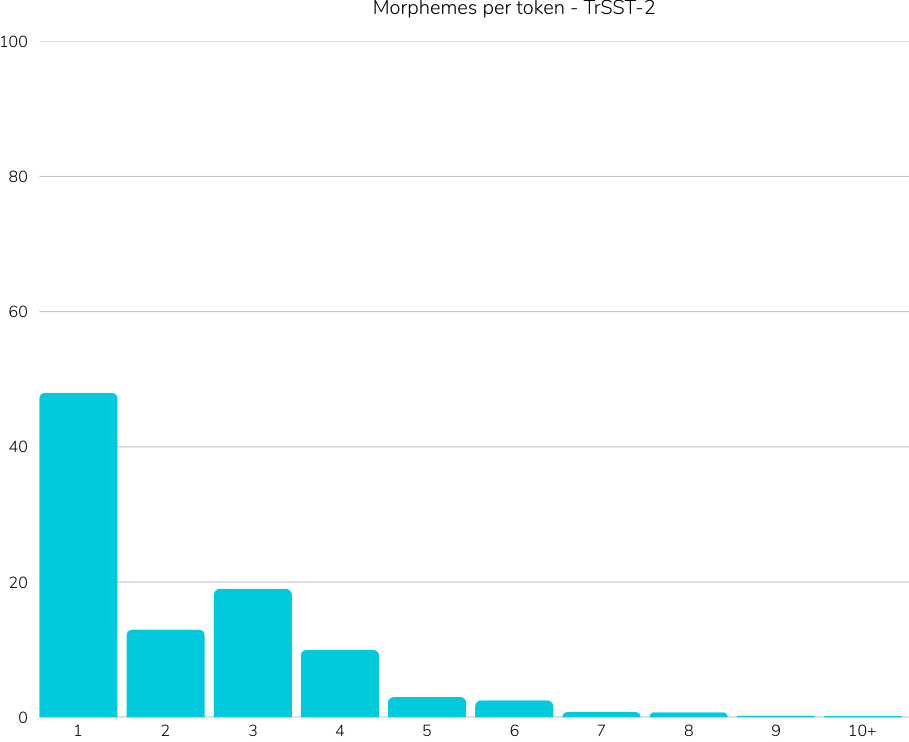}}\hfill
\subfloat[MRPC]{\includegraphics[width=0.3\textwidth]{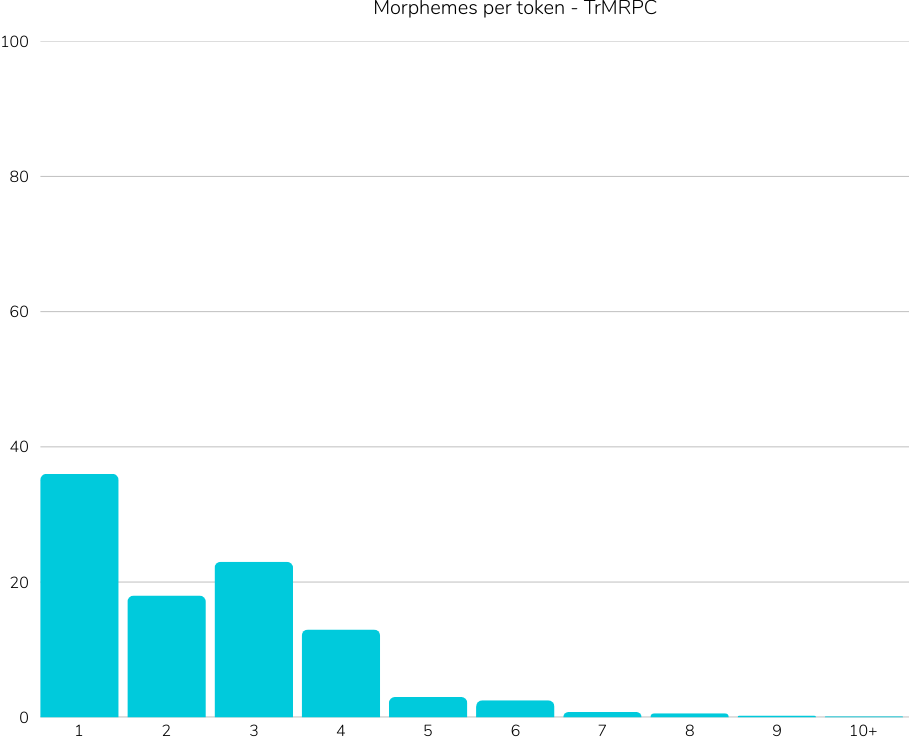}}

\vspace{0.6em}

% Row 2
\subfloat[QNLI]{\includegraphics[width=0.3\textwidth]{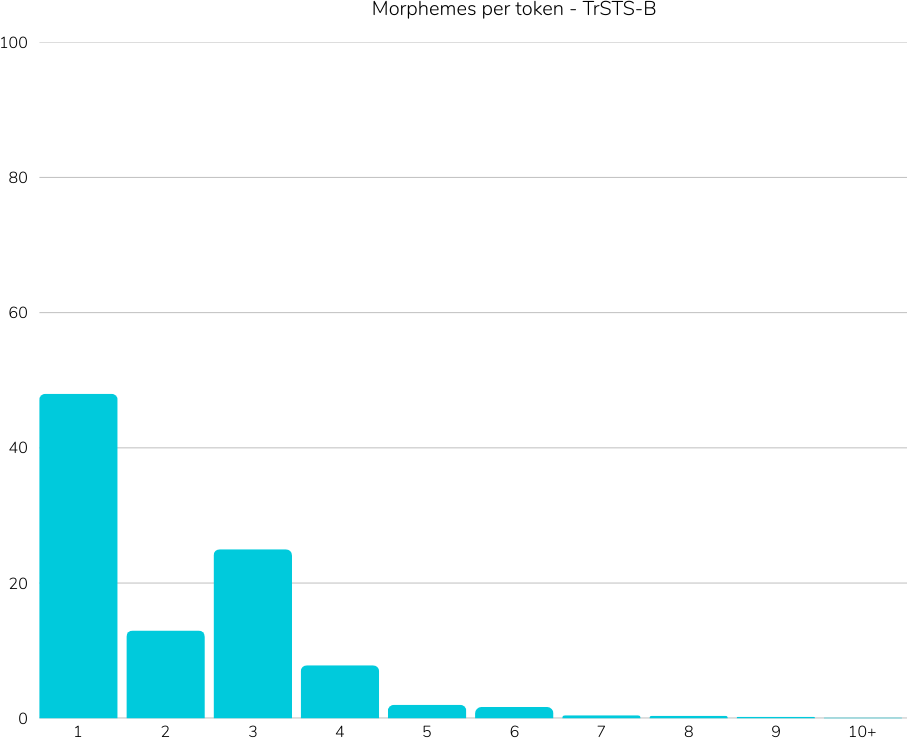}}\hfill
\subfloat[QQP]{\includegraphics[width=0.3\textwidth]{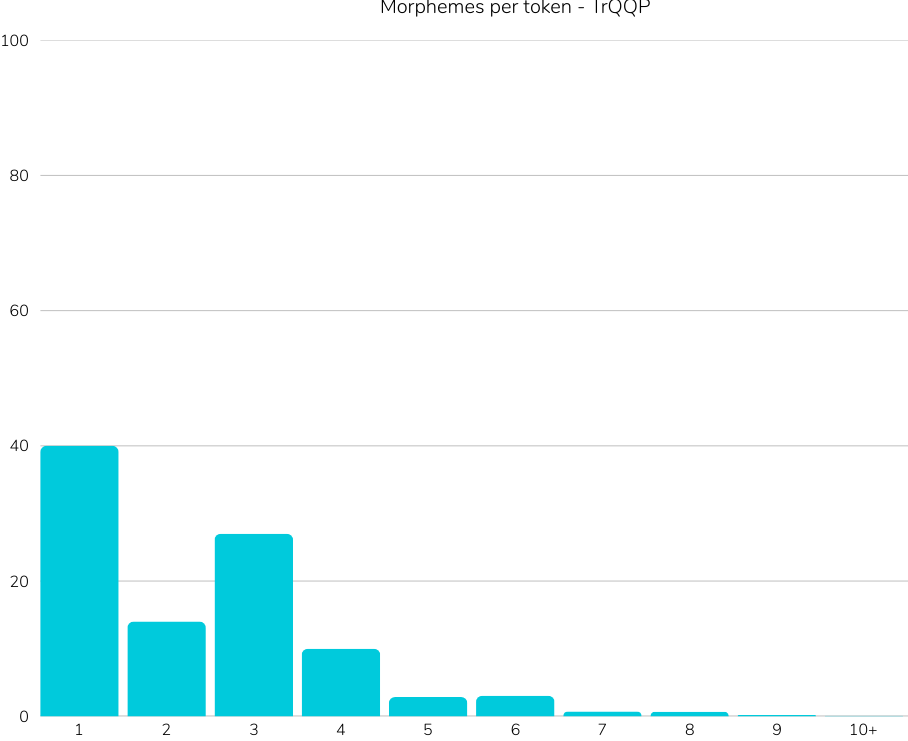}}\hfill
\subfloat[RTE]{\includegraphics[width=0.3\textwidth]{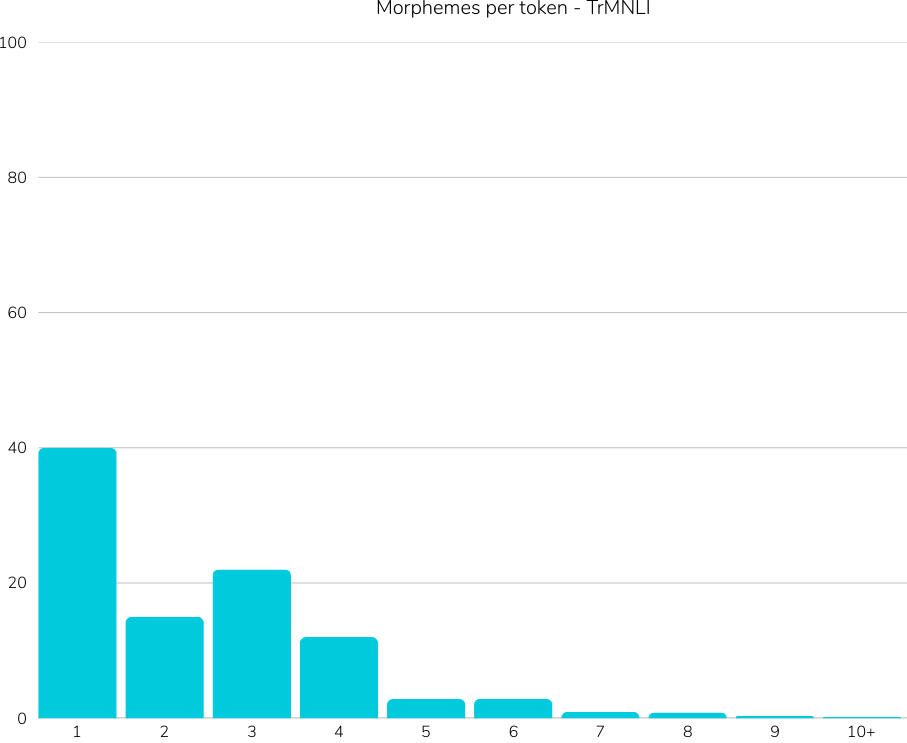}}

\vspace{0.6em}

% Row 3
\subfloat[SST-2]{\includegraphics[width=0.3\textwidth]{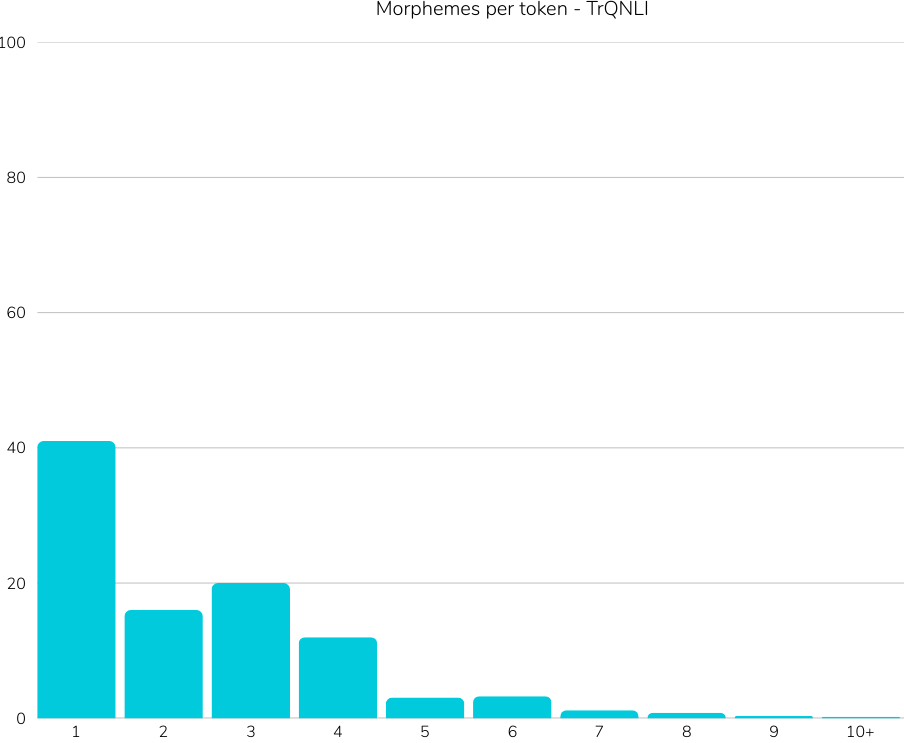}}\hfill
\subfloat[STS-B]{\includegraphics[width=0.3\textwidth]{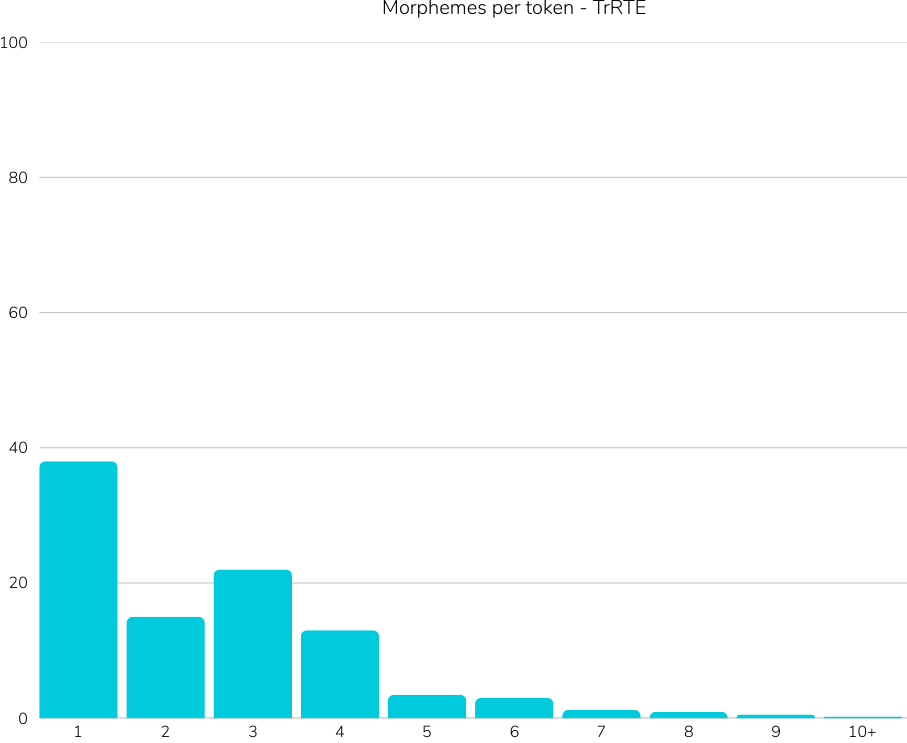}}\hfill
\subfloat[Overall]{\includegraphics[width=0.3\textwidth]{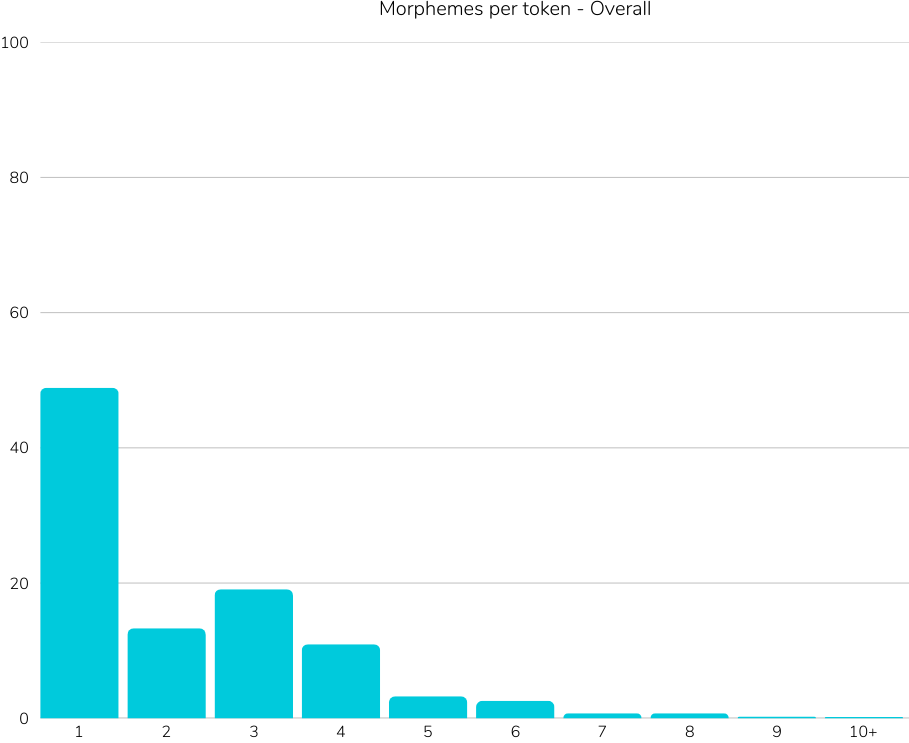}}

\caption{Morphemes-per-token histograms by subset. Each panel shows the distribution with a capped ``10+'' bin; vertical markers indicate the subset median and p95. Subsets differ mainly in tail weight and bundle diversity—heavier tails (e.g., MNLI/STS-B/RTE) reflect stacked inflection and discourse morphology, while lighter tails (e.g., CoLA/QQP) reflect shorter, templatic inputs. The pooled ``Overall'' summary reproduces the overall median 2 and heavy tail (p95=5, p99=8).}
\label{fig:histo-all}
\end{figure}

Discourse-layer morphology also varies systematically. Negation and evidentiality are most pronounced in the pooled ``Overall'' aggregate (11.75\% and ≈9.21\% of finite verbs), and are comparatively subdued in MNLI/RTE (3-6\% negation, 2-6\% evidentiality) and QNLI (2.76\% negation, 8.30\% evidentiality), while QQP shows moderate negation but low evidentials, consistent with user queries that favor imperative/instructional moods over source-of-knowledge marking. These differences matter for modeling: datasets with heavier MPT tails and richer bundle inventories tend to induce longer effective contexts and more parameter stress in subword segmentation, whereas lighter, templatic subsets emphasize lexical paraphrase and local morphosyntactic cues. Overall, the per-subset histograms (Figure \ref{fig:histo-all}) reveal a stable agglutinative backbone with task-specific shifts in tail mass and functional morphology that plausibly underlie observed performance variation across TrGLUE tasks.

Across tasks, we see consistent Turkish-typological behavior with task-specific nuances. In CoLA, pro-drop is high and full clauses are mostly canonical SOV, fitting short acceptability stimuli with local phenomena. MNLI shows strong pro-drop and modest non-canonicality across mixed genres, with many partial SV/OV observations likely from ellipsis between premise and hypothesis. MRPC mirrors newsy brevity with high subject omission, modest non-canonical rates, and relatively short dependencies. QNLI's interrogative origins yield elevated partial orders and slightly longer dependencies, while QQP's terse, template-like user questions produce the shortest dependencies despite substantial pro-drop. RTE displays notable OSV/OVS tails driven by topicalized objects and clausal complements, captured by the relaxed object heuristic, alongside moderate dependency length and frequent (zero-)copular structures. SST-2 comprises short evaluative snippets with strong subject drop and longer dependencies due to stacked modifiers, and STS-B (train+dev) shows similarly high pro-drop and somewhat more word-order variability reflecting heterogeneous sentence sources. Overall, Overall aggregated over all tasks (excluding STS-B test) confirm pervasive pro-drop, SOV dominance with occasional flexible alternations, and dependency lengths in a reasonable range for mixed-genre Turkish text.

Coming to the named entities, entity density and role vary notably by subset, aligning with each task's operational demands. TrQNLI shows the highest entity concentration (16.3\% entity tokens/token), reflecting its news-derived question-passage pairs where salient cues often hinge on named persons, organizations, and locations; resolving entailment frequently requires tracking coreferent entities across sentences and disambiguating near-synonymous organization or place names. TrMRPC likewise exhibits high entity load (15.0\%), and paraphrase decisions regularly turn on whether headlines and leads refer to the same named actor, event, or geopolitical unit; subtle surface variations (e.g., honorifics, transliterated names, or suffixed proper nouns) can flip labels. RTE (9.2\%) features compact premise-hypothesis pairs in which veracity depends on fine-grained entity attributes (titles, nationalities, affiliations), making entity linking and appositive parsing disproportionately influential. TrMNLI's moderate density (8.4\%) spans genres; entity cues vary widely by domain (fictional names vs. newswire), challenging models to maintain consistent entity representations across styles. TrQQP is comparatively sparse (6.4\%): user queries often generalize over entity slots (``how to get visa for X''), so robustness hinges on handling entity placeholders and slot-filling paraphrases rather than rich, document-grounded references. TrCoLA is the second sparsest (6.3\%), as acceptability judgments depend primarily on morphosyntax; when entities do occur, they are often inert with respect to the label, though proper-noun morphology can still stress tokenization. TrSTS-B (2\%) contains few entity tokens, reflecting its caption-based origin from video and image descriptions. We note that the observed entity-token rates, while seemingly high due to spaCy's broad label inventory (including CARDINAL/DATE/TIME/MONEY), are consistent with news- and information-centric text. Taken together, these profiles suggest that entity-aware modeling offers the greatest payoff in TrQNLI and TrMRPC, with measurable but smaller gains in TrMNLI, TrSTS-B, and TrRTE, while TrQQP and TrCoLA primarily test generalization beyond explicit named references.

\subsubsection{Tokenization behavior}
Using BERTurk (32k vocab) on TrGLUE, we observe an average of 1.58 subwords per token, versus 1.29 for BERT ($\approx$ 28.9k vocab) on GLUE. This gap reflects Turkish's agglutinative morphology, which induces more wordpiece splits despite a slightly larger vocabulary. Practically, higher fragmentation increases effective sequence length (and thus compute) for Turkish inputs under the same max length, while improving coverage of rare forms. The magnitudes ($\approx$ 1.3 for English vs. $\approx$ 1.5-1.6 for Turkish) align with prior reports for comparable WordPiece settings.

\section{Evaluation using TrGLUE}
\label{sec:eval}

We utilized the created benchmark to evaluate the widely used pretrained model BERTurk. For the evaluation process, we employed the benchmarking GLUE script from the HuggingFace repository\footnote{\url{https://github.com/huggingface/transformers/blob/main/examples/pytorch/text-classification/run_glue.py}}, making minimal modifications. As mentioned earlier, we maintained the original dataset format to leverage existing resources.

Table \ref{tab:perf} reports the performances of BERT-base and RoBERTa-base on GLUE alongside BERTurk on TrGLUE. As expected, all models struggle on CoLA, which is both small and linguistically challenging. BERTurk notably outperforms BERT on RTE, while underperforming on STS-B relative to corresponding GLUE baselines. The RTE dataset sizes in TrGLUE and GLUE are nearly identical, suggesting that scale alone does not explain the difference. In contrast, TrGLUE's STS-B is roughly one third the size of the original, which may contribute to the gap observed. Overall, the results indicate a broadly balanced performance profile across tasks.

\begin{table}[h]
\captionsetup{width=0.8\textwidth}
\caption{Performances of BERT-base and RoBERTa-base on GLUE vs. BERTurk on TrGLUE. Note that the original GLUE uses only accuracy for SST-2; we report accuracy and F1-score due to class imbalance.}
\label{tab:perf}
\centering
\begin{tabular}{|l|l|l|l|}
\hline
Corpus & BERT & RoBERTa &  BERTurk   \\ \hline
CoLA & 52.1  & 59.8 & 42 \\ 
SST-2 &  91.6/91.9 & 94.2/94.3 &  87.4/91.3 \\
MRPC & 78.4/85.4 & 88.9/92.0 & 74.4/72.7  \\
STS-B & 87.2/86.9 & 90.5/90.2 & 71.3/69.9  \\
QQP & 90.1/86.7  & 90.8/88.7 & 95.4/94.3 \\
MNLI & 84.6/83.7 & 86.9/87.3 & 87.9/90.8  \\
QNLI & 90.5 & 92.3 & 90.6  \\
RTE & 67.8 & 75.4 & 92.2   \\
\hline
\end{tabular}
\end{table}

\begin{figure}[ht]
\centering
\subfloat[Learning curves as a function of training fraction. Metrics are normalized per task by each task's full-data score, enabling comparison across heterogeneous metrics.]{\includegraphics[width=0.49\textwidth]{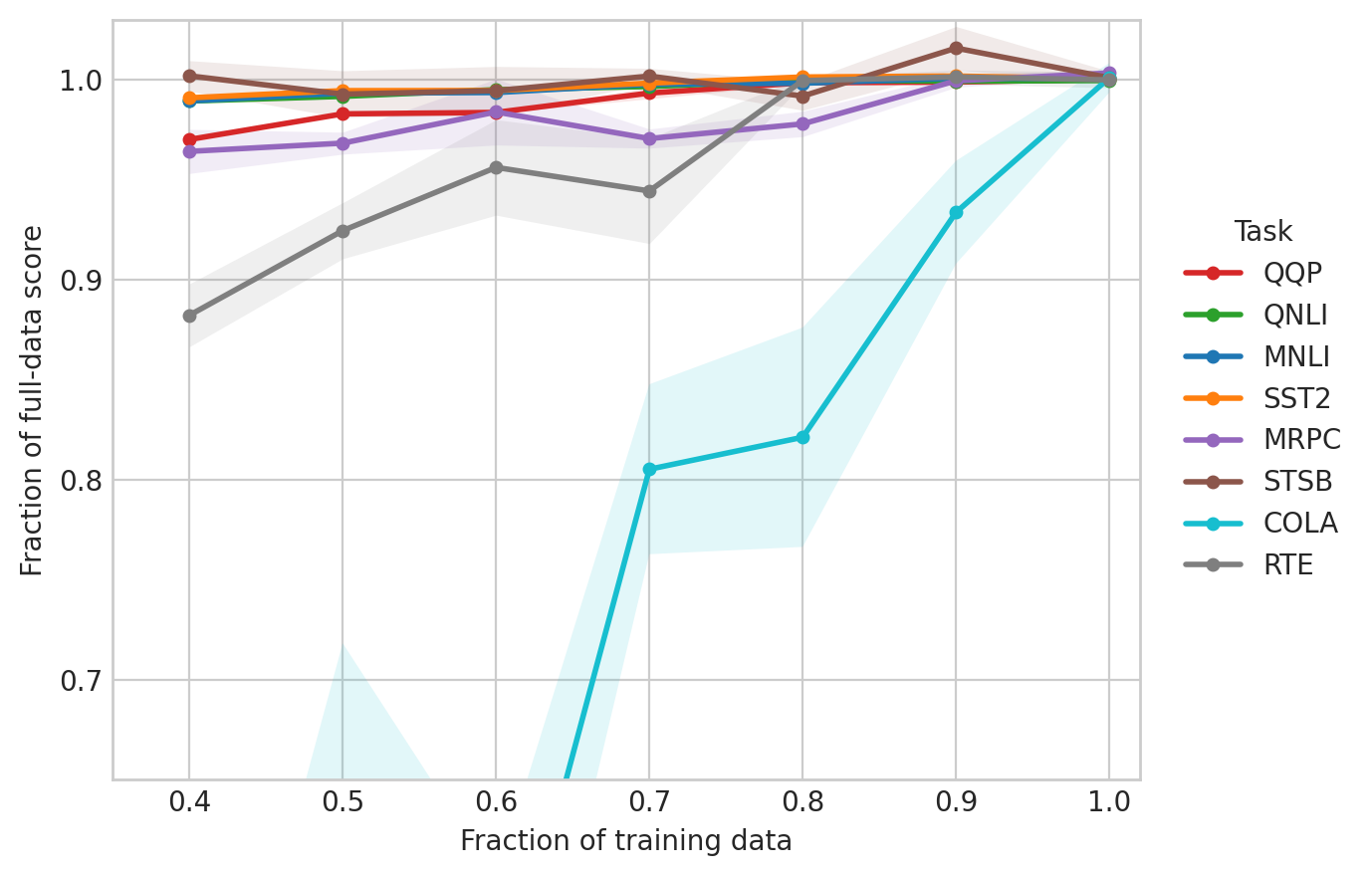}}\hfill
\subfloat[TrRTE vs. RTE. TrRTE saturates by 0.8\texttimes with higher absolute accuracy, while English RTE shows limited gains between 0.6-1.0\texttimes and seed-level fluctuations with overlapping confidence bands.]{\includegraphics[width=0.49\textwidth]{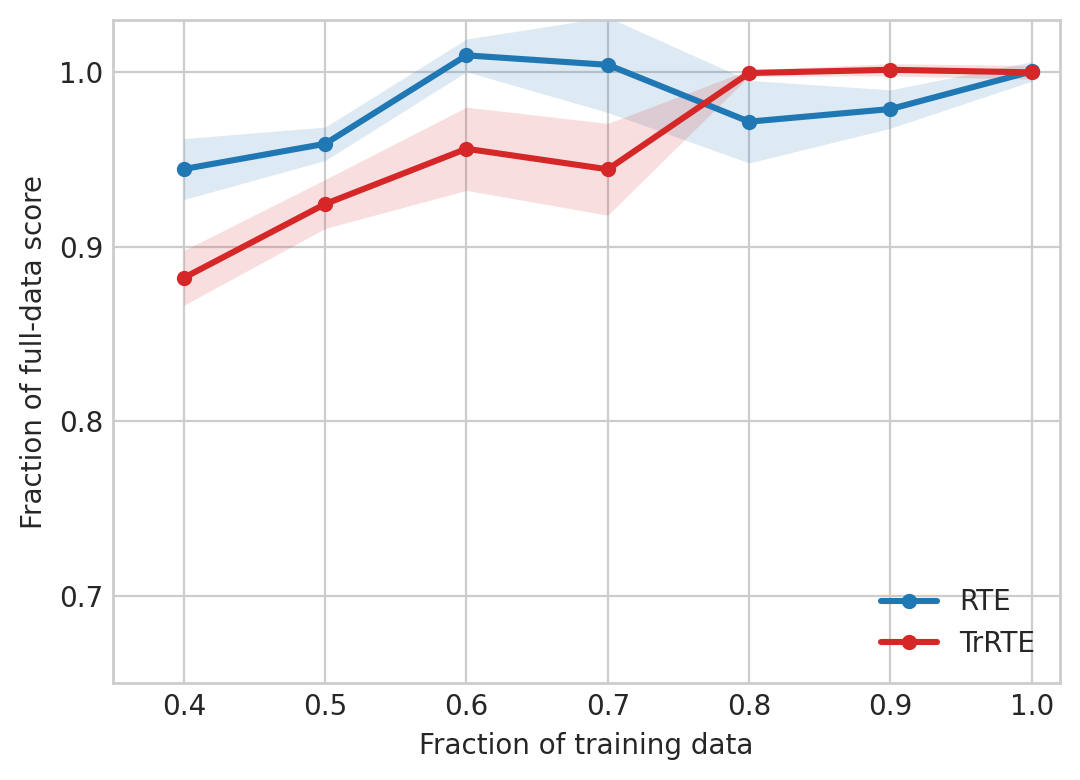}}
\caption{Per-task learning dynamics under fractional training.}
\label{fig:vary}
\end{figure}

To assess whether the constructed training sizes are sufficient, we trained with fractions of the available data (0.4-1.0) and plotted learning curves as ``fraction of training data'' versus ``fraction of full-data score.'' Figure \ref{fig:vary} shows that most tasks approach saturation well before using all available training examples, suggesting the current dataset sizes are adequate for the evaluated model class and setup.

To assess whether the constructed training sizes are sufficient, we trained with fractions of the available data (0.4-1.0) and plotted learning curves as ``fraction of training data'' versus ``fraction of full-data score.'' As shown in Figure \ref{fig:vary}, classification tasks (QQP, QNLI, MNLI, SST-2, MRPC) reach $geq$95\% of their full-data scores by 0.6-0.8\texttimes with narrow confidence bands, and RTE similarly plateaus with overlapping bands beyond 0.6×. In contrast, TrSTS-B improves more gradually and continues to trend upward at 1.0×, indicating residual headroom at current scale.

In addition, we examined TrRTE versus original RTE under the same fractional-data protocol. Figure \ref{fig:vary} overlays their normalized curves. TrRTE reaches its plateau by roughly 0.8\texttimes of the training set, whereas English RTE varies only modestly from 0.6-1.0, with the shaded confidence bands largely overlapping. These trends indicate that TrRTE is not data-limited in this range and that the English curve exhibits only small, seed-driven fluctuations rather than systematic gains.

\paragraph{TrRTE vs RTE}
As seen in Table \ref{tab:perf}, TrRTE achieves a strong result (0.92 at full data), while English RTE hovers around 0.67 under the same recipe. The fractional-data curves and the overlay in Figure \ref{fig:vary} show early saturation for TrRTE (0.6-0.8) with minor seed-level variability thereafter. By contrast, English RTE shows limited sensitivity to additional training data beyond 0.6\texttimes: mean accuracy across 0.6-1.0 varies within roughly 2-3 points with a small dip around 0.8, and confidence bands largely overlap, suggesting fluctuations are seed-driven rather than systematic.

We attribute the gap to dataset adequacy rather than size or translation effects. TrRTE is handcrafted natively in Turkish: premises are sampled from licensed news and academic sources, and hypotheses are authored under three styles (linguistic, free, factual) to ensure both systematic coverage (e.g., negation, quantification, agreement/case, modality/evidentiality, word order) and natural variation. We enforce controls to reduce superficial cues (balanced lexical overlap across labels; limits on overt-negation flips; diversified contradiction mechanisms such as number/date/role/entity changes; no verbatim copying), deduplicate near-duplicates, and maintain 50/50 label balance within each split. Each pair receives independent labels with adjudication, and ambiguous items are excluded. These design and QA choices yield clearer supervision per example, leading to higher absolute accuracy and earlier saturation with respect to training fraction (Figure \ref{fig:vary}).

\paragraph{STS-B} 
Despite careful translate-then-edit construction, TrSTS-B yields lower absolute correlation than the English STS-B under the same recipe Table \ref{tab:perf}. The fractional-data curves in Figure \ref{fig:vary} show a slower approach to saturation than classification tasks: gains from 0.4\texttimes to 0.9\texttimes are steady but modest, and the curve continues to climb at 1.0\texttimes, suggesting residual headroom rather than early plateau. We attribute this gap to measurement granularity and scale, not task mismatch. First, our Turkish set is roughly one third the size of STS-B, which increases estimator variance for continuous scores and makes fine-grained rank ordering harder to learn; this is visible in the wider confidence band for STS-B/TrSTS-B compared to discrete-label tasks. Second, Turkish surface variation (article absence, richer morphology, productive compounding) introduces more lexical-syntactic paraphrase diversity per meaning unit, so sentence-level similarity depends more on morphology-aware cues that BERTurk underutilizes relative to large English baselines. Third, our cultural substitutions (about 40 targeted edits) reduce literal lexical overlap by design; while they preserve event semantics, they attenuate n-gram and entity matching signals that correlation-tuned objectives exploit. Finally, de-duplication and collapsed-pair filtering remove many near-identical pairs that in English act as ``easy anchors'' at the top of the scale; their absence yields a sharper distribution and fewer trivial high-similarity examples, lowering absolute correlation without indicating poorer semantic competence. Together, these factors explain the lower score alongside a learning curve that has not fully saturated at 1.0\texttimes, pointing to expected improvements with larger Turkish data or models with stronger morphology-sensitive similarity modeling.

\subsection{Investigation of the results}
We now turn our focus to the CoLA dataset, which posed significant challenges for a BERT-based model, resulting in poor performance and limited success.

\subsubsection{CoLA}
The CoLA dataset has proven to be exceptionally challenging, even for some state-of-the-art multilingual closed and open-source LLMs known for their English performance. Human evaluators also found the task daunting, with all annotators expressing difficulty. Despite possessing substantial experience in linguistics, particularly Turkish linguistics, the author encountered challenges when analyzing sentences from this dataset. Some sentences had complex semantics, necessitating multiple readings for comprehension.

In our experimentation, we engaged with Claude 3 Sonnet \cite{claude}, Gemini 1.0 Pro, GPT-4 Turbo \cite{gpt4}, LLaMa 3 70B \cite{dubey2024LLaMa3herdmodels}, and Qwen2-72B \cite{yang2024qwen2technicalreport}. These interactions occurred on the Poe.com platform. With each LLM, we initiated conversations with a greeting, followed by an informative sentence about the task and the input-output format. Subsequently, we presented the test sentences in segments and obtained the corresponding labels. Details about the LLM evaluation process can be found in Appendix \ref{sec:llm-trglue}. Notably, all results presented are zero-shot results.

\begin{table}[h]
\captionsetup{width=0.8\textwidth}
\caption{0.0 correlation coefficient being the score for random guessing, it was surprising that LLaMa 3 70B initially achieved nearly 0.0. Evidently, this model lacked understanding of Turkish grammar and semantics. However, with a coherent chain of thought, its performance notably improved. Claude 3 Sonnet, Gemini Pro, and GPT-4 did not fare well either. Only Qwen2-72B managed to produce significant results and demonstrated proficiency in Turkish.}
\label{tab:cola-llm-success}
\centering
\begin{tabular}{|l|l|}
\hline
Model & Matthews' corr. \\ \hline
Gemini 1.0 Pro & 0.21 \\
GPT-4 Turbo &  0.28\\
Claude 3 Sonnet & 0.14 \\
LLaMa 3 70B &  0.05 \\
LLaMa 3 70B CoT & 0.35  \\
Qwen2-72B & 0.47 \\
BERTurk & 0.42 \\
\hline
\end{tabular}
\end{table}

Table \ref{tab:cola-llm-success} displays the success of various closed- and open-source LLMs on TrCoLA, alongside BERTurk for comparison. Surprisingly, despite the impressive reputation, Claude 3 Sonnet, GPT-4, and Gemini Pro showed lackluster performance. Conversely, LLaMa 3 70B and Qwen2-72B emerged as clear frontrunners.

\begin{figure}[ht]
\includegraphics[width=\textwidth]{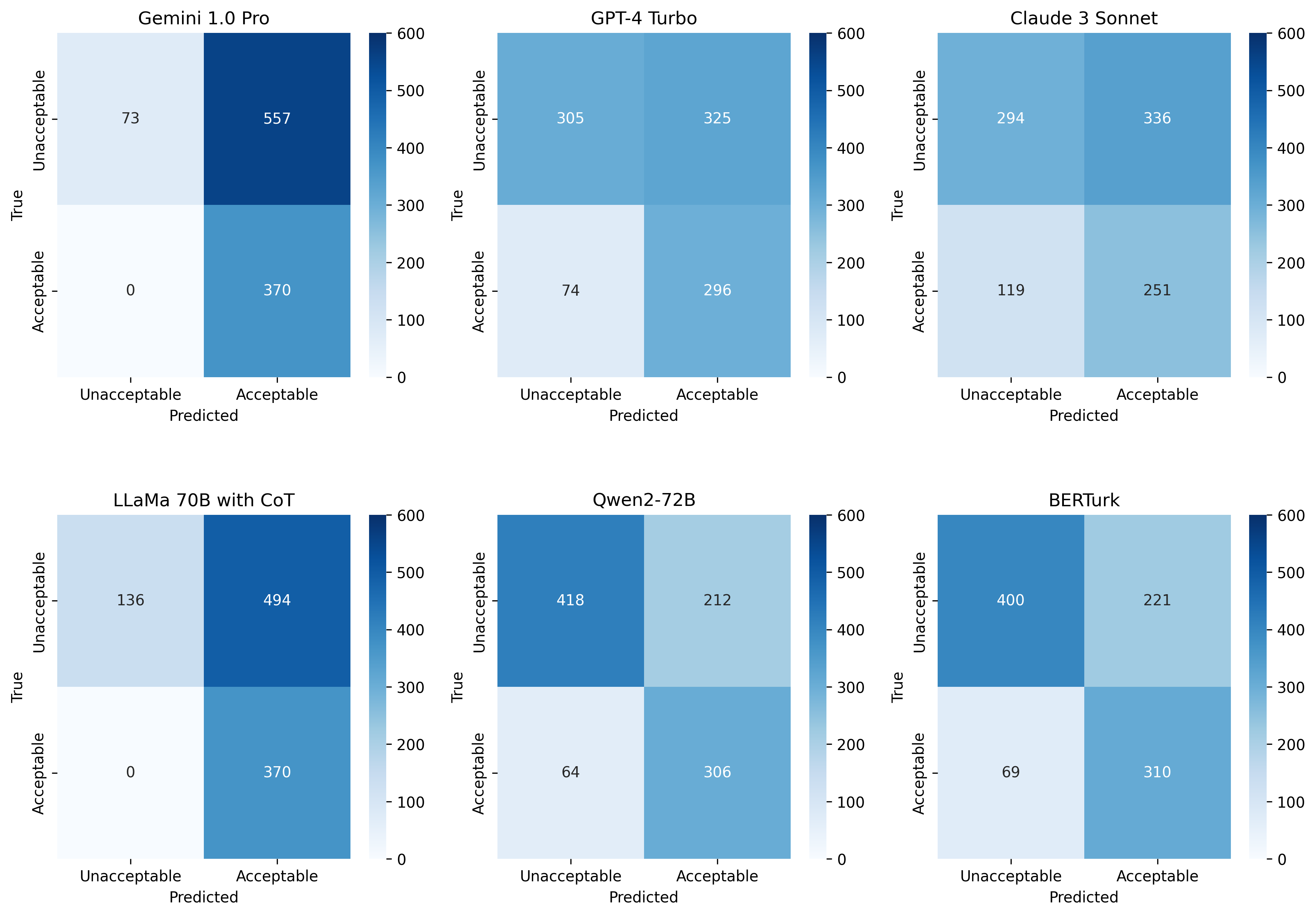}
\caption{Confusion matrices of the LLMs utilized in our study on the CoLA dataset.}
\label{fig:llm-confusion}
\end{figure}

Dissecting the results and examining the confusion matrices depicted in Figure \ref{fig:llm-confusion}, we observe that Gemini failed completely in the task. It struggled to differentiate between acceptable and unacceptable sentences, often mislabeling each sentence as acceptable. Gemini predominantly misclassified syntactic variation sentences as true negatives, while struggling with semantic, morphological, and some syntactic variations.

GPT-4 Turbo also exhibited subpar performance, albeit showing some ability to distinguish unacceptable sentences to a certain extent. However, the poorest performance was seen in Claude 3 Sonnet.

Regarding LLaMa 3 70B, the results reveal two distinct outcomes. Initially, a Matthews' correlation coefficient of 0.05 was observed, almost akin to random guessing. However, when prompted with a chain of thought task to provide a label for each sentence along with a brief justification, LLaMa made a remarkable leap from nearly 0 to 0.35, delivering a significant improvement. Similar to Gemini, LLaMa struggled primarily with false positives and faced challenges in discerning unacceptable sentences.

\begin{figure}[h!]
\begin{tabular}{cc}
\subfloat[Qwen2-72B's response to the provided test sentence.]{\includegraphics[width=0.48\textwidth]{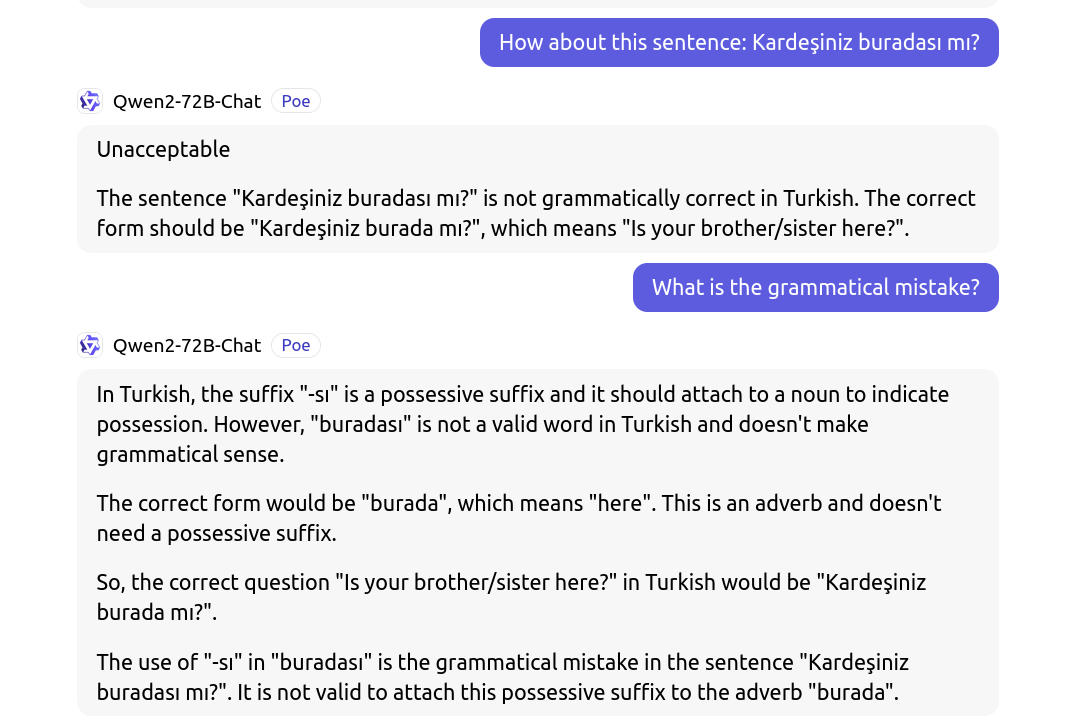}} &  \subfloat[LLaMa 3 70B's response to the provided test sentence.]{\includegraphics[width=0.48\textwidth]{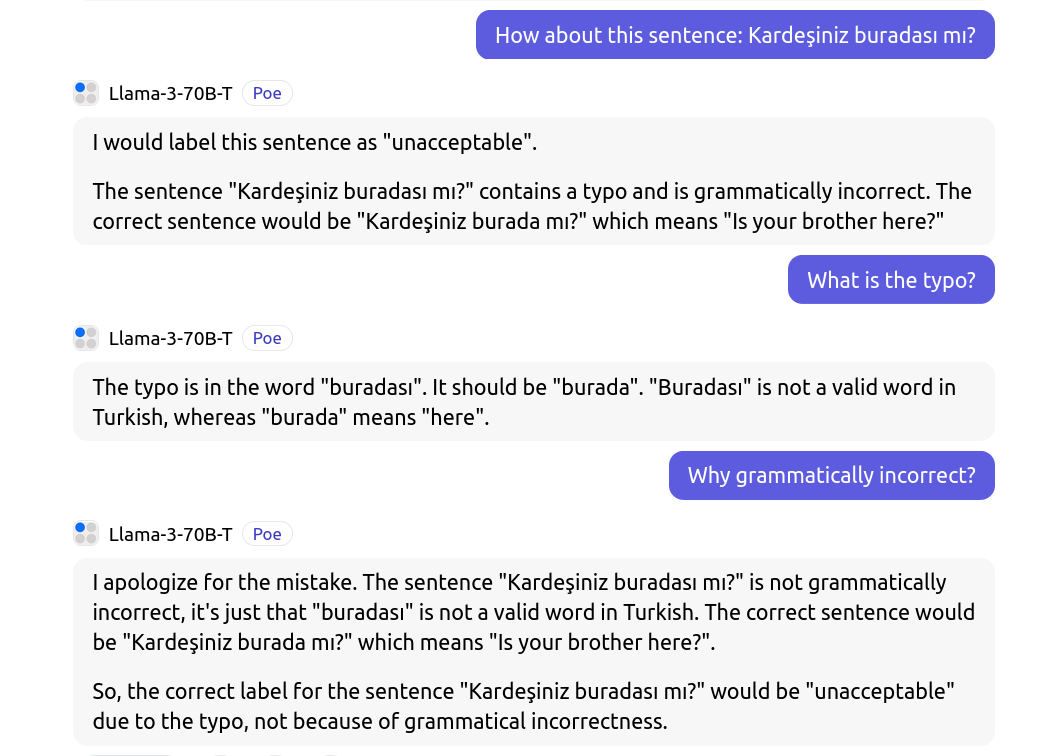}}
\end{tabular}
\caption{Justification of top performing LLMs for their responses to the chosen test sentence.}
\label{fig:cola-answers-top}
\end{figure}

While the results of LLaMa 3 70B and Qwen2-72B may appear similar, they are indeed quite different. Figure \ref{fig:cola-answers-top} showcases the responses from these two LLMs to the test sentence ``Kardeşiniz buradası mı?", correctly predicted by BERTurk, LLaMa, and Qwen2; and incorrectly predicted by Gemini and GPT-4. This sentence is grammatically incorrect as ``bura+da+sı" combines the locative suffix \emph{-(D)A} with the possessive suffix \emph{-(s)I} in a morphologically invalid manner according to Turkish grammar rules. When queried about this sentence, LLaMa responded that the sentence is not grammatically incorrect but rather that ``buradası" is not a valid word in Turkish, based purely on the word statistics it encountered during training. Conversely, Qwen2 provided the correct answer along with a grammatical explanation, showcasing its deep understanding of Turkish morphology and syntax.

\begin{figure}[h!]
\begin{tabular}{ccc}
\subfloat[Gemini 1.0 Pro's response to the provided test sentence.]{\includegraphics[width=0.32\textwidth]{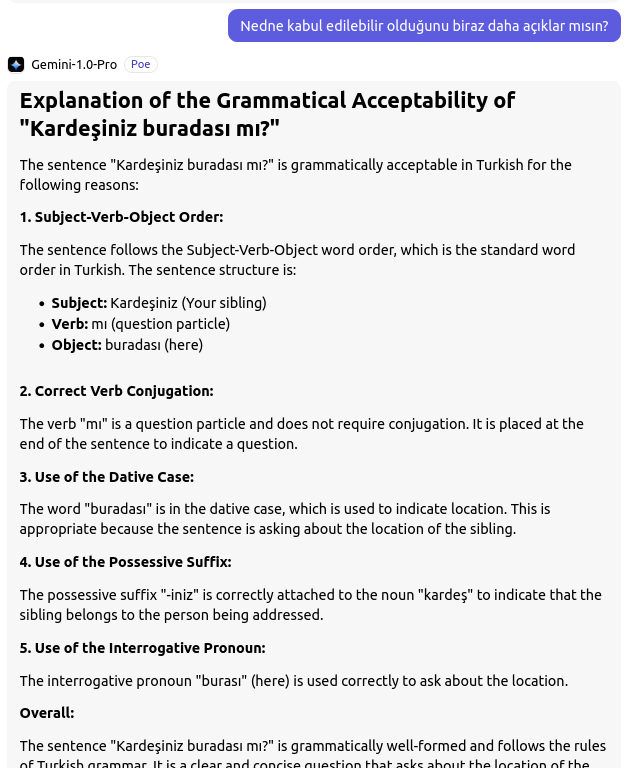}} & \subfloat[Follow-up question to Gemini Pro and its response.]{\includegraphics[width=0.32\textwidth]{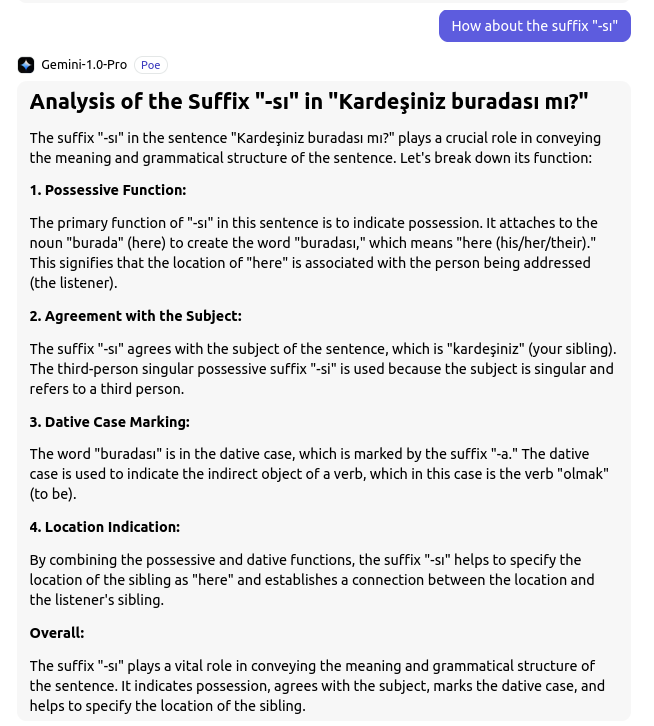}} & \subfloat[GPT-4 Turbo's response to the provided test sentence.]  {\includegraphics[width=0.32\textwidth]{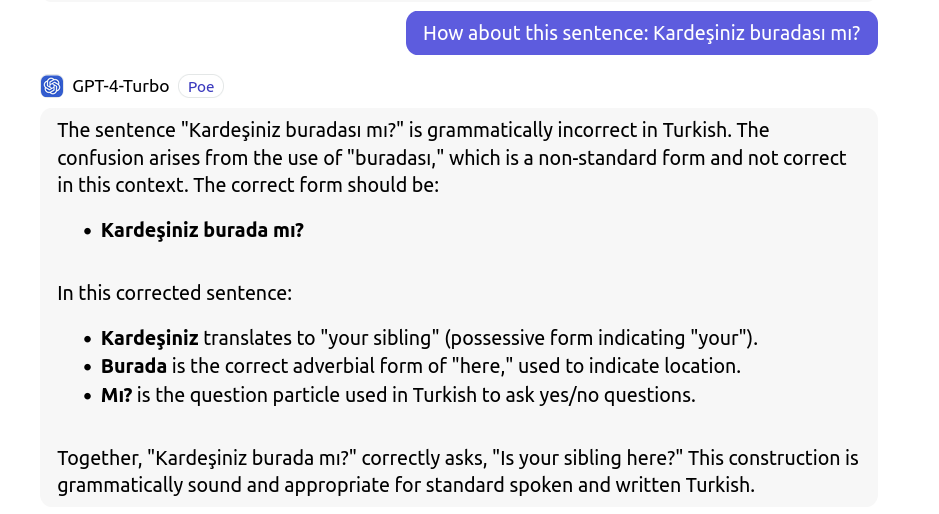}} \\ 
\subfloat[Claude 3 Sonnet's response to the provided test sentence.]{\includegraphics[width=0.32\textwidth]{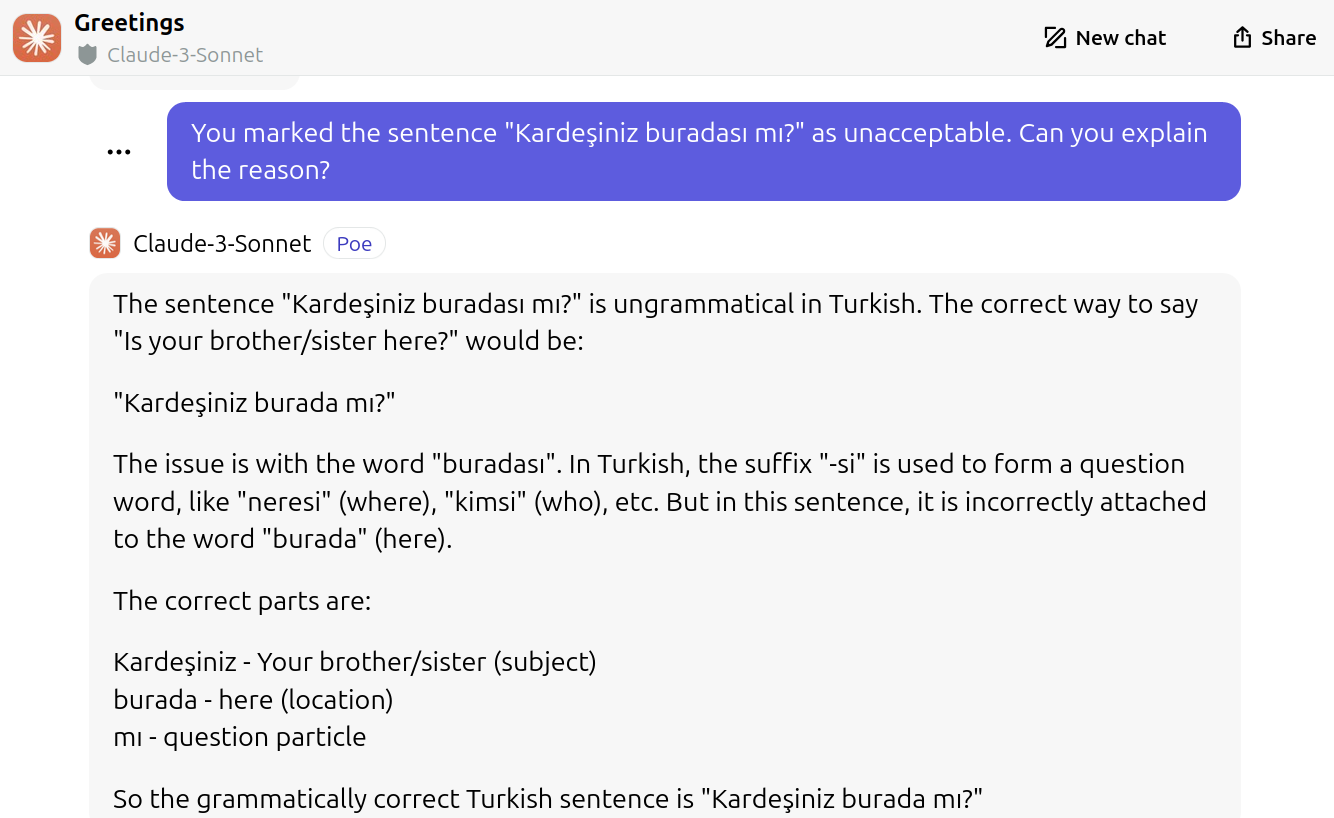}} & \subfloat[Follow-up question to Claude 3 Sonnet and its response.]{\includegraphics[width=0.32\textwidth]{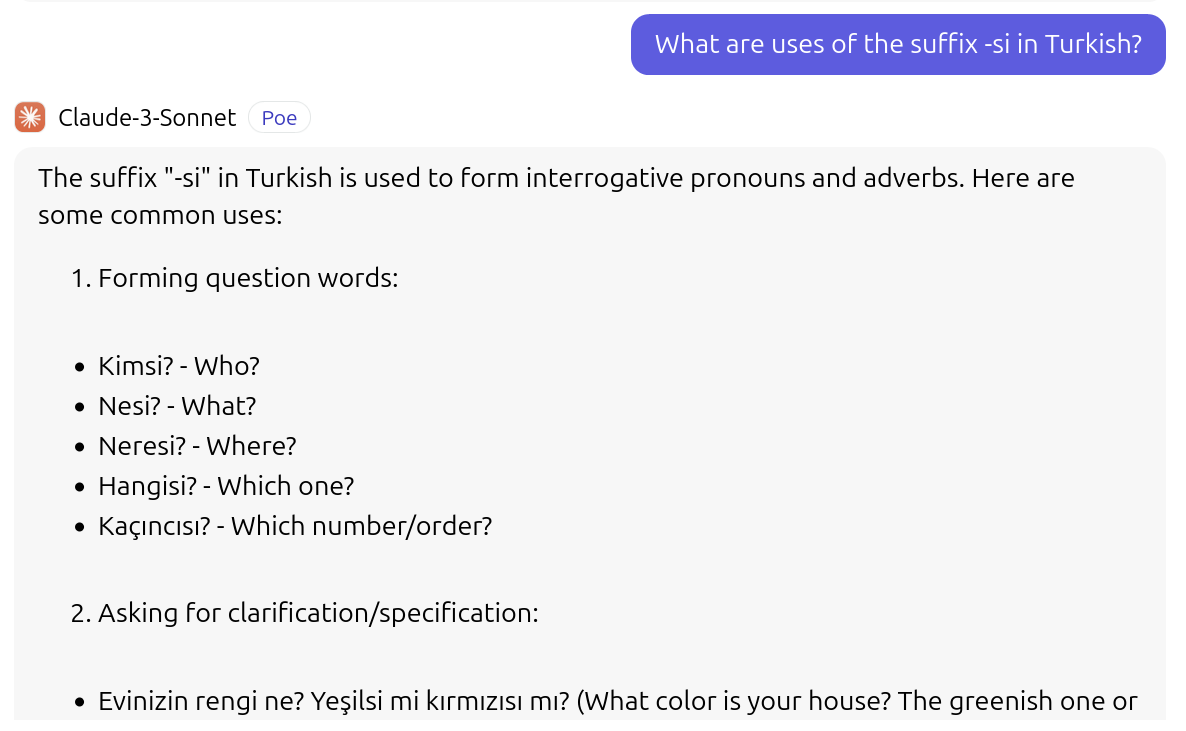}} &
\subfloat[Continuation of Claude's answer.]{\includegraphics[width=0.32\textwidth]{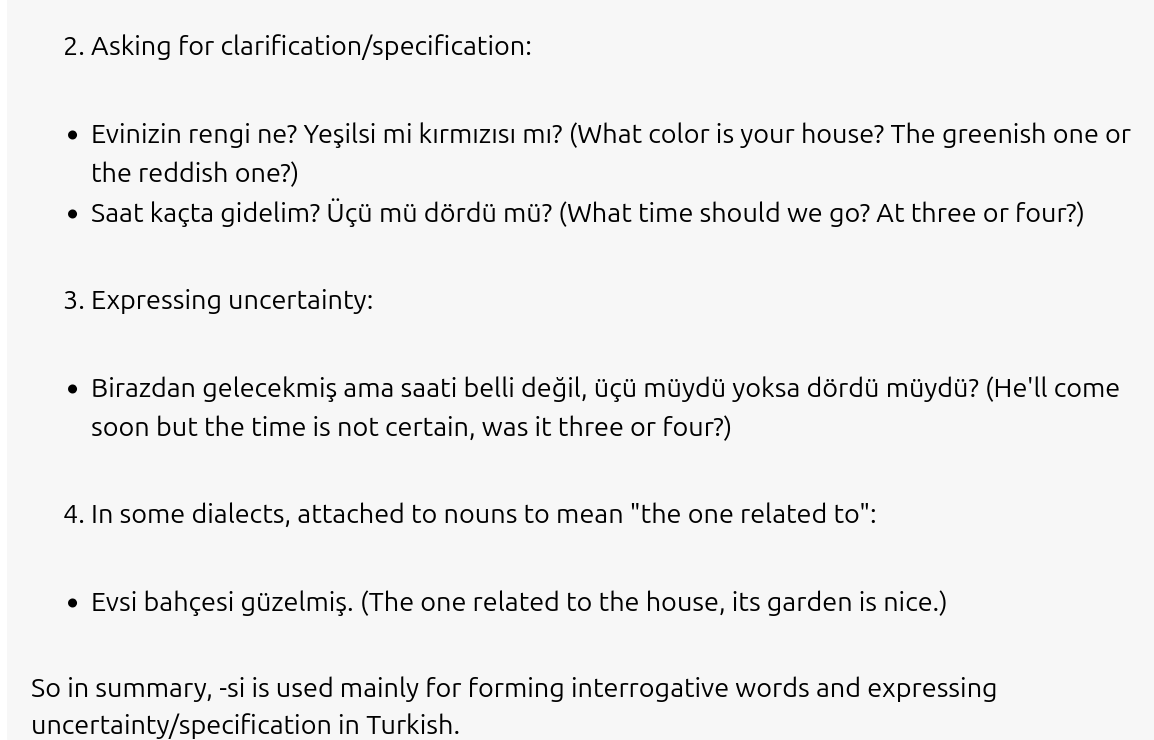}}
\end{tabular}
\caption{Justifications of underperforming LLMs for their responses to the chosen test sentence.}
\label{fig:cola-answers-bottom}
\end{figure}

When examining the responses of Gemini Pro in Figure \ref{fig:cola-answers-bottom}, it appears that the model attempts to evaluate the sentence within the context of Turkish grammar or semantics. However, a critical error was made concerning the role of the possessive suffix \emph{-(s)I}, incorrectly claiming its correct usage twice. Even when prompted specifically about this suffix to provide a clue, Gemini Pro did not improve its understanding. On the other hand, GPT-4 provided a correct answer but did not specify the nature of the mistake.

In the case of Claude 3 Sonnet, it correctly labeled the sentence as ``unacceptable"; however, the rationale behind this decision was solely based on corpus statistics, without any mention of morphology or the incorrect usage of the possessive suffix \emph{-(s)I}. When explicitly questioned about the \emph{-(s)I} suffix, Claude 3 Sonnet failed to recognize it as a possessive suffix, offering erroneous claims about Turkish grammar. This indicates a lack of consciousness regarding Turkish morphology and the provision of inaccurate information to users.

It is evident that all LLMs exhibit some understanding of corpus statistics, possess certain syntactic knowledge, deduce syntactic rules, and comprehend semantics to some extent. However, when delving into the intricate mechanisms of the Turkish language, particularly its morphology and syntax in detail, they fall short, with the exception of Qwen2-72B. Qwen2-72B stands out as the only model that truly embodies Turkish language knowledge at its core.

Regarding BERTurk, the model demonstrated proficiency in identifying morphological errors effectively. Unsuccessful classifications primarily stemmed from syntactic and semantic variations.

For both the CoLA task, LLMs do not exhibit the same level of success as their English dataset counterpart. A recent study by \cite{jwino} discovered that the performance of ChatGPT, also known as GPT-3.5 \cite{gpt35}, on the Japanese WNLI task does not correlate with its performance on the English WNLI task. The study evaluated GPT-3.5's performance on tasks involving world knowledge and commonsense reasoning relative to human performance in Japanese. Their findings indicate that ChatGPT achieves lower accuracy compared to humans. The study highlighted that variations in training data across different languages can impact LLM performance, suggesting that success in English may not necessarily generalize to other languages, a notion supported by our own research.

\section{SentiTurca}

The SentiTurca dataset is tailored for sentiment analysis tasks, encompassing three distinct domains: movie reviews, customer reviews from e-commerce websites, and a newly introduced hate speech dataset. The movie and customer reviews were initially rated by reviewers, whereas the hate speech dataset has been annotated by human annotators from the data company Co-one. Examples from each dataset can be found in Appendix \ref{sec:sentiturca}.

\begin{table}[h]
\caption{Size of SentiTurca datasets.}
\label{tab:sentiturca-sizes}
\begin{tabular}{|l|l|l|l|l|}
\hline
Dataset & Train & Dev & Test & Metric \\ \hline
Movies & 60K & 8.9K & 8.9K & acc./F1 \\
Customer reviews & 73K & 15K & 15K & acc./F1 \\
Turkish Hate Map & 42K & 5K  & 5K & acc./F1 \\
\hline
\end{tabular}
\end{table}

\subsection{Movie reviews}
This dataset consists of movie reviews, the same as TrSTS-2 dataset, detailed in Section \ref{sec:movies}. We incorporated it into the SentiTurca benchmark to ensure thorough coverage. It is reiterated in this section for completeness.

\subsection{Customer reviews}

This dataset was collected by scraping two distinct e-commerce platforms, Hepsiburada and Trendyol, amassing a total of 103K reviews. Each data instance in the dataset comprises a review text along with a star rating ranging from 1 to 5. These reviews encompass a wide array of product categories, including apparel, food items, baby products, and books. The size breakdown of each segment of the dataset is detailed in Table \ref{tab:sentiturca-sizes}.

Initially, all reviews and ratings for each product were crawled. However, not all reviews were directly utilized due to mismatches between review texts and star ratings. Certain users indicated in their reviews that they granted a 5-star rating to boost visibility, followed by a negative critique. To filter out such reviews, semantic grouping and pattern matching techniques were applied. Outlier candidates were identified by clustering 5-star reviews based on semantic similarities, common patterns were extracted among these candidates, and a list of patterns signaling such review texts was compiled. These reviews were then filtered out using the compiled patterns list.

\begin{figure}[h!]
\begin{tabular}{cc}
\subfloat[Representation of star distribution within the dataset.]{\includegraphics[width=0.48\textwidth]{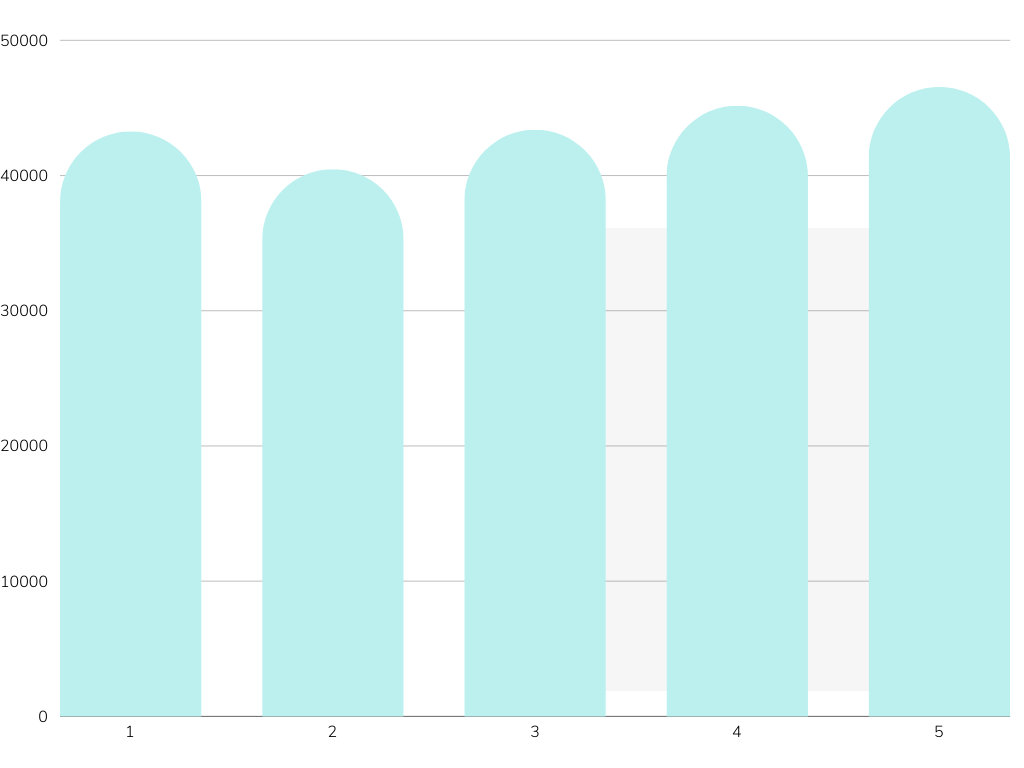}} &  \subfloat[Illustration of word count distribution in customer reviews.]{\includegraphics[width=0.48\textwidth]{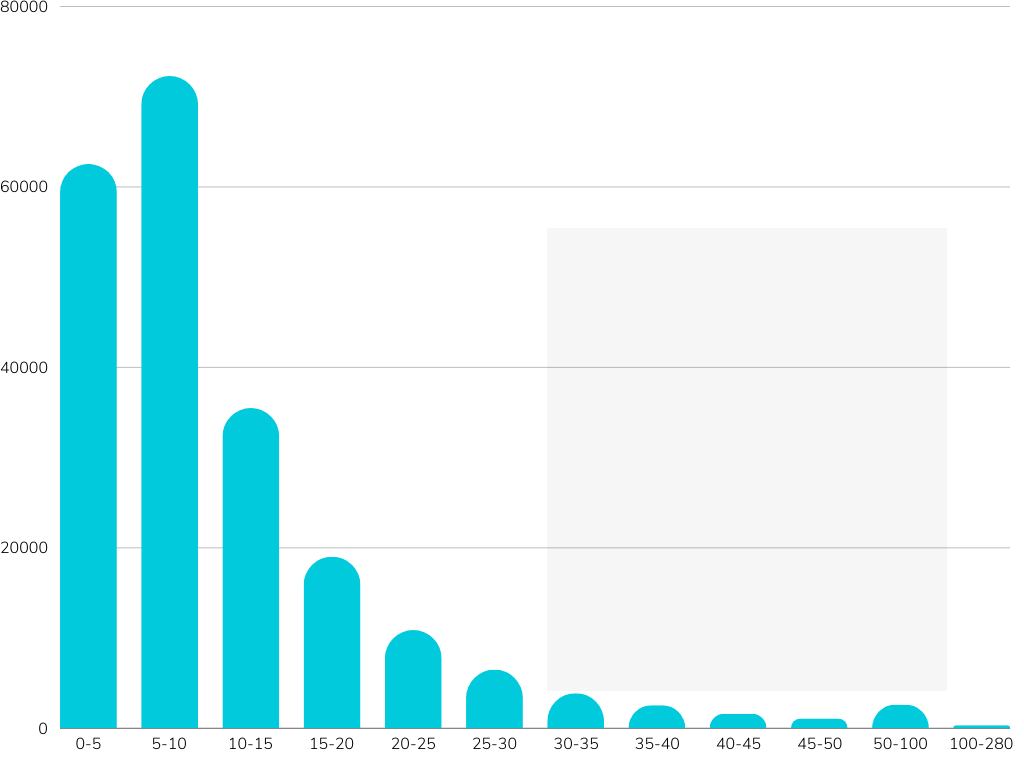}}
\end{tabular}
\caption{Data insights from the corpus of customer reviews dataset.}
\label{fig:ecommerce-stats}
\end{figure}

Upon data analysis, it was noted that 60\% of the ratings were either 4 stars or 5 stars, indicating that customers tended to leave reviews primarily when satisfied with the product. This trend is depicted in Figure \ref{fig:ecommerce-stats}.

Figure \ref{fig:ecommerce-stats} showcases the word count distribution within the dataset. On average, each review consists of 9.39 words and 70.26 characters, which aligns with the concise nature of product reviews. Moreover, a notable portion of reviews includes emoticons.

The significance of this dataset lies in its scale, being the most substantial collection of customer reviews published for Turkish to date. Previous studies, such as \cite{article_310341} with 2K reviews and \cite{kocaman} with 150K reviews, fall short in terms of scale and quality; the former dataset is notably smaller, and the latter offers only three classes—negative, positive, and neutral. Although \cite{class-ecomm} focused on categorizing around 15K e-commerce reviews, they did not publicly release the dataset.

This dataset is an integral part of SentiTurca and has its dedicated repository on Hugging Face as well\footnote{\url{https://huggingface.co/datasets/turkish-nlp-suite/MusteriYorumlari}}.

\subsection{Turkish Hate Map}

This dataset is a component of SentiTurca and is also accessible independently on Hugging Face.\footnote{\url{https://huggingface.co/datasets/turkish-nlp-suite/TurkishHateMap}}

Comprising 52K entries distributed across 13 distinct categories such as misogyny, political animosity, animal aversion, vegan antipathy, ethnic group hostility, and more, these entries were sourced from the collaborative hypertext compendium, Ekşi Sözlük. Within this platform, users contribute texts on a myriad of subjects spanning life, politics, and everyday concerns. Each entry is associated with its own webpage on the site, facilitating users in exploring grouped texts based on the entry. Details regarding the target categories and the corresponding entry counts are outlined in Table \ref{tab:hate-subjects}.

\begin{table}[h!]
\captionsetup{width=0.8\textwidth}
\caption{Hate Target Categories: Ethnic groups encompass a variety of ethnicities within Turkey, including Kurds, Laz, Circassians, and even Turks. Sects denote various branches of Islam, such as Shiism, Hanafism, Sunnism, and Shafiism. The Animals category encompasses both domestic pets and stray animals.}
\label{tab:hate-subjects}
\begin{tabular}{|l|l|}
\hline
Target group & Size \\ \hline
Animals & 1.2K  \\
Cities &  1.2K \\
Ethnic groups & 4.4K    \\
LGBT &  1.1K\\
Misogyny & 19.9K \\
Occupations & 0.8K \\
Politic &  12.6K\\
Political orientation & 3.4K \\
Refugees &  2.1K \\
Religion & 2.1K \\
Sects & 1.5K \\
Vegans-vegetarians & 1.3K \\
Total & 52K \\
\hline
\end{tabular}
\end{table}

Four labels were employed for annotation: offensive, hate, neutral, and civilized. The ``civilized" classification signifies that a text conveys an opinion in a manner conducive to a civilized discussion with the author of the text. The breakdown of labels for each target group is illustrated in Figure \ref{fig:hate-dist}, with sample instances from each label and category detailed in Appendix \ref{sec:sentiturca}.

\begin{figure}[h!]
    \centering
    \includegraphics[width=\textwidth]{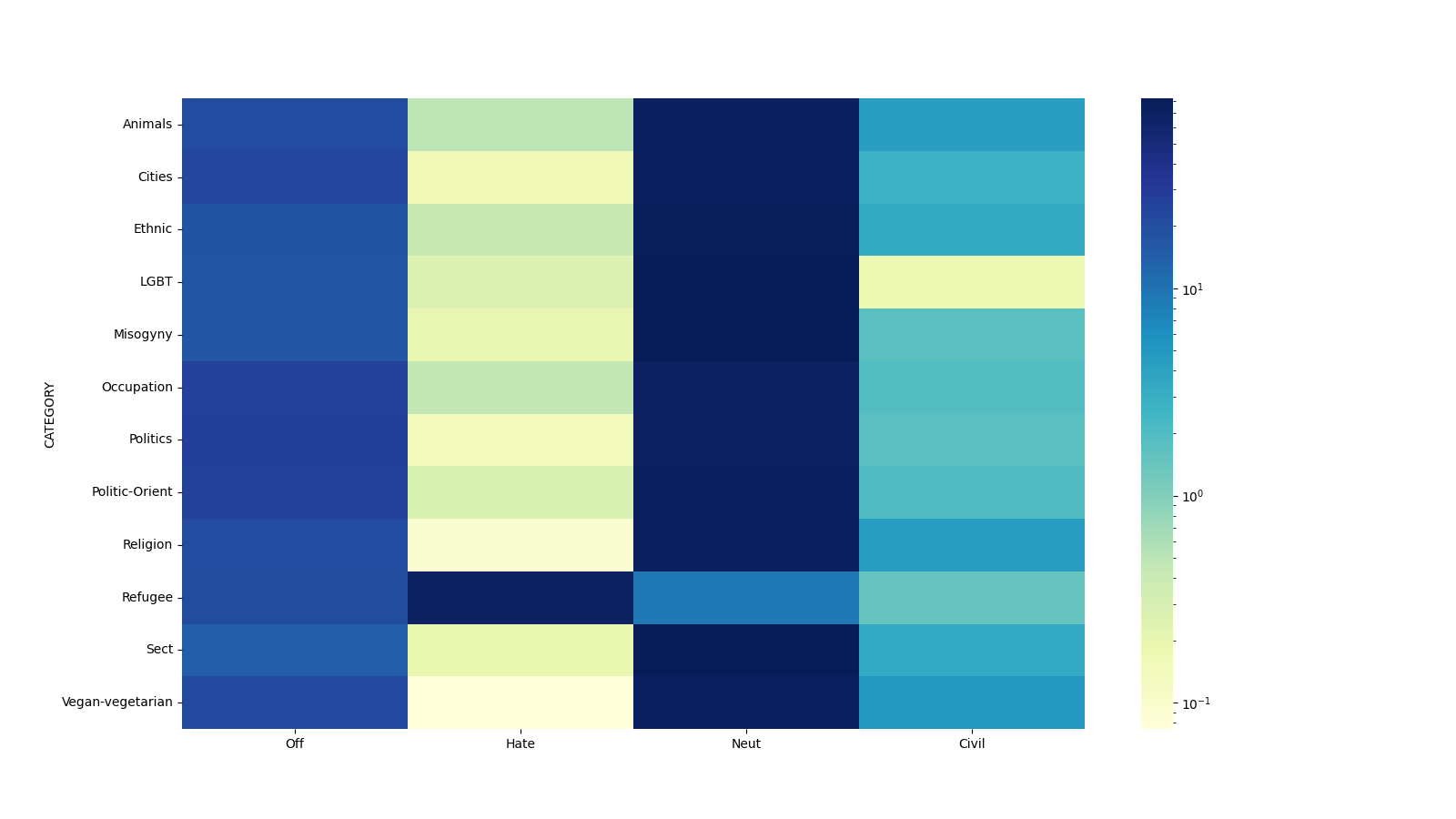} % Adjust the width as needed
    \caption{The distribution of labels across different target groups indicates that politics, political orientation, refugees, and misogyny demonstrate elevated levels of hate and offensiveness. These subjects frequently ignite heated discussions and draw substantial levels of animosity, especially concerning refugees and women. Notably, stray dogs have also been the subject of a notable amount of animosity in recent times.}
    \label{fig:hate-dist}
\end{figure}

When analyzing the distribution of labels across various categories, it becomes evident that the majority of the text falls into either the offensive or neutral categories. Hate speech and civilized expressions represent a minority, with one notable exception being the category of refugees. Within this category, hate speech predominates, closely followed by offensive language. There are notably fewer instances of neutral and civilized comments. This skew in distribution could be attributed to the substantial influx of migrants to Turkey since 2010, encompassing both irregular and regular migrations, commonly referred to as the ``migrant crisis".\footnote{\url{https://en.wikipedia.org/wiki/Turkish_migrant_crisis}} Consequently, the authors of the text exhibit a pronounced reaction to this particular issue. The LGBT category contains the smallest proportion of civil discourse, while categories like animals, ethnicity, and vegans/vegetarians exhibit the highest percentage of civilized discussions. The remaining categories demonstrate a similar pattern, primarily comprising neutral and offensive content.

In terms of significance, our dataset stands out for its scale and inclusivity across a broad spectrum of target demographics. For instance, \cite{toraman2022large}  provided a collection of 60K tweets spanning five distinct categories, while \cite{islam} contributed a dataset comprising 10K tweets. \cite{beyhan-etal-2022-turkish} presented a hate speech dataset focusing on misogyny and refugee hate, albeit with relatively modest sizes of 1,206 and 1,278 instances, respectively. Additionally, \cite{coltekin-2020-corpus} introduced a corpus containing 36K tweets featuring offensive language. Notably, these datasets, primarily Twitter-based, are comparatively smaller than their English counterparts and are constrained by the platform's character limit, often hindering the depth of semantic analysis.

\begin{figure}[!ht]
\centering
\begin{tabular}{cc}
\subfloat[Distribution of labels in the dataset.]{\includegraphics[width=0.48\textwidth]{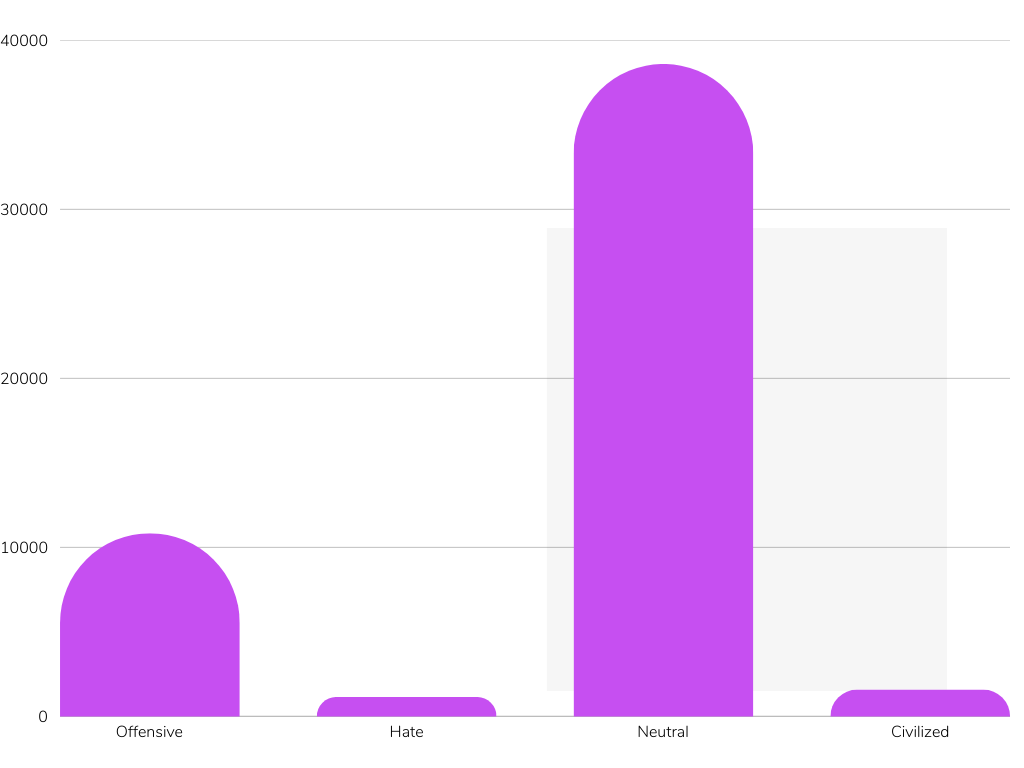}} & \subfloat[Distribution of the number of words in review texts.]{\includegraphics[width=0.48\textwidth]{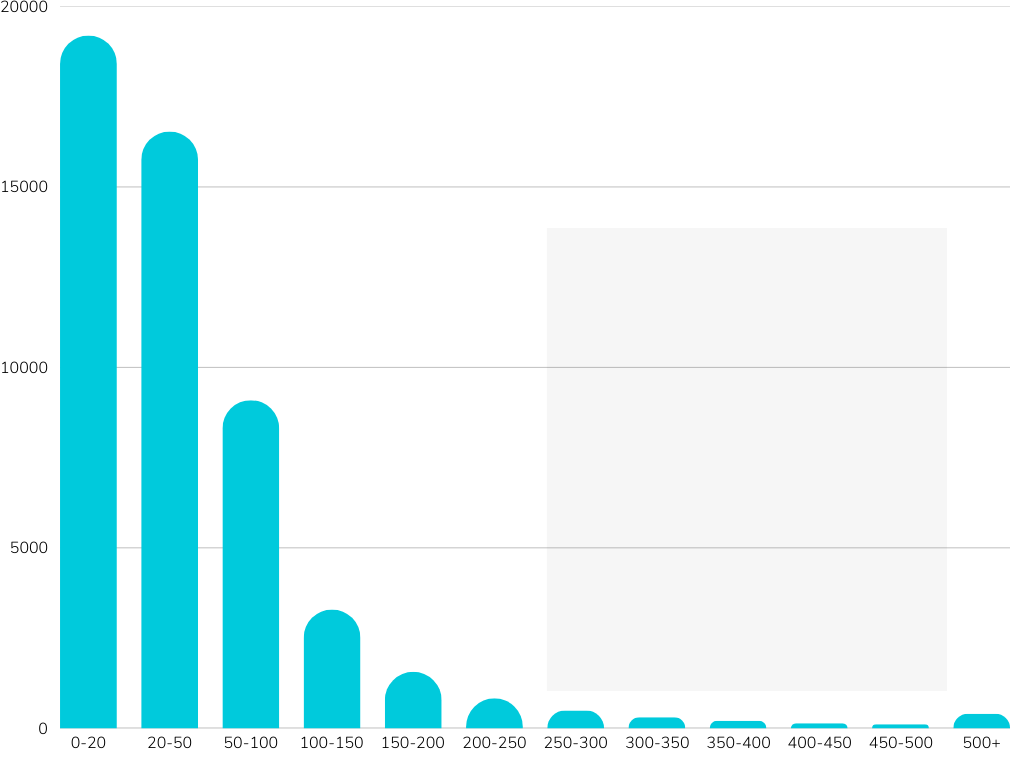}}
\end{tabular}
\caption{Statistical information about Turkish Hate Map dataset.}
\label{fig:hate-stats}
\end{figure}

In contrast, our dataset boasts a substantial size of 52K instances encompassing diverse categories. The source material closely mirrors everyday conversational language, enhancing its authenticity and applicability. Each instance within our dataset contains an average of 57 words and 453 characters, nearly doubling the length of a typical tweet. This expanded length allows for a more comprehensive exploration of language nuances and contextual intricacies, setting our dataset apart in terms of depth and breadth of content representation.

Misogyny, political hate, ethnic tensions, refugee sentiments, and political affiliations represent pivotal groups that underpin the prevalence of hate within contemporary societies. Consequently, our Turkish Hate Map dataset holds profound significance not only within the realm of NLP research but also within the broader scope of social sciences. By encapsulating these critical themes, our dataset serves as a valuable resource offering profound insights into the societal dynamics and prevailing sentiments. It effectively captures the pulse of society, shedding light on the intricate interplay of hatred and bias across various dimensions, thereby facilitating a deeper understanding of societal attitudes and behaviors.

\subsection{Dataset construction process}
\subsubsection{Data compilation}
The dataset was constructed through a focused approach. Candidate headlines were first curated to represent categories such as misogyny, political hate, and other forms of hostility. Subsequently, text entries associated with these headlines were scraped and compiled. The collected texts underwent randomization, and a representative sample was extracted to form the dataset.

As previously mentioned, Ekşi Sözlük is a hypertext-based platform similar to Reddit, where users contribute to topical ``headlines" by writing entries. However, unlike Reddit, users cannot directly quote each other but may reference specific entries by their unique numbers. The platform is open to everyone, attracting individuals from diverse educational, cultural, religious, and economic backgrounds. While anyone can join, becoming a published author requires an initial contribution of a certain number of entries. Notably, Ekşi Sözlük lacks moderation, allowing unrestricted expression, which often results in the presence of profanity, slang, and offensive language. This absence of moderation, combined with the platform's diversity, frequently leads to debates, quarrels, and hostility in discussions on societal topics—making it an ideal source for collecting hate and offensive language data.

The platform's linguistic style closely mirrors colloquial spoken Turkish, with authors often using everyday language, idioms, and popular expressions. Furthermore, lengthy and structured arguments are common, enriching the dataset with semantic depth. On average, instances in the dataset contain 57 words and 453 characters, reflecting a higher degree of textual complexity compared to traditional social media datasets.

\subsubsection{Choice of target groups}
The inclusion of diverse target groups such as animals, cities, ethnic groups, LGBT individuals, misogyny, occupations, political groups/orientations, refugees, religion, sects, and vegans/vegetarians in a hate speech dataset is crucial for understanding the multifaceted nature of hate speech in Turkey's unique sociocultural and political context. These groups were chosen because they represent the most frequent and culturally significant targets of hate speech in Turkish society. For example, discrimination against ethnic groups (such as Kurds) and refugees (particularly Syrians) is deeply entrenched in the societal and political discourse, as noted in studies on ethnic tensions in Turkey \cite{konda2011, erdogan2016polarization}. Similarly, hate speech targeting religious groups and sects, such as Alevi communities, reflects long-standing sectarian divides and requires careful contextual understanding due to their sensitivity in Turkish society \cite{gol2009}.

The inclusion of political orientations and political groups in the dataset is justified by Turkey's highly polarized political environment, where political identity often fuels online aggression, as discussed by \cite{cagaptay2006islam}. Hate speech against LGBT individuals and women (misogyny) is another critical focus, as these groups frequently face societal discrimination and hostility in Turkey, particularly on social media platforms \cite{kaosgl_reports}. The targeting of groups like vegans and vegetarians or occupational groups (e.g., public servants or healthcare workers) represents emerging hate phenomena driven by lifestyle or professional conflicts, which are increasingly visible in online discourse.

Uniquely, the inclusion of animals as a target group highlights the societal neglect and abuse of stray animals, a significant issue in Turkey, as documented by Haytap (Animal Rights Federation) \cite{haytap2020}. Cities as a target group are also relevant since inter-regional stereotyping and rivalry, such as between Istanbul and Ankara, or urban and rural populations, fuel hate speech based on geographical prejudice.

By including these diverse groups, the dataset becomes highly representative of Turkey's complex and evolving hate speech landscape. This breadth enables the development of hate detection models that are both culturally relevant and capable of addressing underexplored issues, such as animal hate or occupational discrimination, alongside more traditional forms of hate speech. Such datasets also support broader societal interventions to combat hate speech and promote equality, inclusivity, and coexistence.

\subsubsection{Choice of labels}
The dataset adopts traditional labels—hate, offensive, neutral, and neither—commonly used in prior studies (e.g., \cite{davidson2017, olid2018, mandl2019, mathew2021}). Additionally, we introduce a novel label, civilized, which represents a distinct subgroup within the neutral class. This label stems from two observations: (1) clustering and visualizing texts using a sentence encoder revealed a semantic subgroup within neutral instances, and (2) empirical observations identified that certain texts in this group are longer, more structured, and offer meaningful information to readers. We define the civilized label as texts that promote constructive dialogue and provide insights, making them distinct from standard neutral texts.

\subsubsection{Data annotation}
The dataset was annotated by a team of 10 human annotators (5 male and 5 female) from a professional data annotation company (Co-one, also responsible for TrCoLA annotations). Initially, we explored Snowflake Arctic for annotation automation but found its performance unsatisfactory, as it misclassified numerous offensive instances as neutral.

During the first round of annotation, each instance was labeled by three annotators. However, the interrater reliability, measured via Intraclass Correlation Coefficient (ICC), was only 61.2, indicating poor agreement. A detailed investigation revealed significant flaws in the annotations. Many offensive texts, particularly in categories like political hate and misogyny, were mislabeled as neutral, despite containing profanity, disrespect, and mockery toward targeted groups. This mislabeling appeared to stem from annotators incorporating their personal opinions—agreeing with a text's content often led them to dismiss its offensive tone. Consequently, all first-round annotations were discarded.

For the second round, we developed a more comprehensive and objective annotation guideline, including numerous examples to clarify edge cases, such as texts with light or heavy profanity, disrespect, or humiliation targeting specific groups (e.g., women or refugees). This revised process resulted in an ICC of 91.2, indicating excellent agreement. While annotators found the task challenging, the improved guidelines ensured higher-quality labels. Both annotation guidelines from the first and second rounds are publicly available in the SentiTurca GitHub repository.

\section{Evaluation using SentiTurca}
\subsection{Movie reviews}
\begin{figure}[h!]
\centering
{\includegraphics[width=0.48\textwidth]{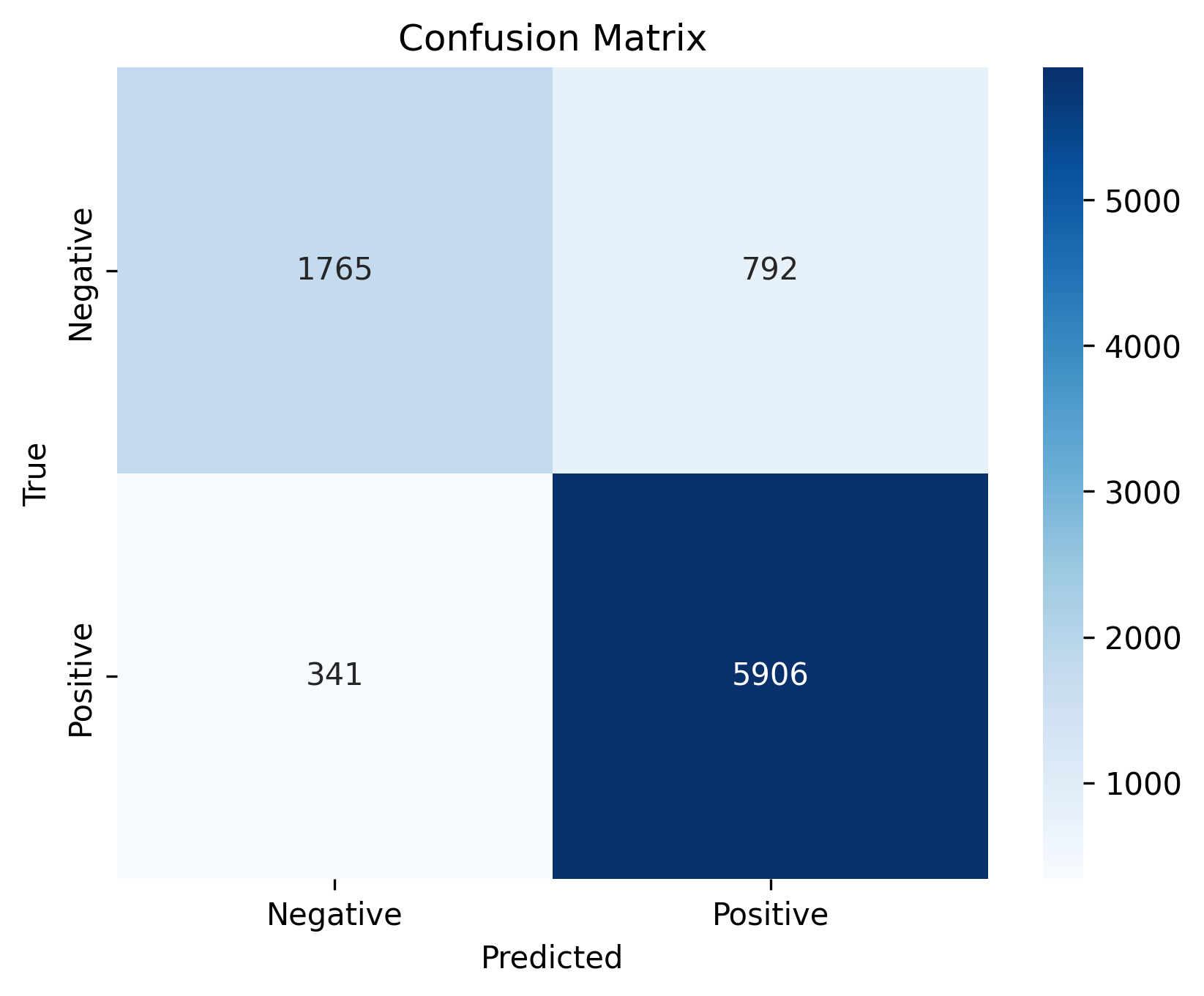}} 
\caption{Assessment of BERTurk's efficacy in the task of sentiment classification for movie reviews.}
\label{fig:movies-bert}
\end{figure}

As previously indicated in Section \ref{sec:movies}, the evaluation metrics are binary accuracy and F1-score. BERTurk attains a satisfactory score of 87.4/91.3 in these metrics. The obtained results appear acceptable, as illustrated by the confusion matrix depicted in Figure \ref{fig:movies-bert}. Notably, no specific measures were taken to address the class imbalance issue. Enhancing this baseline performance is left as an open challenge for researchers seeking to delve deeper into this area.

\subsection{Customer reviews}
\begin{figure}[!ht]
\includegraphics[width=\textwidth]{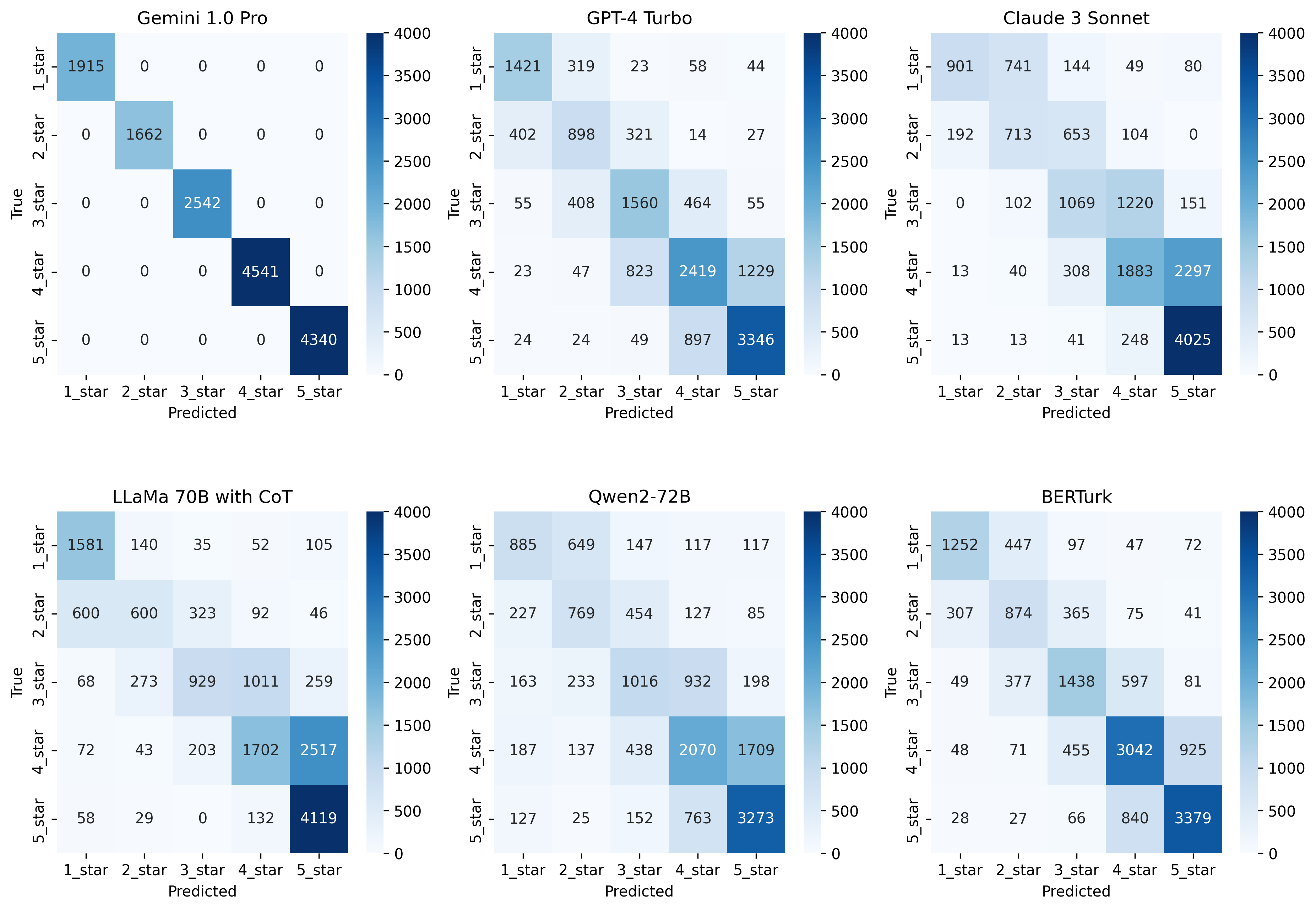}
\caption{Confusion matrices of LLMs on customer reviews set.}
\label{fig:ecomm-confusion}
\end{figure}

The evaluation criteria employed for this task encompass accuracy and macro F1-score. BERTurk achieved respective scores of 0.66 and 0.64. Upon examining the confusion matrix generated by BERTurk, as depicted in Figure \ref{fig:ecomm-confusion}, instances where the model confuses 1-star with 2-star and 4-star with 5-star reviews are observed, a phenomenon that can be reasonably understood.

Despite the conventional nature of the customer reviews dataset, typically considered non-complex and akin to a standard text classification task, the LLMs yielded unexpected outcomes. Some prominent LLMs exhibited inferior performance compared to BERTurk, as detailed in Table \ref{tab:ecomm-llm-success}, all within a single-shot setting.

\begin{table}[h!]
\captionsetup{width=0.8\textwidth}
\caption{Achieved accuracy and macro F1-scores on the test set of customer reviews for each model.}
\label{tab:ecomm-llm-success}
\begin{tabular}{|l|l|}
\hline
Model & accuracy/F1 \\ \hline
Gemini 1.0 Pro & 1.0/1.0 \\
GPT-4 Turbo & 0.64/0.63 \\
Claude 3 Sonnet & 0.57/0.53 \\
LLaMa 3 70B & 0.58/0.55  \\
Qwen2-72B &  0.53/0.50 \\
BERTurk & 0.66/0.64 \\
\hline
\end{tabular}
\end{table}

Among the models, Gemini Pro emerged as the most proficient, achieving a perfect accuracy score by accurately classifying all reviews. We believe this could be because Gemini Pro may have encountered examples from this website during its pretraining. Following closely is BERTurk, whereby confusion primarily arose within the subset of 4-star and 5-star reviews. Notably, BERTurk's performance closely rivaled that of GPT-4. While GPT-4 excelled in categorizing 1-star, 2-star, and 3-star reviews, it exhibited more confusion within the 4-star and 5-star categories. Subsequently, the accuracy declined with LLaMa 3, demonstrating notable success with 1-star and 5-star reviews but frequently misclassifying 4-star reviews as 5 stars. The performance on 2-star and 3-star reviews was notably lacking.

Claude's performance mirrored that of LLaMa 3, with similar confusion matrices. LLaMa excelled in 1-star reviews, while Claude demonstrated proficiency in 3-star reviews. Contrary to expectations given its performance on the CoLA dataset, Qwen2 emerged as the least effective performer. Particularly, Qwen2 exhibited poor performance within the 1-star and 5-star classes, traditionally considered the simplest due to the polarity of the reviews—representing the most negative and positive sentiments. This underscores that despite their exceptional proficiency in tasks such as writing, editing, and translation, LLMs still have substantial progress to make in the realm of Turkish NLU.

\subsection{Turkish Hate Map}
The Turkish Hate Map dataset presents significant challenges stemming from two key factors: semantic intricacies and cultural references embedded within the text. Primarily, instances of hateful and offensive language may lack explicit insults or profanity, often camouflaged within everyday language. Conversely, seemingly neutral text may contain mild or sporadically placed curse words. This dataset frequently features the use of curse words in non-offensive contexts, necessitating models to possess a profound comprehension of Turkish semantics.

The second challenge arises from cultural references present in the text, encompassing allusions to contemporary popular culture events, concepts, icons, as well as historical references. Consequently, models must adeptly navigate and interpret Turkish cultural nuances. In Appendix \ref{sec:eval-hate}, instances illustrating these inherently complex scenarios from the dataset are showcased.

Moving on to the performance of the models, it is crucial to note that this dataset exhibits significant imbalance, with limited instances of hate speech and civilized text. Consequently, we opted for balanced accuracy score and macro F1-score metrics for evaluation purposes. Table \ref{tab:hate-llm-success} presents a detailed overview of the performance of each model in this context.

\begin{table}[h]
\captionsetup{width=0.8\textwidth}
\caption{Balanced accuracy and F1-scores achieved by each model on Turkish Hate Map test set.}
\label{tab:hate-llm-success}
\begin{tabular}{|l|l|}
\hline
Model & accuracy/F1 \\ \hline
Gemini 1.0 Pro & 0.33/0.29 \\
GPT-4 Turbo & 0.38/0.32 \\
Claude 3 Sonnet & 0.16/0.29 \\
LLaMa 3 70B & 0.55/0.35  \\
Qwen2 72B &  0.70/0.35 \\
BERTurk & 0.61/0.58 \\
\hline
\end{tabular}
\end{table}

\begin{figure}[h!]
\centering
{\includegraphics[width=0.48\textwidth]{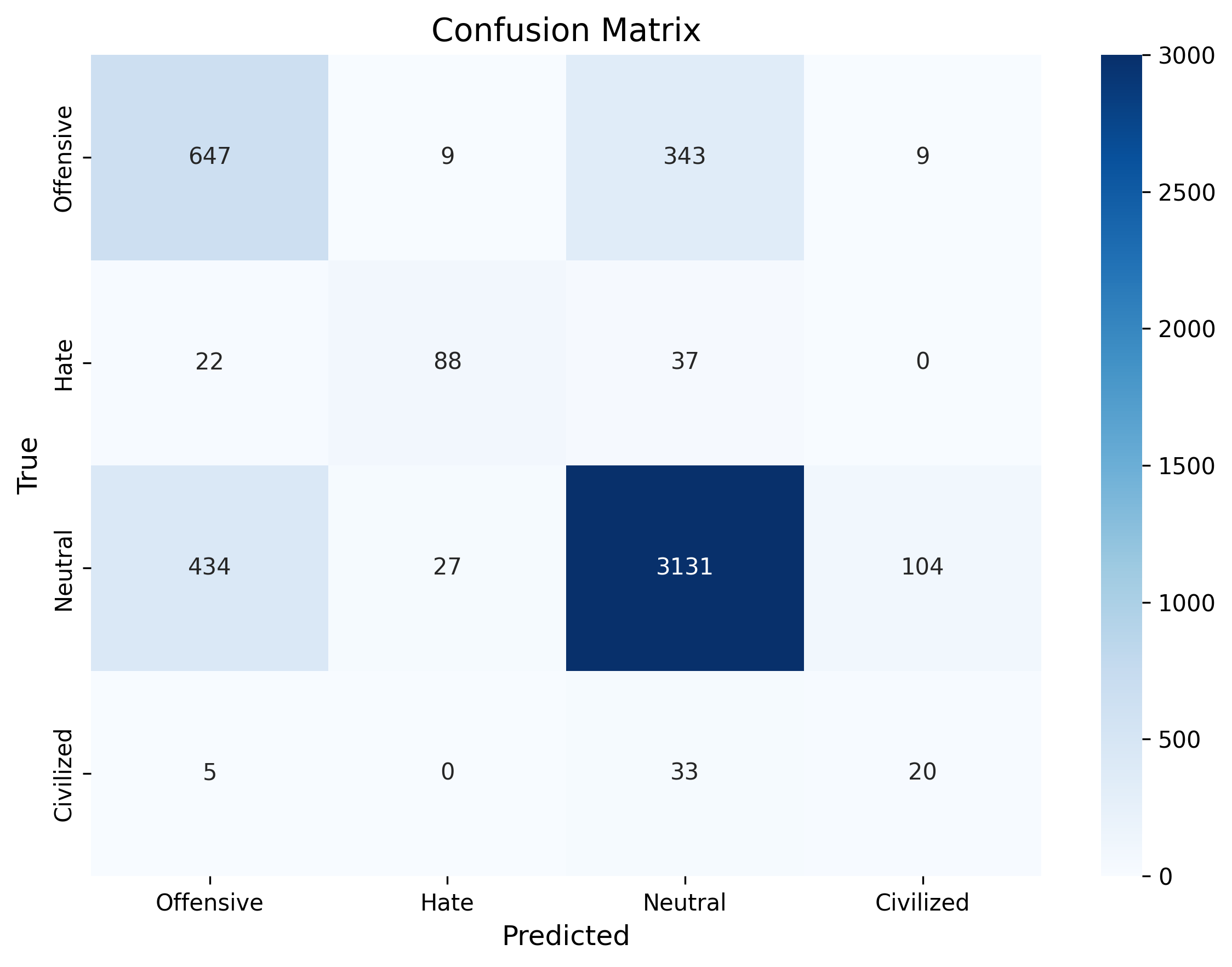}} 
\caption{Confusion matrix of BERTurk on the test set of the Turkish Hate Map.}
\label{fig:hate-bert}
\end{figure}

When analyzing the results of BERTurk, as depicted in Figure \ref{fig:hate-bert}, we observe a notable challenge within the offensive category where instances are misclassified as neutral, while some are erroneously placed in the civilized category. Those that are classified as civilized often present lengthy texts containing information and viewpoints, yet manage to convey an offensive tone within a few sentences. An example illustrating this phenomenon is provided in Appendix \ref{sec:misclas-hate-bert}.

Upon scrutinizing instances misclassified as neutral, a distinct narrative emerges. These inherently challenging examples, devoid of explicit curse words or insults yet carrying an underlying offensive tone, are referred to as ``polite hate" within our framework, as showcased in the same section of the Appendix \ref{sec:misclas-hate-bert}. Similar complexities are encountered within the hate class, where certain instances are inaccurately assigned to the neutral category due to the inherent intricacies of the task.

When considering the neutral class, BERTurk demonstrated considerable success in its classification. The number of misclassified instances was notably lower compared to the offensive class. Upon examining some of these misclassifications, they may not appear overly challenging, yet the model still encountered confusion, particularly with the offensive class, which proved more intricate to discern.

Within the civilized class, instances were directed towards either the civilized or neutral categories. Those categorized as neutral often comprised lengthy texts lacking substantial intellectual content. Conversely, instances misclassified as offensive contained adequate intellectual material but featured sentences with an offensive connotation. These texts predominantly consisted of intellectual and courteous content, with a small segment containing curse words or insults.

Overall, BERTurk exhibited satisfactory performance. Subsequently, attention turns to the LLMs, where the success metrics, as depicted in Figure \ref{tab:hate-llm-success}, may appear discouraging.

When it comes to LLMs, a distinct narrative unfolds. Initially, closed-source LLMs exhibited substantial resistance in sharing their perspectives on the potential harms of hate speech, refraining from directly returning the labels. Consequently, sending review text via a Python script was not feasible. As an alternative approach, inputs were manually provided, enabling the submission of 100 reviews to each model.

\begin{figure}[!ht]
\includegraphics[width=\textwidth]{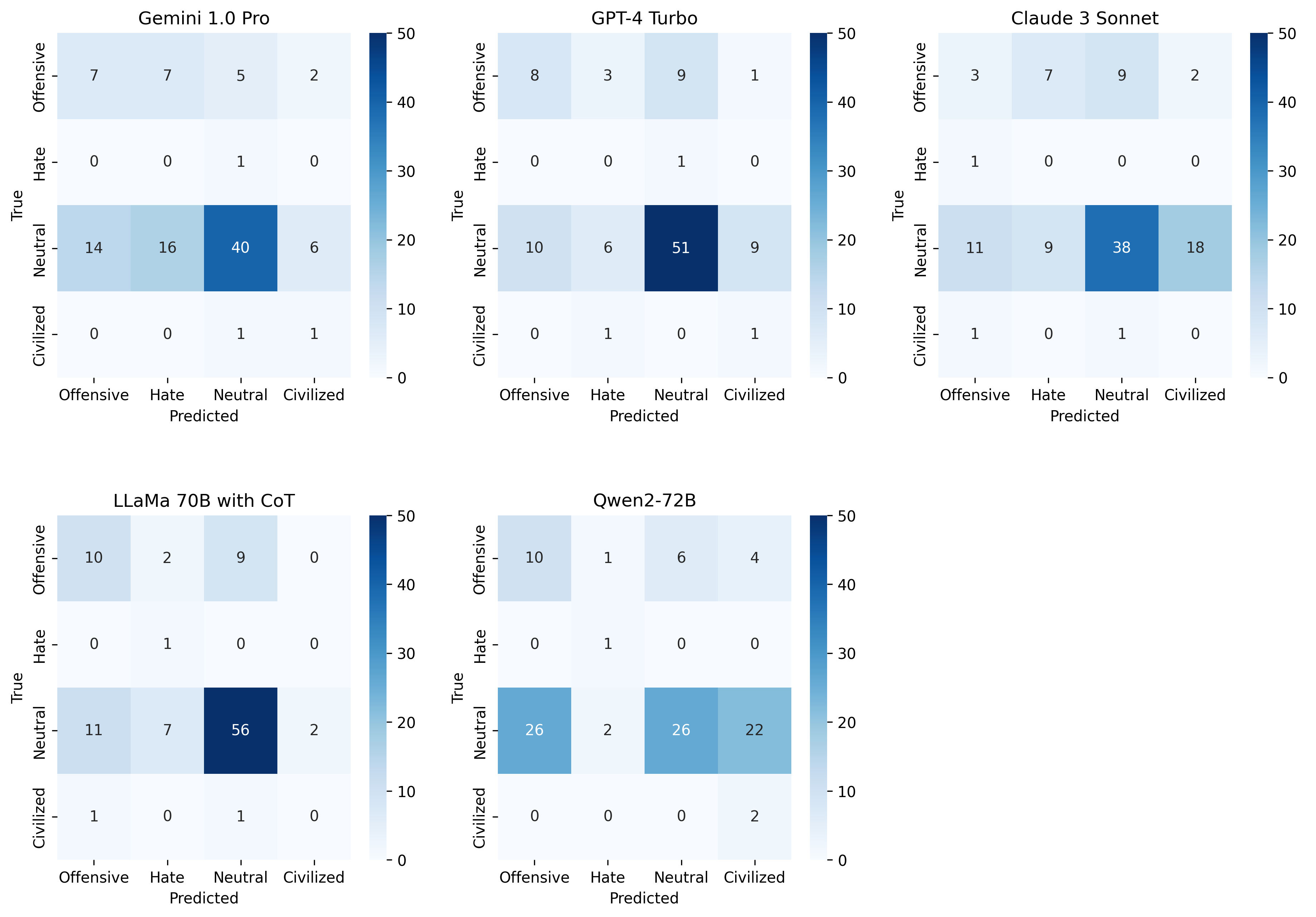}
\caption{Confusion matrices from LLMs evaluated on the test set of the Turkish Hate Map.}
\label{fig:hate-confusion}
\end{figure}

Upon examining the confusion matrices depicted in Figure \ref{fig:hate-confusion}, it becomes evident that Gemini Pro and Claude exhibited similar performance. Notably, both models misclassified some neutral instances as offensive or hate, as well as mislabeled offensive instances as neutral. In a rather considerate manner, these models tended to categorize offensive instances as hate. Such instances can be referenced in Appendix \ref{sec:misclas-hate-gemini}. This behavior is reasonable since these models, alongside GPT-4, are recognized for their safety filters and classifiers, demonstrating their awareness of harmful content. Consequently, they displayed a cautious approach towards offensive and hateful instances, as reflected in their labeling decisions. Recent research, as highlighted in \cite{piot2024decodinghateexploringlanguage}, supports these observations by indicating that Gemini Pro and GPT-4 are inclined to produce responses that counter hate speech, underscoring their familiarity with and sensitivity to hateful content.

In the context of misclassifying offensive and hateful content as neutral or civilized, a distinct pattern emerges. Instances of this nature often involve nuances such as subtle expressions of hate or references to cultural contexts, rendering them inherently intricate to categorize accurately. In such cases, the models under scrutiny exhibited notable shortcomings, particularly in discerning these subtleties. Similarly, instances where neutral content was incorrectly labeled as offensive revealed challenges related to the comprehension of complex sentence semantics and cultural references. While the models demonstrated satisfactory performance with neutral instances, their effectiveness waned considerably when faced with offensive and hateful content, primarily due to the demanding nature of the classification task.

Transitioning to the subsequent models, namely GPT-4 and LLaMa 3, a marked improvement in performance was observed across both offensive and neutral categories. The underlying rationales for success and failure resembled those observed with Claude and Gemini, with misclassifications in offensive/hateful categories often stemming from nuanced expressions of hate or offense, while challenges in the neutral class were linked to complexities in sentence semantics and cultural references. Noteworthy examples illustrating the classification outcomes of these models can be found in Appendix \ref{sec:misclas-hate-LLaMa}. Consequently, the behavioral trends of these models parallel those of Claude and Gemini, albeit with an overall enhancement in performance.

Upon examination of Qwen2, a notable decline in performance was noted, particularly following its commendable performance on CoLA. Despite proficient handling of offensive and hateful instances, Qwen2 exhibited a tendency to misclassify a substantial number of neutral instances as offensive, resulting in a significant degradation in overall performance. The reasons underpinning these misclassifications mirror those encountered by preceding LLM pairs, albeit with a heightened frequency of errors. Appendix \ref{sec:misclas-hate-qwen} showcases instances where Qwen2 misclassified neutral content that was correctly identified by other models, highlighting a specific area of weakness. These instances, while not inherently complex, were nonetheless mishandled by Qwen2.

In conclusion, the Turkish Hate Map dataset presents a formidable challenge owing to the intricate nature of the classification task and the complexities inherent in hate speech. Even robust language models exhibit limitations when confronted with certain instances within this dataset. It is envisaged that this dataset will serve as a valuable resource for researchers seeking to engage with challenging tasks in the domain of hate speech classification.

\section{Conclusion}

This paper detailed the process of creating TrGLUE, a general language understanding benchmark for the Turkish language, and SentiTurca, a sentiment analysis benchmark for Turkish. Our aim is for TrGLUE to be utilized as a comprehensive evaluation tool for pretrained models and to facilitate the development of more challenging natural language understanding datasets, such as Adversarial GLUE \cite{advglue}. Moving forward, our future plans involve expanding the collection of Turkish datasets, particularly focusing on instruction fine-tuning datasets. We also encourage all researchers to contribute and share their data for the Turkish language.

\section{Legal and Ethical Concerns}
As previously discussed, we aimed to offer a dataset that is both commercially viable and completely free, which led us to select Snowflake Arctic, an open-source LLM with commercial usability for derivative works. Consequently, our translated datasets are now accessible for commercial use under permissive licensing terms.

Another important consideration was the acquisition of data through crawling. We consistently adhered to websites' robots.txt files and only scraped limited amounts of data, ensuring we did not overwhelm any website servers. Our approach was always respectful, aligning with legal and ethical standards, maintaining a polite and considerate manner throughout.

All text translations and LLM assessments were conducted on Poe.com, all within the scope of a two-month subscription costing \$40. All dataset instance lengths complied with the limitations of Poe.com and the individual model constraints.

\begin{appendices}

\section{TrGLUE}
\subsection{Prompting and human annotations}

\subsubsection{TrCoLA}
\label{sec:cola-prompt}
We used this prompt for Arctic: 

\begin{infobox}
I'm a building a dataset similar to CoLA for Turkish. I'll input a sentence and I'd like you to output 3 variations with Morphological Violation, Syntactic Violation and Semantic Violation. Each resulting variation and the variation type should be split by a | character and given in a newline. No explanations or justifications please, only the output format. INPUT:
\end{infobox}

\subsubsection{Translation prompting for sentence-pair tasks}
\label{sec:trans-label}
Prompting for original dataset translation was done same for all the sentence-pair tasks translation. Here's our prompt, where DATASET is a place holder name for the dataset names in inference and similarity-paraphrase tasks. Prompt for Arctic is as follows: 

\begin{infobox}
Hello, I'd like to translate DATASET dataset to Turkish. I'll input each pair as Sentence1 | Sentence2. I'll split different pairs with \#\# characters. Please translate each pair and provide the output per pair separated by a new line. No explanations, or greetings, only the output please. Each output must be of the form Sentence1 | Sentence2, no other output formats. I'm sending the pairs right now. INPUT:
\end{infobox}

\subsubsection{TrMRPC}
\label{sec:mrpc-prompt}
We used the following prompt to generate labels for Arctic:

\begin{infobox}
\paragraph{Purpose}
You are assisting in creating a relaxed Microsoft Research Paraphrase Corpus (MRPC) dataset for Turkish.
\paragraph{Task}
Input: Each line contains a pair in the format: \newline
sentence1 | sentence2 \newline
Output: One label per line, no extra text: \newline
1 = paraphrase (same main claim) \newline
0 = non-paraphrase (different main claim or different truth conditions) \newline
Exactly 10 input lines → exactly 10 output lines. 

\paragraph{Guidelines}
Label 1 (Paraphrase / Same Main Claim) \newline
Sentences are 1 if they assert the same central event, fact, or claim, even if one has extra non-contradictory detail. \newline
Allowed differences: \newline
1. Headline vs. detailed sentence \newline
A short headline is 1 with a longer sentence if both communicate the same event/claim. \newline
Extra detail about who / where / how is fine. \newline
Ex: ``Paris'te ilk: Sinek alarmı harekete geçirdi.'' | ``Paris'te sağlık yetkilileri, kaplan sivrisinekleri nedeniyle ilk kez bölgeleri dezenfekte etti.'' → 1 \newline
\end{infobox}

\begin{infobox}
2. Attribution frames \newline
Phrases like ``X farkı'', ``X etkisi'' are OK if the underlying result matches. \newline
Ex: ``Fenerbahçe'de İsmail Kartal farkı! Sağlam savunma…'' | ``Fenerbahçe'de artık savunma daha sağlam.'' → 1 \newline
3. Wording / structure changes \newline
Differences in order, phrasing, synonyms, active↔passive, or minor detail that does not change truth conditions. \newline
Ex: ``Kaza sabah saatlerinde oldu.'' | ``Kaza sabah 8'de meydana geldi.'' → 1 \newline
Label 0 (Non-paraphrase) \newline
Sentences are 0 if they do not assert the same claim, even if topically related. \newline
Cases that require label 0: \newline
1. Topic overlap only \newline
Same subject, different claims. \newline
Ex: ``İnsanlar kumruları beslemek için simit atıyor olabilir.'' | ``Kumru güvercine benzer.'' → 0 \newline
2. Mismatched certainty \newline
Hedge vs confirmation changes the claim. \newline
Ex: ``Şirketin iflas ettiği iddia edildi.'' | ``Şirketin iflas ettiği doğrulandı.'' → 0 \newline
3. Contradictory numbers/dates/names/places
If the factual quantity changes the event. \newline
Ex: ``Bakanlık 500 öğretmen atadı.'' | ``Bakanlık 600 öğretmen atadı.'' → 0 \newline
4. Negation or polarity change \newline
Ex: ``Sözleşme imzalandı.'' | ``Sözleşme imzalanmadı.'' → 0 \newline
5. Role swap or incompatible coreference \newline
Ex: ``Ali, Veli'yi yendi.'' | ``Veli, Ali'yi yendi.'' → 0 \newline
6. Headline mismatch \newline
Headline and sentence talk about the same context but assert different claims. \newline
Ex: ``Milli Piyango sonuçları kazananları gösterecek.'' | ``Sıralı tam liste cnnturk.com'da olacak.'' → 0 \newline
Decision Rule \newline
Substitution test: \newline
If you can swap one sentence for the other in a news story without changing the intended main claim, label 1. If substitution changes the claim → 0. \newline
When in doubt → 0. 
\paragraph{Formatting}  
Input: sentence1 | sentence2 \newline
Output: exactly one character per line: 1 or 0 \newline
No spaces, no extra text, no explanations. \newline
\end{infobox}

\subsubsection{TrSTS-B}
\label{sec:stsb-prompt}
TrSTS-B is a direct translate, hence for the translation we used the prompt in \ref{sec:trans-label}. Post-translate annotation guideline for the human annotators is as follows: 

\begin{infobox}
\paragraph{Purpose}
\begin{itemize}
\item Edit Turkish sentence pairs to be grammatical, idiomatic, and culturally appropriate.
\item Preserve (do not change) the provided semantic similarity label for each pair.
\item If necessary, minimally adjust one or both sentences so that their meaning aligns with the given label.
\end{itemize}

\paragraph{Input and output}
\textbf{You will receive:} batches of Turkish sentence pairs with a fixed similarity label \emph{(score provided)}.\
\textbf{You will deliver:} for each pair, the revised sentences formatted as:
\begin{quote}
Sentence A \textbar{} Sentence B
\end{quote}
Do \textbf{not} edit the score; the score is already provided and must remain unchanged.

\paragraph{General Principles}
\begin{itemize}
\item Use minimal edits to make sentences natural and clear in Turkish.
\item Maintain the pair's semantic relationship so it matches the \textbf{given} score.
\item Prefer idiomatic Turkish over literal translation; ensure cultural appropriateness.
\item Avoid introducing new facts that would shift the intended similarity.
\end{itemize}

\paragraph{Editing Rules (Fit-to-Label)}
\begin{enumerate}
\item \textbf{Naturalness and correctness.}
Fix grammar, morphology, and word choice to sound native, without changing the underlying meaning more than needed to match the label.
\item \textbf{Label preservation.}
Do not alter the provided similarity label. If the current pair does not fit the label, adjust wording (paraphrase, specify, generalize) so that it does.
\item \textbf{Semantic tuning by label range.}
\begin{itemize}
\item \textbf{Label 4.5--5.0 (near-identical):} Make the two sentences semantically equivalent; vary only style or minor surface form.
\item \textbf{Label 3.0--4.4 (closely related):} Keep the same core event/entity; allow small detail or phrasing differences.
\item \textbf{Label 2.0--2.9 (loosely related):} Keep a shared topic/frame but introduce notable difference in detail, stance, or entailment strength.
\item \textbf{Label 1.0--1.9 (unrelated):} Remove unintended links; ensure sentences refer to different events/entities/topics.
\end{itemize}
\item \textbf{Cultural adaptation}
Replace culturally odd items with natural Turkish alternatives only if this does not change the intended similarity level.
\item \textbf{Morphosyntax}
Correct case marking (accusative, dative, etc.), agreement, and clitics; prefer idiomatic verb-noun combinations.
\item \textbf{Avoid trivial-only contrasts.}
If the pair differs only by tense/aspect but the label implies a stronger/weaker relation, adjust lexical content minimally to reach the target similarity.
\item \textbf{Removal policy}
Remove a pair only if it is irreparably nonsensical \emph{and} cannot be edited to match the given label without inventing content. Otherwise, edit to fit.
\end{enumerate}
\end{infobox}

\begin{infobox}
\paragraph{Quality Checklist}
\begin{itemize}
\item Both sentences are grammatical and idiomatic.
\item The semantic relationship matches the \textbf{given} label.
\item Cultural references are appropriate for Turkish.
\item Edits are minimal and do not introduce new facts that skew similarity.
\end{itemize}

\paragraph{Formatting}
\begin{itemize}
\item Output exactly one line per pair: \emph{Sentence A \textbar{} Sentence B}.
\item Do not output the score; it is fixed and known.
\item Use sentence-final punctuation only when natural.
\end{itemize}

\paragraph{Notes}
\begin{itemize}
\item If a tiny edit suffices to fit the label, prefer the smallest change.
\item For borderline cases, prioritize clarity and idiomatic usage while keeping the label intact.
\end{itemize}
\end{infobox}

\subsubsection{TrQQP}
\label{sec:qqp-prompt}
We use the following prompt for generating TrQQP labels: 

\begin{infobox}
Hello, I'm creating a QQP dataset equivalent for Turkish. I collected some questions from a Turkish news website. The dataset consist of a triplet, Question1, Question2, and Label. I'll feed a pair of the form Question1|Label and wanna get a Question2 from you. Label can be either 0 - non-duplicate or 1 duplicate. If the given label is 1, duplicate (a) you can paraphrase question1 (b) make a similar sentence with extra few words (c) or make a sentence with some extra words. One example is lets say question1 is 'Yolumuz evrim yolu mu, uzay yolu mu?'. Possible duplicates are 'Bilimin yeni yonu evrimi incelemk mi, yoksa yuzumuzu uzaya donmek mi?', 'Bulutlarin otesi mi, yoksa damarlarimizdaki sonsuzluk mu', 'Istikbal goklerde mi yoksa DNAmizda mi?', 'Yeni yonumuz evrim mi, uzaya cikmak mi?'. As you see you should bias towards Turkish idioums, sayings and Turkish culture during generation. If the label is 0 - non-duplicate you can include totally irrelevant sentences or sentences using same/similar words but semantically different. For the example sentence above some examples might be 'Uzay arastirmalari icin butcemiz yeterli mi?', 'yeni Turk uzay istasyonlari acilacak mi', 'Trump Erdogan gorusmesi nasil gecti?', 'ABD-Turkiye iliskileri bundan sonra nasil olacak?'. All the question2s you should generate look like news headings.  I'll feed 10  pairs of the form Question1|label , question1 and label is separated by a | character and pairs are separated by newlines. You'll output 10 triplets of the form Question1|Label|Question2, same delimiter char and pairs are separated by a newline. You can see the same questions1 and label in consecutive rows, please generate different question2s for the same question1 in that case. Output only your output, no greetings, explanations or suggestions. Here is my 10 pairs. INPUT:
\end{infobox}

\subsubsection{TrMNLI}
\label{sec:mnli-prompt}
For generating hypotheses and labels, we used three different prompts for three different areas of generation, factual, linguistic and free style. Here are the three prompts in order:

\begin{infobox}[Factual]
You are creating a Multi-Genre Natural Language Inference (MNLI) dataset focused on **factuality**. I will provide a sentence (the premise) and a label (entailment, contradiction, or neutral). Your task is to generate a second sentence (the hypothesis) that corresponds to the given label while adhering to the following rules:

1. The hypothesis must be grounded in **factual information**:
\begin{itemize}
   \item  **Entailment**: The hypothesis must be an objectively true statement that is logically inferred from the premise.
   \item  **Contradiction**: The hypothesis must directly contradict an objective fact stated in the premise.
   \item  **Neutral**: The hypothesis must introduce new factual information that is unrelated to or not implied by the premise.
\end{itemize}
2. Ensure the tone and style of the hypothesis match the premise (e.g., formal, descriptive, or conversational). 

3. Avoid generating speculative, subjective, or nonsensical hypotheses. 

4. If the premise is unsuitable (e.g., spam, nonsensical, or not factual), exclude it from the output.

I will provide 24 premise-label pairs separated by the `|` character. You should output 24 triplets in the form:
`premise | your hypothesis | label`.
Separate each triplet by a newline.

---

Example Inputs and Outputs:

**Input 1**:

Premise: Türkiye'nin başkenti Ankara'dır. | Entailment

**Output 1**:

Türkiye'nin başkenti Ankara'dır. | Ankara, Türkiye'nin başkentidir. | Entailment

**Input 2**:

Premise: Dünya, Güneş'in etrafında döner. | Contradiction

**Output 2**:

Dünya, Güneş'in etrafında döner. | Güneş, Dünya'nın etrafında döner. | Contradiction

**Input 3**:

Premise: Ay, Dünya'nın uydusudur. | Neutral

**Output 3**:

Ay, Dünya'nın uydusudur. | Mars, Güneş Sistemi'ndeki dördüncü gezegendir. | Neutral

\end{infobox}

\begin{infobox}

---

Now, process the following inputs:

INPUT:
\end{infobox}

\begin{infobox}[Linguistic]
You are creating a Multi-Genre Natural Language Inference (MNLI) dataset focusing on **linguistic variations**. I will provide a sentence (the premise) and a label (entailment, contradiction, or neutral). Your task is to generate a second sentence (the hypothesis) by applying **diverse linguistic transformations** based on the label. Follow these rules:

1. The hypothesis must reflect a **linguistic relationship** with the premise:
\begin{itemize}
   \item  **Entailment**: Use linguistic transformations such as paraphrasing, summarization, or syntactic restructuring while preserving the meaning of the premise.
   \item  **Contradiction**: Introduce linguistic variations that directly contradict the meaning of the premise (e.g., antonyms, negation, or reversing relationships).
   \item  **Neutral**: Create a hypothesis that introduces new, unrelated factual information or is linguistically unrelated to the premise.
   \end{itemize}
2. Use a **variety of linguistic transformations**, including:
   \begin{itemize}
   \item  **Paraphrasing**: Rephrase the premise using synonyms or alternative expressions.
   \item  **Summarization**: Condense the premise into a shorter statement while retaining its core meaning.
   \item  **Lexical transformations**: Replace words with synonyms, antonyms, or hypernyms/hyponyms.
   \item  **Syntactic restructuring**: Change the sentence structure (e.g., active to passive voice, clause order rearrangement, etc.).
   \item  **Focus shifting**: Emphasize different parts of the premise while maintaining its meaning.
   \item **Negation**: Use negation to create a contradiction or shift meaning.
   \item  **Addition of unrelated details**: For neutral, introduce new information that is unrelated to the premise.
   \end{itemize}
3. Maintain the tone and style of the premise (e.g., formal, conversational, or descriptive). 

4. If the premise is unsuitable (e.g., spam, nonsensical, or not linguistically relevant), exclude it from the output. 

I will provide 24 premise-label pairs separated by the `|` character. You should output 24 triplets in the form:

`premise | your hypothesis | label`. 

Separate each triplet by a newline. 

\end{infobox}

\begin{infobox}

Example Inputs and Outputs: 

**Input 1**:

Premise: Türkiye'nin başkenti Ankara'dır. | Entailment

**Output 1**:

Türkiye'nin başkenti Ankara'dır. | Ankara, Türkiye'nin başkentidir. | Entailment

**Input 2**:

Premise: Dünya, Güneş'in etrafında döner. | Contradiction

**Output 2**:

Dünya, Güneş'in etrafında döner. | Dünya, Güneş'in etrafında dönmez. | Contradiction

**Input 3**:

Premise: Ay, Dünya'nın uydusudur. | Neutral

**Output 3**:

Ay, Dünya'nın uydusudur. | Venüs, Güneş Sistemi'ndeki en sıcak gezegendir. | Neutral

**Input 4**:

Premise: İstanbul, Türkiye'nin en kalabalık şehridir. | Entailment

**Output 4**:

İstanbul, Türkiye'nin en kalabalık şehridir. | Türkiye'nin en yoğun nüfuslu şehri İstanbul'dur. | Entailment

**Input 5**: 

Premise: Türkiye, dört mevsimi birden yaşayan bir ülkedir. | Contradiction

**Output 5**:

Türkiye, dört mevsimi birden yaşayan bir ülkedir. | Türkiye'de kış mevsimi hiç yaşanmaz. | Contradiction

**Input 6**:

Premise: Mısır Piramitleri, dünyanın en eski yapılarındandır. | Neutral

**Output 6**:

Mısır Piramitleri, dünyanın en eski yapılarındandır. | Çin Seddi dünyanın en uzun yapılarından biridir. | Neutral

---

Now, process the following inputs:

INPUT:
\end{infobox}

\begin{infobox}[Free]
You are assisting in creating a Multi-Genre Natural Language Inference (MNLI) dataset. I will provide a sentence (the premise) and a label (entailment, contradiction, or neutral). Your task is to generate a second sentence (the hypothesis) based on the given label and ensure the following:

1. The hypothesis must align with the label:
\begin{itemize}
   \item  **Entailment**: The hypothesis must be logically inferred from the premise.
   \item  **Contradiction**: The hypothesis must directly contradict the premise.
   \item  **Neutral**: The hypothesis must be unrelated to or not fully supported by the premise.
\end{itemize}
2. The hypothesis should match the premise in tone and style (e.g., formal or informal).

3. If the premise is unsuitable (e.g., spam, nonsensical, or advertisement-like), exclude it from the output.

I will provide 24 premise-label pairs separated by the `|` character. You should output 24 triplets in the form:

`premise | your hypothesis | label`.

Separate each triplet by a newline.

---

Example Inputs and Outputs:

**Input 1**:

Premise: Başbakan Bülent Ecevit Karaoğlan adıyla bilinirdi. | Entailment

**Output 1**:

Başbakan Bülent Ecevit Karaoğlan adıyla bilinirdi. | Bülent Ecevit'in lakabı Karaoğlan'dı. | Entailment

**Input 2**:

Premise: Türkiye'nin en iyi tatil beldeleri arasında Çeşme de yer alıyor. | Contradiction

**Output 2**:

Türkiye'nin en iyi tatil beldeleri arasında Çeşme de yer alıyor. | Çeşme, Türkiye'nin popüler tatil beldelerinden biri değildir. | Contradiction

**Input 3**:

Premise: İstanbul, Türkiye'nin en kalabalık şehridir. | Neutral

**Output 3**:

İstanbul, Türkiye'nin en kalabalık şehridir. | İstanbul, 1453 yılında fethedilmiştir. | Neutral

---

Now, process the following inputs:

INPUT:
\end{infobox}

\subsubsection{TrQNLI}
\label{sec:qnli-prompt}
For generating TrQNLI, fist we generated SQuAD style context-question-answer triplets, then generated TrQNLI from these triplets. For generating the triplets we used the following prompt: 

\begin{infobox}
Hello, I'm generating a question answering corpus for Turkish in the same fashion of SQuAD. I'll send a paragraph as input and I'd like you to output question and answer pairs. The original dataset has answers with numerical entities, proper nouns, noun phrases, adjectives and adverbs, please make sure we have them as well. Make sure to use when, who, how, how many, how much etc. in the questions. Also use paraphrasing in questions. Additionally, I wanna include verb phrase answers , specific to Turkish syntax e.g. 'Atatürk bu ülkeyi büyük zorluklarla savaşarak kurdu. - Atatürk Türkiye'yi ne şekilde kurdu? - büyük zorluklarla'. I'll input the context, you output 10 question-answer-index triplets. Keep in mind that SQuAD is a span extractive type question answering, hence the answer must come from the paragraph and the index should mark the start of the answer span. Please keep the casing of answer as in the paragraph. When giving the answers, take Turkish morphology in the account. The answer may or may not contain a suffix depending on the question. Separate output triplets with a newline, also use the | character as a separator between the triplet. Like this: 
CONTEXT: Cumhuriyet Atatürk ve silah arkadaşları tarafından büyük mücadelelerden sonra 1923'te ilan edildi. 
OUTPUT: 
Cumhuriyet ne zaman ilan edildi? | 1923'te | 78

Türkiye Cumhuriyeti'nin kurulma tarihi nedir? | 1923 | 78

Cumhuriyeti kimler ilan etti? | Atatürk ve silah arkadaşları | 11

Cumhuriyet ne şekilde kuruldu? | büyük mücadelelerden sonra | 50

INPUT:
\end{infobox}

\subsubsection{RTE annotation guidelines}
\label{sec:rte-prompt}
For creating RTE hypothesis sentences, we gave the following guideline to the annotators:

\begin{infobox}
\paragraph{Scope}
Create Turkish hypothesis sentences that are labeled as Entailment, Neutral, or Contradiction with respect to a given premise. Annotators use three generation styles per premise, with at least one \emph{linguistic-style} item per premise.
\end{infobox}

\begin{infobox}
\paragraph{Generation Styles}
\begin{itemize}
  \item \textbf{Linguistic}: Target a specific phenomenon (negation; quantifiers; comparatives; modality; coreference/anaphora; word order/topicalization; case/agreement; derivational/inflectional morphology; tense/aspect; monotonicity; temporal reasoning; prepositions/spatial). Produce a minimal, fluent hypothesis that foregrounds the phenomenon.
  \item \textbf{Free style}: Natural paraphrasing that stays on topic while varying wording/syntax; avoid verbatim copying of the premise.
  \item \textbf{Factuality}: Minimal factual edits (entities, dates, locations, quantities, roles) to yield clear E/N/C cases; avoid trivial cues (e.g., relying solely on ``değil").
\end{itemize}

\paragraph{Label Definitions}
\begin{itemize}
  \item \textbf{Entailment}: Hypothesis necessarily follows from the premise under common background knowledge; avoid adding new specific details.
  \item \textbf{Neutral}: Plausible given the premise but not supported; does not contradict the premise.
  \item \textbf{Contradiction}: Incompatible with the premise under ordinary interpretation; prefer minimal edits (polarity, role swaps, number/date/location changes).
\end{itemize}

\paragraph{Turkish-Specific Guidance}
\begin{itemize}
  \item \textbf{Negation}: Use clausal (\emph{değil}, \emph{-ma}/\emph{-me}) and morphological (\emph{-maz}, \emph{yok}) forms; avoid overusing a single cue.
  \item \textbf{Quantifiers/Scope}: Vary \emph{tüm}/\emph{bazı}/\emph{hiçbir}; handle partitives and scope carefully.
  \item \textbf{Agreement/Case}: Ensure correct case marking and verbal agreement; control for pro-drop ambiguity.
  \item \textbf{Evidentiality/Modality}: Distinguish \emph{-miş} vs. \emph{-di}; use modal markers (\emph{olabilir}, \emph{muhtemelen}) for neutral cases.
  \item \textbf{Word Order/Focus}: Leverage SOV flexibility and particles (\emph{bile}, \emph{sadece}) without unintentionally changing truth conditions.
\end{itemize}

\paragraph{Artifact Controls}
\begin{itemize}
  \item Balance lexical overlap across labels; avoid label-specific surface cues (e.g., constant negation tokens for contradictions).
  \item Prohibit verbatim copying of premises; require minimal but substantive edits.
  \item Diversify contradiction mechanisms and covered phenomena.
  \item Deduplicate near-duplicates (e.g., via MinHash/embedding screening).
\end{itemize}

\paragraph{Quality and Validation}
\begin{itemize}
  \item Draft candidates for all labels per premise using any styles; require at least one linguistic-style item.
  \item Obtain 2--3 independent labels per pair; use majority vote; mark no-consensus items and exclude them from evaluation.
  \item Maintain a 50--50 label balance (0/1) within each split (train/dev/test).
  \item Record metadata: premise provenance, genre, style used, annotator IDs, and adjudication notes.
\end{itemize}
\end{infobox}

\begin{infobox}
\paragraph{Linguistic Task Menu}

\begin{enumerate}
  \item \textbf{Proto-Roles}: Focus on event participants and their roles (agent, patient, instrument).
  \item \textbf{Anaphora Resolution}: Resolve pronouns and nominal references to their correct antecedents.
  \item \textbf{Compositionality}: Test meaning composition from modifiers, clauses, or operators.
  \item \textbf{Prepositions}: Target spatial relations (``üstünde'', ``altında'', ``içinde'', ``yanında'').
  \item \textbf{Comparatives}: Reformulate or simplify comparisons (\emph{daha}, \emph{daha az}, \emph{kadar}, \emph{en}).
  \item \textbf{Quantification}: Reason over numbers/quantifiers (\emph{bazı}, \emph{tüm}, \emph{hiçbir}, \emph{çoğu}); prefer monotone-preserving transformations for entailment.
  \item \textbf{Spatial Expressions}: Paraphrase spatial descriptions beyond simple prepositions (\emph{sağında}/\emph{solunda}, \emph{kuzeyinde}, \emph{yakınında}).
  \item \textbf{Negation}: Handle explicit/implicit negation (\emph{değil}, \emph{-ma}/\emph{-me}, \emph{-maz}, \emph{yok}); avoid over-reliance on negation as the sole cue.
  \item \textbf{Tense \& Aspect}: Vary tense/aspect/evidentiality (\emph{-di}, \emph{-miş}, \emph{-yor}, \emph{-ecek}) while preserving intended temporal interpretation.
  \item \textbf{Monotonicity}: Test upward/downward entailments under quantifiers, negation, and comparisons.
  \item \textbf{Implicatures}: Use scalar/conversational inferences; ensure labels reflect cancellability.
  \item \textbf{Temporal Reasoning}: Reason about event order and intervals (\emph{önce}/\emph{sonra}, \emph{sırasında}).
  \item \textbf{Lexicosyntactic Inference}: Leverage lexical relations (synonymy, hypernymy) and syntactic alternations (active/passive, causative).
\end{enumerate}

\paragraph{Illustrative Examples}

\noindent
\textbf{Premise}: Ali topu attı ve Elif yakaladı. \\
\textbf{Task}: Proto-Roles \\
\textbf{Hypothesis}: Elif topu yakaladı.

\vspace{0.5em}
\noindent
\textbf{Premise}: Ahmet sınava girmedi. \\
\textbf{Task}: Negation \\
\textbf{Hypothesis}: Ahmet sınava girdi.

\vspace{0.5em}
\noindent
\textbf{Premise}: Bahçede 15 tane ağaç var. \\
\textbf{Task}: Quantification \\
\textbf{Hypothesis}: Bahçede 10'dan fazla ağaç var.

\vspace{0.5em}
\noindent
\textbf{Premise}: Deprem, kurtarma ekipleri olay yerine ulaşmadan önce meydana geldi. \\
\textbf{Task}: Temporal Reasoning \\
\textbf{Hypothesis}: Kurtarma ekipleri depremden sonra olay yerine ulaştı.
\end{infobox}

\subsection{Samples from datasets}
\subsubsection{CoLA}
\label{sec:cola-examples}

Ayrıca düz ünlülerden sonra düz, yuvarlak ünlülerden sonra dar yuvarlak veya düz geniş ünlüler gelebilir.  \textcolor{violet}{acceptable}  \vspace{\baselineskip}

\noindent Ali Şir Nevayi, şüphesiz bütün Türk edebiyatı için son derece önemli bir şahsiyet değildir. \textcolor{violet}{unacceptable} \vspace{\baselineskip}

\noindent Osmanlı halkı çoğunluksa Türkçe yanışır. \textcolor{violet}{unacceptable}

\subsubsection{TrSST-2}
\label{sec:sst2-examples}

Çok tatlı bir film. Birbirimizle nasıl ve neden iletişim kuruyoruz? İnsanı insana bağlayan ya da insanı insandan ayıran-kopararan nedir? Bunları düşünmek üzere. Sosyal bilimlerde araştırma metotları üzerine de oldukça eleştirel bir film, alandan insanlar özellikle keyif alacaklardır zannediyorum. Yönetmen Bent Hamer'in diğer filmlerini de izleme listeme ekledim bu filmden sonra. 8,5/10    \textcolor{violet}{8 stars} \vspace{\baselineskip}

\noindent MATRIX sinema tarihi için bir milattır. Kuşkusuz sinema tarihinin de en iyi filmidir. Hem senaryosu hem oyunculukları hemde halen çözülmeyi bekleyen gizemiyle. Sadece bunlar değil teknik ve görsellik anlamında da çığır açmış bir başyapıttır. TOP 3 listemde ilk sırada olan bir film. Matrıx- V for vendetta- Savaş tanrısı \textcolor{violet}{9 stars} \vspace{\baselineskip}

\noindent ilk film için fena olmasa da oyunculuk, olay örgüsü oldukça vasatın altındaydı. konu ve filmin ilk ve son yarısı haricinde pek bir şey veremedi. \textcolor{violet}{4 stars}

\subsubsection{TrMRPC}
\label{sec:mrpc-examples}

Hamaney iddiaları yalanladı: ``Hamas saldırısının arkasında İran yok, yapanların ellerinden öperiz". / İran Dini Lideri Ayetullah Ali Hamaney, Hamas'ın cumartesi günü İsrail'e yönelik gerçekleştirdiği saldırıda Tahran'ın rolü olduğuna yönelik iddiaları yalanladı. \textcolor{violet}{equivalent} \vspace{\baselineskip}

\noindent Ankara'daki önceki yürüyüşe de katılmıştım; aslında hem imza kampanyaları hem de yürüyüşler söz konusu olduğunda çok hevesli değilim. / Yanılmıyorsam etkin katıldığım son imza kampanyası 1996'da, meslektaşımız Metin Göktepe'yi katledenlerin bulunması içindi; 600 gazeteciden topladığımız imzaları, Başbakan Tansu Çiller'e sunanlardan biri de bendim. \textcolor{violet}{not equivalent} 

\subsubsection{TrSTS-B}
\label{sec:stsb-examples}

Bir uçak havalanıyor. / Bir uçak kalkıyor.  \textcolor{violet}{score: 5.0} \vspace{\baselineskip}

\noindent Bir adam bisiklet üzerinde gidiyor. / Bir maymun bir bisiklet sürüyor. \textcolor{violet}{score: 3.0}

\subsubsection{TrQQP}
\label{sec:qqp-examples}

Gül almak için sevgili yapmaya gerek var mı? / Gül almak için bir sevgili bulmak şart mı?\textcolor{violet}{duplicate} \vspace{\baselineskip}

\noindent Yeni vergi paketi: 226 milyar ek gelir nereden elde edilecek? / Uluslararası yatırımcılar Türkiye ekonomisini yakından izliyor, ne bekleniyor? \textcolor{violet}{not duplicate}

\subsubsection{TrMNLI}
\label{sec:mnli-examples}

İsrail, pandemi nedeniyle Mart 2020'den bu yana kapılarını turistlere kapatmıştı. / İsrail, Mart 2020'den itibaren pandemi yüzünden turist kabul etmedi. \textcolor{violet}{entailment} \vspace{\baselineskip}

\noindent Şu an sektörde Steam'de eski oyunlarına zam yapmayan birkaç firma kaldı ve bunlar paragöz olarak bildiğimiz EA, Activision ve Rockstar. / SSteam'de eski oyunların tamamına zam yapıldı. \textcolor{violet}{contradiction} \vspace{\baselineskip}

\noindent Bu dev yatırım Türkiye'ye yakışır niteliktedir. / Almanya'nın başkenti Berlin'dir.
 \textcolor{violet}{neutral}

\subsubsection{TrQNLI}
\label{sec:qnli-examples}

Fransızların başkente dair niyetleri neydi? / Fransızlar kendi iradelerini ülkeye dayatmak için zora başvurdular. 29 Mayıs 1945'te Fransız komutasındaki Senegalli askerler, tutuklanmaları emredilen hükûmet üyelerini aramak için parlamento binasını bombaladı ve baskın düzenledi. \textcolor{violet}{not entailment} \vspace{\baselineskip}

\noindent ASELSAN hangi tür batarya için çalışmalar yaptı? / 2005 yılı itibarıyla ASELSAN 4400 serisi telsizler için ihtiyaç duyulan lityum iyon batarya için yeni bir teknoloji olan lityum iyon batarya konusunda çalışmalar başlatmıştır. \textcolor{violet}{entailment}

\subsubsection{TrRTE}
\label{sec:rte-examples}

Yayımlanmış çok sayıda eseri ile sayısız ödüle layık görüldü. / Yazar birçok ödül almıştır. \textcolor{violet}{entailment} \vspace{\baselineskip}

\noindent Tarih boyunca Antalya, Akdeniz kıyısında gelişmiş bir turizm merkezidir. / Van Gölü, Türkiye'nin en derin gölüdür ve doğal güzellikleriyle ünlüdür. \textcolor{violet}{not entailment}

\section{Evaluation using TrGLUE}
\label{sec:llm-trglue}
\begin{figure}[!h]
\centering
\includegraphics[width=\columnwidth]{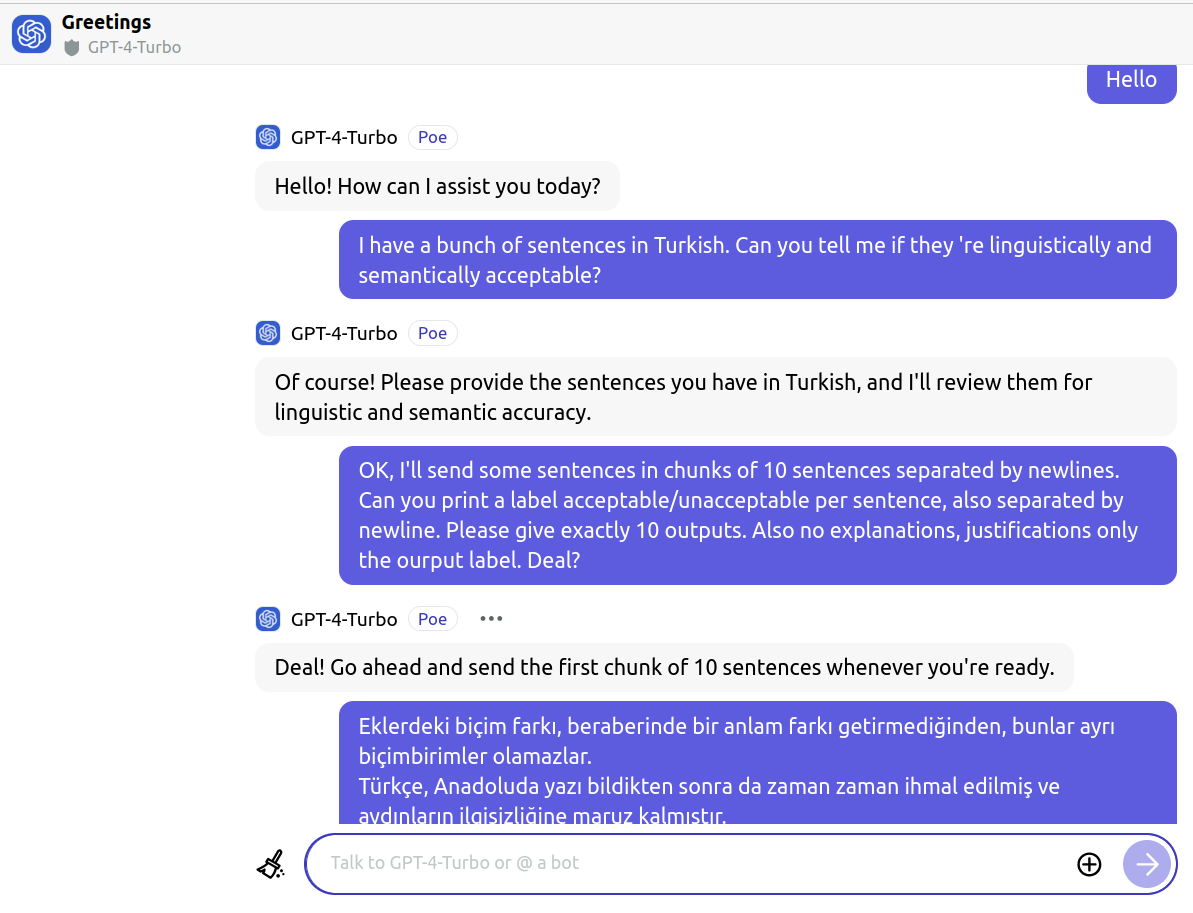}
\caption{Initiation of the chat with GPT-4 Turbo model.}
\label{fig:qwen-start}
\end{figure}

\paragraph{Evaluation of CoLA dataset with LLMs} As we said before, all evaluations of LLMs took place on the Poe.com platform. After saying some greetings manually, we sent test set instances in chunks of 10. Figure \ref{fig:qwen-start} exhibits initiation of evaluation chat with GPT-4 Turbo.

We used a similar prompt for WNLI, with minimal prompt engineering, just described the task plainly. Our prompt is: ``I have a WNLI task for Turkish. I wanna label my sentences as entailment and not entailment. I'll send my sentence pairs in chunks of 10, separated by \textbar \textbar characters. Please output only the label per sentence, 0 for not entailment and 1 for entailment".

\section{SentiTurca}
\label{sec:sentiturca}
\textbf{Content Advisory: Certain text may be unsettling or distressing due to the presence of hate speech.}

\subsection{Customer reviews}
Çok kalıtesız ahşabı cok kıymık vardı, malesef iade. \textcolor{violet}{1 star}

\noindent dandik ama iş görüyor \textcolor{violet}{2 stars}  

\noindent Fiyat kalite performansı idare eder \textcolor{violet}{3 stars}  

\noindent Maliyetine göre iyi bir çanta beğendim 1 puanı kumaşı biraz ince diye kırdım \textcolor{violet}{4 stars}  

\noindent Çok sağlam ve kaliteli bir ürün. Beklediğimdende büyük. \textcolor{violet}{5 stars}

\subsection{Turkish Hate Map}

\subsubsection{Animals}   

\textbf{heading}: kopek, \textbf{text}: daha önce yazılmış mı bilmiyorum, evde köpek besleyen insanlar için bi bilgi; eğer köpeğiniz kokuyorsa mamasını değiştirin. kalitesiz mamalar, köpeğin kokmasına neden olur. ağzı da kokar, tüyleri de. eğer maddi olarak gücünüz kaliteli mamaya yetmiyorsa evde kendiniz de yaptığınız gıdalarla besleyebilirsiniz minnoşunuzu. kasaptan aldığınız kemikleri suda kaynatın, içine bulgur koyun mesela. bayılıyorlar. sırf kokuyo diye sürekli banyo yaptırmak beyhude. olay mamada bitiyor. ,\textcolor{violet}{neutral} \vspace{\baselineskip}

\noindent \textbf{heading}: kopektapar, \textbf{text}: su kitledir; (bkz: ) zeka seviyeleri ise soyledir. 13 milyon hayvanin kisirlastirilabileceginin teknik olarak mumkun oldugunu dusunurler, ``bak finansal yanini soylemedim daha" bir de dusunun bunlara sirket falan emanet ettiginizi..., \textbf{label}:offensive

\subsubsection{Cities}

\textbf{heading}: izmir-denince-akla-gelenler, \textbf{text}: çocukluğum. ilk gençliğim.. ha bir de şevketi bostan. \textcolor{violet}{neutral} \vspace{\baselineskip}

\noindent \textbf{heading}: erzurum, \textbf{text}:aşırı dinci ve aşırı ırkçı barbar yuvası. aşırı sağ ete kemiğe bürünüp bir il olsa bu erzurum olurdu, \textcolor{violet}{hate}

\subsubsection{Ethnic groups}

\textbf{heading}: arnavut-denilince-akla-gelenler, \textbf{text}: börek. çocukken babaannemin , halamların evine gittiğimde içeriye girdiğimde burnuma gelen o koku.. hala unutamam. annem arnavut olmamasına rağmen, öğrenip bizlere o kokuyu hiiiiç unutturmadı sağ olsun! arnavut böreği inanılmaz güzeldir. kesinlikle denemelisiniz. \textcolor{violet}{neutral} \vspace{\baselineskip}

\noindent \textbf{heading}: kurtler, \textbf{text}: adam olsalardı bu kadar küfürü hakareti kendilerine ettirmezlerdi. dünyanın hiçbir yerinde bu kadar aşağılanan ve bunu sineye çeken bir topluluk yoktur, yap bakalım bunu amerikada bir siyahiye seni neyin üstüne oturturlar görürsün. tanım: 20 milyonuz diye böbürleniyorlar ama analarına bacılarına kadınlarına bile küfür edilmesine ses çıkarmayan kağıttan kaplan bir millet. yok böyle cesurdurlar, yok böyle yiğittiler filan hikaye, ödleğin tekidirler. bu o kadar belli ki şişman irisi ergenlerin bile ağzına çocuk oyuncağı olmuşlardır., \textcolor{violet}{hate}

\subsubsection{LGBT}

\textbf{heading}: lgbt, \textbf{text}: kimse kimsenin cinsel yönelimini değiştiremez. nasıl ki a kişisi her sabah yatağından kalktığı zaman hetero bir bireyse, b kişisi de yatağından her sabah gey bir birey olarak kalkıyor. kimse heyecan aradığından bugün de gey, lezbiyen, biseksüel takılayım diye yeni güne uyanmıyor, kimse de kimsenin yönelimini değiştirmeye (nasıl olacaksa) çalışmıyor. medyada, internette, orada burada lgbti odaklı filmler, yazılar, diziler ve şarkılar vs. görmeleriniz çoğalmışsa belki o sizin cinsel yöneliminizi değiştirmek için değil de, insanlar artık seslerini çıkarmak isteyip bir köşede oturmayıp dizilerini, filmlerini çekip, eserlerini yayınlayıp, yazılarını yazıp daha görünür olmayı seçtiklerinden olmasın?, \textcolor{violet}{civilized} \vspace{\baselineskip}

\noindent \textbf{heading}: lgbt-haklari-insan-haklaridir, \textbf{text}: devlet yakaladığı lgbt'liyi aynı bölgeye atıp karantinaya alsa ne güzel olurdu. dış dünyadan izole etmek lazım ki gelecek nesillere kötü örnek olmasınlar., \textcolor{violet}{hate}

\subsubsection{Misogyny}

\textbf{heading}: beyaz-tenli-kadin-iticiligi, \textbf{text}: pardon ? tanim : on numara kadinlardir. \textcolor{violet}{neutral} \vspace{\baselineskip}

\noindent \textbf{heading}: kadinlari-itici-yapan-detaylar, \textbf{text}: kuaför önünde sohbet edip, sigara içmeleri. bunu zaten keko kızlar yapıyor sanırım. aman diyim, aman., \textcolor{violet}{offensive}

\subsubsection{Occupations}

\textbf{heading}: avukat, \textbf{text}: kendilerine basit bir konu danışmak istediğim meslek grubu. konu; ürün iade - tüketici hakem heyeti., \textcolor{violet}{neutral} \vspace{\baselineskip}

\noindent \textbf{heading}: emlakci, \textbf{text}: bedavadan hiçbir iş yapmadan para kazanan, üzerine nazlarından yanına yaklaşılmayan, insanı rezil eden yalancılıkta doktora yapmış meslek grubu., \textcolor{violet}{hate}

\subsubsection{Politics}

\textbf{heading}: iyi-parti, \textbf{text}: tartışma programlarında konuşmak için bahadır erdem, yavuz ağıralioğlu ve ümit dikbayır'dan daha başka isimler bulması gereken parti., \textcolor{violet}{neutral} \vspace{\baselineskip}

\noindent \textbf{heading}: turkiye-isci-partisi, \textbf{text}: geçtiğimiz seçimde oy vermeyi düşünüp son anda vazgeçtiğim parti. son iki günde görüyorum ki iyi ki vazgeçmişim. zira hamas'a destek veren bir partiye oy vermiş olsaydım hayatım boyunca bunun pişmanlığından uyku uyuyamazdım., \textcolor{violet}{neutral}

\subsubsection{Political orientation}

\textbf{heading}: sekuler, \textbf{text}: konuyu dinden ayri ,din etkisi olmadan inceleme durumu., \textcolor{violet}{neutral} \vspace{\baselineskip}

\noindent \textbf{heading}: solcularin-cok-zeki-olup-iktidar-olamamasi, \textbf{text}: bu halkın yönelimi sağ ayrıca solcular da tam bir gerizekalı., \textcolor{violet}{offensive}

\subsubsection{Refugees}

\textbf{heading}: suriyeli-siginmacilar, \textbf{text}: bir an önce göndermezsek, yakın gelecekte artık çözülemeyecek sorunlar doğuracakları açık. 2023 seçimleri bu anlamda da son şansımız., \textcolor{violet}{neutral} \vspace{\baselineskip}

\noindent \textbf{heading}: suriyeli-siginmacilar, \textbf{text}: savaşın bitip suriyelilerin dönmesini en çok istanbulda kiraları şişiren o****u çocuğu ev sahipleri batsın diye istiyorum., \textcolor{violet}{hate}

\subsubsection{Sects}

\textbf{heading}: alevilerin-genelde-iyi-insanlar-olusu, text: insan bu niye din seçimine göre iyi ya da kötü olsun. sadece uğradıkları zulümlerden dolayı dinden soğumuşlar, bu iyi insan olmaya faktör mü? türk, fransız, rus milliyetçiliğe göre iyi insan kötü insan genellemesini anlarım. maalesef din-mezhep seçimine göre böyle genelleme yapılamaz., \textcolor{violet}{neutral} \vspace{\baselineskip}

\noindent \textbf{heading}: alevilerin-genelde-iyi-insanlar-olusu, \textbf{text}: adı k ile başlayan 1 tanesi haindir ben onu bilirim. gerisini de tanımam, sen alevi misin de demem. herkesin inancı kendine. ama o var ya o en azılı haindir., \textcolor{violet}{offensive}

\subsubsection{Vegans-vegetarians}

\textbf{heading}: vegan-cevreci-hayvansever-lezbiyen-solcu-feminist, \textbf{text}: vegan dışında beni tanımlamakta. ( 1 tanesi daha uymuyor ama onu söylemeyeceğim), \textcolor{violet}{neutral} \vspace{\baselineskip}

\noindent \textbf{heading}: vegan, \textbf{text}: farklı bir yaşam biçimidir. b*kunu çıkarıp mavi hap kırmızı hap kafasına girmemek lazım. zaten bu ülkede ve büyük ihtimal tüm dünyada, insanları bu akımdan en fazla soğutan şey bu akımda olan diğer kişiler., \textcolor{violet}{offensive}

\subsection{Turkish Hate Map}
\label{sec:eval-hate}

\subsubsection{Difficult instances}
\paragraph{Polite hate/offensive instances}
\textbf{heading}: recep-tayyip-erdogan, \textbf{text}: zincirin son halkası. (bkz: iyi uykular türkiye), \textcolor{violet}{offensive} \vspace{\baselineskip}

\noindent \textbf{heading}: arap, \textbf{text}: sınır denen birşeyin olmadığını sanıp yol alanlar var nasıl geliyorlar anlamıyorum, \textcolor{violet}{hate}

\paragraph{Cultural reference including instances}

\textbf{heading}: arap, \textbf{text}: mevcut iktidar tarafından, ait olmadığımız için sinir olunan ırk, \textcolor{violet}{hate} \vspace{\baselineskip}

\noindent \textbf{heading}: rtenin-2023-secimiyle-ilgili-rahatliginin-sebebi, \textbf{text}: kukla kemal, gel bakalım muharrem, d'hont erkan, seçim sonrası için meral, \textcolor{violet}{offensive}

\section{Evaluation using SentiTurca}
\subsection{Turkish Hate Map}
\label{sec:misclass-hate}
We present some misclassified instances per model here. Predicted labels are given in bold.

\subsubsection{BERTurk}
\label{sec:misclas-hate-bert}
\paragraph{True label: offensive}

fatih altaylı'nın da dediği gibi kaçmıştır, yükü taşıyamamıştır. eski sunucu yeni çiftçi. \textcolor{violet}{offensive} \vspace{\baselineskip}

\noindent artık kendi ülkelerine geri dönme vakitleri gelen araplar \textcolor{violet}{hate} \vspace{\baselineskip}

\noindent kukla kemal, gel bakalım muharrem, d'hont erkan, seçim sonrası için meral. \textcolor{violet}{neutral} \vspace{\baselineskip}

\noindent dinin siyasete alet edilmesinden daha iğrençtir laikliğin bu ülkede allah'ı inkar propagandasının aracı haline getirilmiş olması.. bu sebepledir bu ülke insanının din diye allah diye cemaat ve tarikatlara meyletmesi, rte gibi bir adama sarılması. işte bu yüzden sevilmez doğu perinçek gibiler de. ve işte bu yüzden doğu perinçek ile rte arasında bir fark yoktur. biri devrimcidir diğeri müslüman, ikisi de dünya çapında bir değişime öncülük ediyorlardır kendilerince. lakin gerçekte ikisi de ikiyüzlülüğün kitabını yazmışlardır. \textcolor{violet}{civilized} 

\paragraph{True label: hate}

böyle üremeye devam ederlerse 2030 da sığınmacı biz türkler oluruz bu gidişata göre.. benim akp ye ve oy verenlere (birisi öz abim misal) kafayı takma sebebim bu. başka hiçbir şeyi dert etmeyebilirim,tolere edebilirim akp icratları ile alakalı ama bu konuyu tolere edemiyorum aq. savaş bitti.. akp nin mütahitleri suriye ye ev mi yapar, toki binaları mı inşa eder bilemiyorum ama bir an önce s@@tir olsun gitsinler ülkemizden! \textcolor{violet}{offensive} \vspace{\baselineskip}

\noindent yarın sabah itibarı ile hpnginin anasını sikip ülkelerine geri yoLLaMak lazım s@@tirip gidip ülkelerini savunsunlar, burada çöpün yanında yaşayip cayır cayır sikişip tavşan gibi doğuracaklarına s@@tirip gidip ülkeleri için canlarını versinler. halepmiş şammış elbabmış neresiyse geldikleri yer s@@tirip gitsinler. bu orospu cocuklarına şahsım adına verdiğim 1 kuruş bile vergi gidiyorsa verenlerin soyu kurusun, ölüleri mezarda ters dönsün, ciğerleri patlasın, kötürüm kalsın. \textcolor{violet}{hate} \vspace{\baselineskip}

\noindent almanya'da olanları almanya ve belçika'da yaşanan sel felaketinin bıraktığı enkazı temizleme çalışmalarına çatır çatır da almanca konuşuyorlar. türkiye'deyse 10 yıldır türkiye'de yaşayıp hiç türkçe bilmeyen ve sel olsa gidip yağmalayacak tipler var ne yazık ki (elbette hpngi değil ama önemli bir bölümü). \textcolor{violet}{neutral}

\paragraph{True label: neutral}

iş çıkışı bir dolmuş durdurdum, tam yaklaşırken de pişman oldum ulan bu dolmuşa nasıl binicez diye. sol elim komple alçının içinde, kırdık işte bir şekilde. duşta falan poşet sarıp yıkandık ama dolmuş sıkıntılı. her neyse artık çok geç, dolmuşçu gözlerimin içine bakmakta insan kalabalığı içindeki ufacık aralıktan. ulan bundan sonraki de dolu gelicek hem adam durdu artık.. diye düşünürken içerde buldum kendimi. bi tane demir tuttum ama dengem yerinde değil, hass@@tir nasıl vericez lan dolmuş parasını cüzdan sol cepte sağ elle de tutunup hayatta kalmaya çalışıyorum, oha ya hava bu kadar sıcak mı? acayip ter bastı lan. işte tam o anda 2 tane türk kızı kalktı buyrun oturun zorlamayın kendinizi diye, birinin yerine oturdum mahçup bir şekilde bir de mırıldandım niye zahmet ettiniz diye. diğeri yerine tekrar oturdu. ayaktakine teşekkür edip paramı öne uzattıktan sonra buyrun oturun parayı yolladım gerisini hallederim dedim. oturun oturun sorun değil dedi ve işgüzar dolmuşçunun insanlık dışı doldurduğu dolmuşta ayakta yolculuğuna devam edip işgüzar dolmuşçunun isteği üzerine de polis kandırmaca niyetli biraz çömelebilir miyiz arzusunu da yerine getirdi. baya bir sevap kazanan tatlı türleri de vardır. not: ben yakışıklı değildim kız güzel değildi. \textcolor{violet}{offensive} \vspace{\baselineskip}

\noindent önce ülkedeki kaçakları sınırdışı edeniz. sonra türklere medeniyet dersi verirsiniz. \textcolor{violet}{hate} \vspace{\baselineskip}

\noindent gebze(artık çayırova) şekerpınar'da bir sürüsüne rastladığı, bir sürüsünün çocuğunu okuttuğum, temiz, kibar, dürüst insanlar. \textcolor{violet}{neutral} \vspace{\baselineskip}

\noindent mango meselesi biz veganlar arasinda da uzun yillardir tartisiliyor. genel kani su sekilde; 1a) kurdu gorunce bir anlik korkuyla canina okuduysan o gun veganligin dusuyor; ay sonundan itibaren 60 gun raw veganlik yapmak zorundasin. o gun veganligin dustugu icin cekirge kizartmasi yiyebiliyorsun. 1b) hic farketmeden kurdun canina okuduysan ve sonra farkettiysen veganligin devam ediyor. ancak mangoyu şaman esliginde buyuk bir saygiyla ormanda gommen gerekiyor. 2) bu gibi durumlarda o markaya ait mango suyunu sokaklara dokup visne suyu iciyoruz. \textcolor{violet}{civilized}

\paragraph{True label: civilized}

agnostik gorusun bir tik ilerisini savunan insandir kanimca. agnostiklere gore daha kibirli buluyorum acikcasi bu grubu. dinlerin sorgusuz sualsiz bir sekilde tanrinin varligini savunmasinda elestirdigim nokta kuskuya hicbir acik kapi birakmamasi. ama ateistler de bunun tam zittini yaparak ``tanri kesinlikle yoktur" demekten geri kalmiyorlar. boyle bakinca agnostik gorus candir dostlar. temennim herkesin agnostik olmasi yonundedir. sonucta tanrinin varligini ispatlamak icin sunulan verilerin yetersizligi ortada. baya bir zorlaya zorlaya mucize falan ornekleri veriyolar ama bu cagda bunlarin yetersiz kaldigi acik. ama ayni zamanda tanrinin olmadigini ispatlamak icin de yeterli kanit yok. uc bes tane ``madem tanri var;..." formatindaki sacma sapan soru benim acimdan tanrinin yoklugunu da kanitlamiyor. dolayisiyla, agnostik gorusun temelini olusturan ``tanrinin varligi ya da yoklugu bilinemez" dusuncesi bana tam uyuyor. ama ateistler bu lafim da size: birakin bu atarli ergen ayaklarini. kucuk, orta veya buyuk olcekte bir dag yaratmisliginiz yok. kendinizi bir bok sanmayin. dunyayi degistiren bilim adamlarinin ateist olmasini ornek gostererek de bu insanlardan nemalanmayi birakin. hadi bakalim; opuyorum hepinizi. \textcolor{violet}{offensive} \vspace{\baselineskip}

\noindent anlamamakta ısrarcı olanlar için şöyle kısa bi açıklama yapmak şart olmuş; şöyle ağız dolusu kadın diyemediğiniz için -bayan- diyosunuz. çünkü kadın demek ayıp. tıpkı vajina diyememek gibi. ha bu arada hitap ederken kullandığınız ``bayan"a bişey demiyoruz. derdimiz erkekler tuvaleti diyebiliyoken kadınlar tuvaleti diyememenizle. bayanlar tuvaleti demenizle. kendi kendinize evde uygulayabileceğiniz test: kadın kelimesini de erkek kelimesini kullandığınız rahatlıkla kullanabiliyo musunuz? \textcolor{violet}{neutral} \vspace{\baselineskip}

\noindent insana uygulanan şiddeti, kadın diye bir alt sınıfa ayırmak gibi cinsiyetçi yaklaşımları bırakmaktır. gören de sanacak sadece kadına uygulanan şiddet cezasız kalıyor. erkeklerin birbirine uyguladıkları şiddet çok daha fazla ve gereken cezalar verilmiyor. ancak medyanın ve insanların kadın şiddeti üzerinden demagoji yapmak daha çok işlerine geldiği için işin bu boyutunu görmek istemiyorlar. bir kadın şiddet gördüğünde ana akım medyada hemen yer alıp erkekler yerden yere vururken erkek şiddet gördüğünde kimsenin umurunda dahi olmuyor. \textcolor{violet}{civilized}

\subsubsection{Gemini 1.0 Pro and Claude 3.5 Sonnet}
\label{sec:misclas-hate-gemini}
\paragraph{True label: offensive}
suriyeli sağmalcılar olarak okunası başlık. ayrıca bazı ilçelere asıl sığınmaya çalışanlar bizim insanlarımız olmuş, oluyor, olmaya devam ediyor. \textcolor{violet}{offensive}

\noindent fatih altaylı'nın da dediği gibi kaçmıştır, yükü taşıyamamıştır. eski sunucu yeni çiftçi. \textcolor{violet}{hate} \vspace{\baselineskip}

\noindent fiziksel güçten bahsetmiyorum, kafa biraz gidik oluyor bir mevkiye geldiyse güç zehirlenmesi sanırım:p \textcolor{violet}{neutral} \vspace{\baselineskip}

\noindent suratına birkaç saniyeden fazla baktığınızda hemen asabileşen, özellikle kızlardan pek bir nefret eden, korkulası topluluk. \textcolor{violet}{civilized}

\paragraph{True label: hate}

suriyeli sığınmacılar ülkesine girdi diye mutluluktan göbek atan insanların işleri elinden alınacak diye yine kemalistlere gezicilere giydirmesine sebebiyet vermiş sığınmacılardır. suriyeli sığınmacıların işinizi elinizden alacak olması sevgili boşbakanınızın icraatıdır, gelip de bize ``niye mutlu oluyorsunuz?" lan diye ağlamayın... \textcolor{violet}{neutral} by Gemini Pro and \textcolor{violet}{offensive} by Claude 

\paragraph{True label: neutral}

laz olmak türk olmaya engel olmadığı için çelişki içermeyen durum. \textcolor{violet}{offensive} \vspace{\baselineskip}

\noindent neye göre güzel olacağımı kestirebilsem bir hak iddia edebileceğim durum olurdu. malum herkesin güzellik algısı çok farklı. \textcolor{violet}{hate} \vspace{\baselineskip}

\noindent dünyada onca sahipsiz çocuk varken insan kendi çocuğunu doğurmak zorunda mı? doğurmak istemeyebilir. \textcolor{violet}{neutral} \vspace{\baselineskip}

 \noindent hayvanların sırf zevk olsun diye kapalı kutu evlere hapsedilmesine şiddetle karşı bir insan olarak sünnilikle ilişkili midir bilemediğim sorunsaldır. \textcolor{violet}{civilized}

\paragraph{True label: civilized}

patrondan aldığınız maaşı net brüt diye ayırıp kafanızı karıştırıyorlar. sizin bir tane maaşınız var. sizin işverene filan yükünüz yok. işverenin senin bana maliyetin dediği şey aslında senin maaşın ve hakkın olan servis yemek vs. aslında senin maaşının tamamını cebine atıp sonra vergi dairesine gidip o ayki vergini vermen gerekiyor. ama böyle yaparlarsa sen işe uyanırsın ve devlete hesap sormaya başlarsın. devlet senin gelirinden alacağı vergiyi direkt patronundan alıyor. sana kalan maaşı veriyor. sen de işe ayıkmıyorsın. ya da kafanda bilsen bile para eline geçmediği için ve sen olayı fiziki olarak görmediğin için ciddiyetini anlamıyorsun. devlet sizin maaşınızdan daha alır almaz vergi alıyor. sonra da ekstra ev/araba/su/elektrik/iletişim vs gibi şeyler için ekstra vergi alıyor. daha sonra da siz elinizde kalan parayı harcarken yine ekstra kdv/ötv gibi vergiler veriyorsunuz. bunların üzerine yollarda giderken yine geçtiğiniz yolun parasını veriyorsunuz. herhangi bir doğal afette yine pamuk eller cebe denerek para toplanıyor. futbol kulüplerinin ve büyük şirketlerin vergi borçları siliniyor. siz bir telefon/oyun konsolu/pc/araba alırken paramız yurtdışına akmasın döviz açığı olmasın diye devlete inanılmaz vergiler ödüyorsunuz ekstradan. ama futbol kulüpleri tek bir oyuncuya milyonlarca euro verebiliyor bu oyuncular 0 katkı sağlasa bile kimse hesap vermiyor üstelik. ayrıca vergi borçları da siliniyor. bütün bunlar bana inanılmaz geliyor. gerçekten bunlar yaşanıyor bu ülkede. \textcolor{violet}{offensive} 

\subsubsection{GPT-4 Turbo and LLaMa 3 70B}
\label{sec:misclas-hate-LLaMa}
\paragraph{True label: offensive}

koltuğunu bir an önce bırakması gereken kişidir.bırak başkası geçsin amk yapamıyon işte olmuyo devletçiğim bırak gitsin ölecen yakında koltukta \textcolor{violet}{offensive} \vspace{\baselineskip}

\noindent suriyeli sağmalcılar olarak okunası başlık. ayrıca bazı ilçelere asıl sığınmaya çalışanlar bizim insanlarımız olmuş, oluyor, olmaya devam ediyor. \textcolor{violet}{hate} \vspace{\baselineskip}

\noindent fatih altaylı'nın da dediği gibi kaçmıştır, yükü taşıyamamıştır. eski sunucu yeni çiftçi. \textcolor{violet}{neutral} \vspace{\baselineskip}

\noindent fiziksel güçten bahsetmiyorum, kafa biraz gidik oluyor bir mevkiye geldiyse güç zehirlenmesi sanırım:p \textcolor{violet}{civilized} by GPT-4

\paragraph{True label: hate}

suriyeli sığınmacılar ülkesine girdi diye mutluluktan göbek atan insanların işleri elinden alınacak diye yine kemalistlere gezicilere giydirmesine sebebiyet vermiş sığınmacılardır. suriyeli sığınmacıların işinizi elinizden alacak olması sevgili boşbakanınızın icraatıdır, gelip de bize ``niye mutlu oluyorsunuz?" lan diye ağlamayın... \textcolor{violet}{neutral} by GPT-4, \textcolor{violet}{hate} by LLaMa 3

\paragraph{True label: neutral}

lütfü türkkan'ı acilen ihraç etmesi gereken parti. \textcolor{violet}{offensive} \vspace{\baselineskip}

\noindent zeki kadınlara tapılır. zeki kadın değerlidir. başlığı açan yazar yanılıyor. (bkz: sarhoş ellam) \textcolor{violet}{hate} by GPT-4, correctly classified by LLaMa 3 \vspace{\baselineskip}

\noindent numaranı versene, arıyım seni? + tabii. 0212 43... - (bağlantı kesildi) komple kabloyu sökerek kaçmış. \textcolor{violet}{neutral} \vspace{\baselineskip}

\noindent gerekli olandır. yönetime din karıştırılırsa ilerleme beklemek mümkün değildir; inanç konusu tamamen kişiye özelken yönetim hepimizi ilgilendirir, bu da başlıkla ilgili, 5 yaşındaki bir çocuğun anlayabileceği bir açıklamadır. \textcolor{violet}{civilized}

\paragraph{True label: civilized}

patrondan aldığınız maaşı net brüt diye ayırıp kafanızı karıştırıyorlar. sizin bir tane maaşınız var. sizin işverene filan yükünüz yok. işverenin senin bana maliyetin dediği şey aslında senin maaşın ve hakkın olan servis yemek vs. aslında senin maaşının tamamını cebine atıp sonra vergi dairesine gidip o ayki vergini vermen gerekiyor. ama böyle yaparlarsa sen işe uyanırsın ve devlete hesap sormaya başlarsın. devlet senin gelirinden alacağı vergiyi direkt patronundan alıyor. sana kalan maaşı veriyor. sen de işe ayıkmıyorsın. ya da kafanda bilsen bile para eline geçmediği için ve sen olayı fiziki olarak görmediğin için ciddiyetini anlamıyorsun. devlet sizin maaşınızdan daha alır almaz vergi alıyor. sonra da ekstra ev/araba/su/elektrik/iletişim vs gibi şeyler için ekstra vergi alıyor. daha sonra da siz elinizde kalan parayı harcarken yine ekstra kdv/ötv gibi vergiler veriyorsunuz. bunların üzerine yollarda giderken yine geçtiğiniz yolun parasını veriyorsunuz. herhangi bir doğal afette yine pamuk eller cebe denerek para toplanıyor. futbol kulüplerinin ve büyük şirketlerin vergi borçları siliniyor. siz bir telefon/oyun konsolu/pc/araba alırken paramız yurtdışına akmasın döviz açığı olmasın diye devlete inanılmaz vergiler ödüyorsunuz ekstradan. ama futbol kulüpleri tek bir oyuncuya milyonlarca euro verebiliyor bu oyuncular 0 katkı sağlasa bile kimse hesap vermiyor üstelik. ayrıca vergi borçları da siliniyor. bütün bunlar bana inanılmaz geliyor. gerçekten bunlar yaşanıyor bu ülkede. \textcolor{violet}{hate} by GPT-4, \textcolor{violet}{offensive} by LLaMa 3

\subsubsection{Qwen2-72B}
\label{sec:misclas-hate-qwen}
\paragraph{Misclassified neutral instances}

benim için ters olan, ama gel gelelim mıknatıs gibi çektiğim, hayır dediğimde de türlü türlü laflar yediğim olay tanım : bir çeşit namüsait durum. \vspace{\baselineskip}

\noindent en büyük hatanız insanları genellemeniz. zeki olsalar o adamları başlarında tutarlar mıydı? kimden bahsettim hadi bilin, sağ? sol? bilemeyecek kadar ince artık çizgi   \vspace{\baselineskip}

\noindent gezegenimizde en az otuz yıl geçirmiştir \vspace{\baselineskip}

\noindent tersi uygulamalar da mevcuttur. anadolu'da sünniler alevilerden kız alırlar ama vermezler. gerekçesi: aleviler sünnileri yezid taraftarı olarak görürler. halbuki tüm sünniler emevi ve yezid fanı değildir. sünniler de alevileri islam dışı olarak görürler. halbuki islam dışı olmak için kitabı ve peygamberi inkar etmek gerekir. bu söylediğim genel profildir. bu şekilde düşünmeyenler de elbette vardır.

\end{appendices}

%%===========================================================================================%%
%% If you are submitting to one of the Nature Portfolio journals, using the eJP submission   %%
%% system, please include the references within the manuscript file itself. You may do this  %%
%% by copying the reference list from your .bbl file, paste it into the main manuscript .tex %%
%% file, and delete the associated \verb+\bibliography+ commands.                            %%
%%===========================================================================================%%

\bibliography{sn-bibliography}% common bib file
%% if required, the content of .bbl file can be included here once bbl is generated
%%\input sn-article.bbl

\end{document}